\documentclass{egpubl}
\usepackage{amsmath,amsfonts}
\usepackage{algorithmic}
\usepackage{algorithm}
\usepackage{array}
\usepackage[caption=false,font=normalsize,labelfont=sf,textfont=sf]{subfig}
\usepackage{textcomp}
\usepackage{stfloats}
\usepackage{multirow}
\usepackage{url}
\usepackage{verbatim}
\usepackage{booktabs}
\usepackage{cite}
\usepackage{colortbl}
\usepackage[x11names]{xcolor}
\definecolor{gold}{HTML}{D4AF37}
\definecolor{silver}{HTML}{C0C0C0}
\definecolor{bronze}{HTML}{CD7F32}
\definecolor{first}{HTML}{F48288}
\definecolor{second}{HTML}{FAC791}
\definecolor{third}{HTML}{FFFF8F}
\newcommand{\coloredcircle}[1]{\textcolor{#1}{\rule{6pt}{6pt}}}

\makeatletter
\def\ps@titlepage{%
  \let\@mkboth\@gobbletwo
  \def\@oddhead{}\def\@evenhead{}%
  \def\@oddfoot{}\def\@evenfoot{}%
  \let\sectionmark\EmptySectionmark
  \let\subsectionmark\EmptySubsectionmark}
\makeatother
\usepackage[T1]{fontenc}
\usepackage{dfadobe}
\BibtexOrBiblatex
\electronicVersion
\PrintedOrElectronic
\usepackage{graphicx}
\ifpdf \pdfcompresslevel=9 \fi
\usepackage{egweblnk}

\title{Struct-GStream: Towards Efficient Free-Viewpoint Video Streaming at Low-Bitrates  with Structured 3D Gaussians}

\author[H.\,Jiao, J.\,Sun, L.\,Zhao, et al.]
{\parbox{\textwidth}{\centering 
        H.\,Jiao$^{1\dagger}$, 
        J.\,Sun$^{1\dagger}$, 
        L.\,Zhao$^{1*}$, 
        W.\,Xing$^{1}$, 
        H.\,Lin$^{1}$, 
        Z.\,Zhang$^{1}$ 
        and A.\,Ma$^{2,3}$\\[-2pt]
        {\small $^{\dagger}$ Joint first authors. \quad $^{*}$ Corresponding author: lzhao@zju.edu.cn}
        }
        \\
{\parbox{\textwidth}{\centering 
        $^1$ College of Computer Science and Technology, Zhejiang University, China\\
        $^2$ Chinese Academy of Sciences, Beijing, China\\
        $^3$ University of Science and Technology Beijing, China
       }
}
}

\begin{document}


\maketitle
\hypersetup{
    pdftitle={Struct-GStream: Towards Efficient Free-Viewpoint Video Streaming at Low-Bitrates with Structured 3D Gaussians},
    pdfauthor={H. Jiao, J. Sun, L. Zhao, W. Xing, H. Lin, Z. Zhang, A. Ma},
    pdfsubject={Preprint},
    pdfkeywords={free-viewpoint video, 3D Gaussian Splatting, streaming, neural rendering}
}
\begin{abstract}
    Constructing photorealistic Free-Viewpoint Videos (FVVs) of
    dynamic scenes from a set of posed 2D images has
    been an intriguing yet challenging task in computer vision.
    Methods based on neural rendering achieve high-fidelity image quality in FVV construction.
    However, most of these methods are unable to achieve real-time rendering and often
    require complete video sequences to train.
    Despite the existence of some online training methods capable of rendering FVVs in real time,
    they struggle to meet the requirements for storage and training time for downstream applications.
    To overcome this problem, we propose Struct-GStream, which can achieve
    efficient FVVs streaming using structured 3D Gaussians (3DGs).
    Specifically, we introduce dynamic anchor points to generate structured 3DGs to construct basic scenes and
    model approximate scene movements based on the assumption of local rigidity in object motion.
    Besides, we introduce a global free 3DGs patching strategy involving free 3DGs' generation,
    pruning, and optimization to patch and model deficient areas and emerging objects.
    Our method achieves fast training at low-bitrates while maintaining high rendering quality.
    Extensive experiments demonstrate that Struct-GStream
    significantly outperforms existing online training methods of FVV construction in terms of training time, storage, and rendering quality
    while maintaining competitive rendering speed.

\ccsdesc[500]{Computing methodologies~Reconstruction}

\printccsdesc   
\end{abstract}  
\section{Introduction}
Constructing Photorealistic Free-Viewpoint Videos of real-world
dynamic scenes with multi-view inputs has
been a continuous endeavor in computer vision.
This technology enables users to freely explore dynamic scenes from novel views at different times,
demonstrating significant potential and value in multiple fields,
including virtual reality/augmented reality.
Its broad application prospects have attracted the attention of many researchers.
Conventional methods for constructing FVVs are primarily divided into two
categories: geometry-based methods~\cite{HSFVV,Motion2fusion} that explicitly reconstruct dynamic meshes or points,
and image-based methods~\cite{interpolation_video, Immersivelightfield} that use interpolation to obtain novel views.
However, both of these methods struggle to handle scenes with complex occlusions and textureless regions.

  
\begin{figure}[!t]
  \centering
  \subfloat[\textnormal{I-NGP: Per-frame}]{\includegraphics[width=0.23\textwidth]{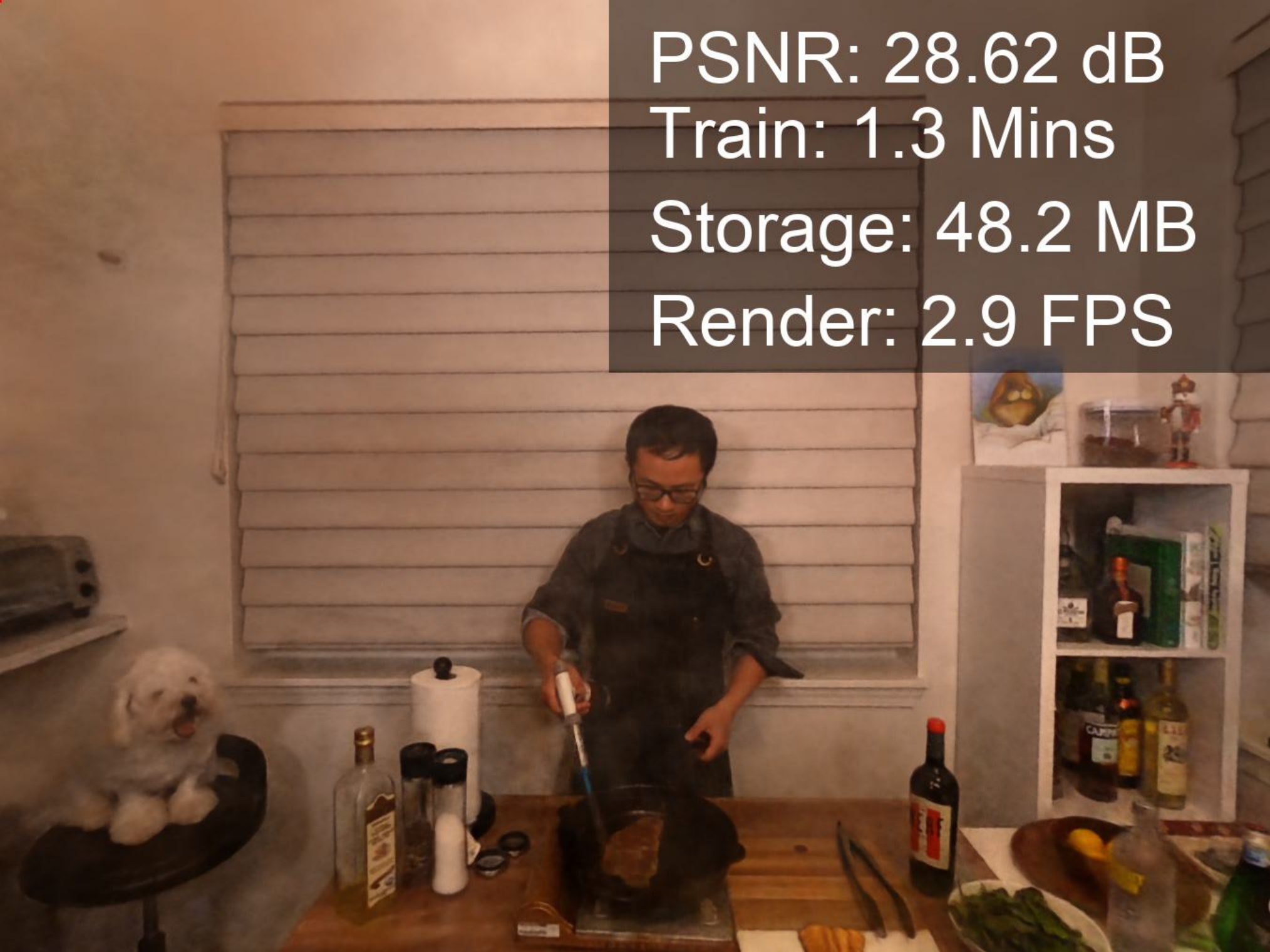}}
  \hfill
  \subfloat[\textnormal{StreamRF: Online}]{\includegraphics[width=0.23\textwidth]{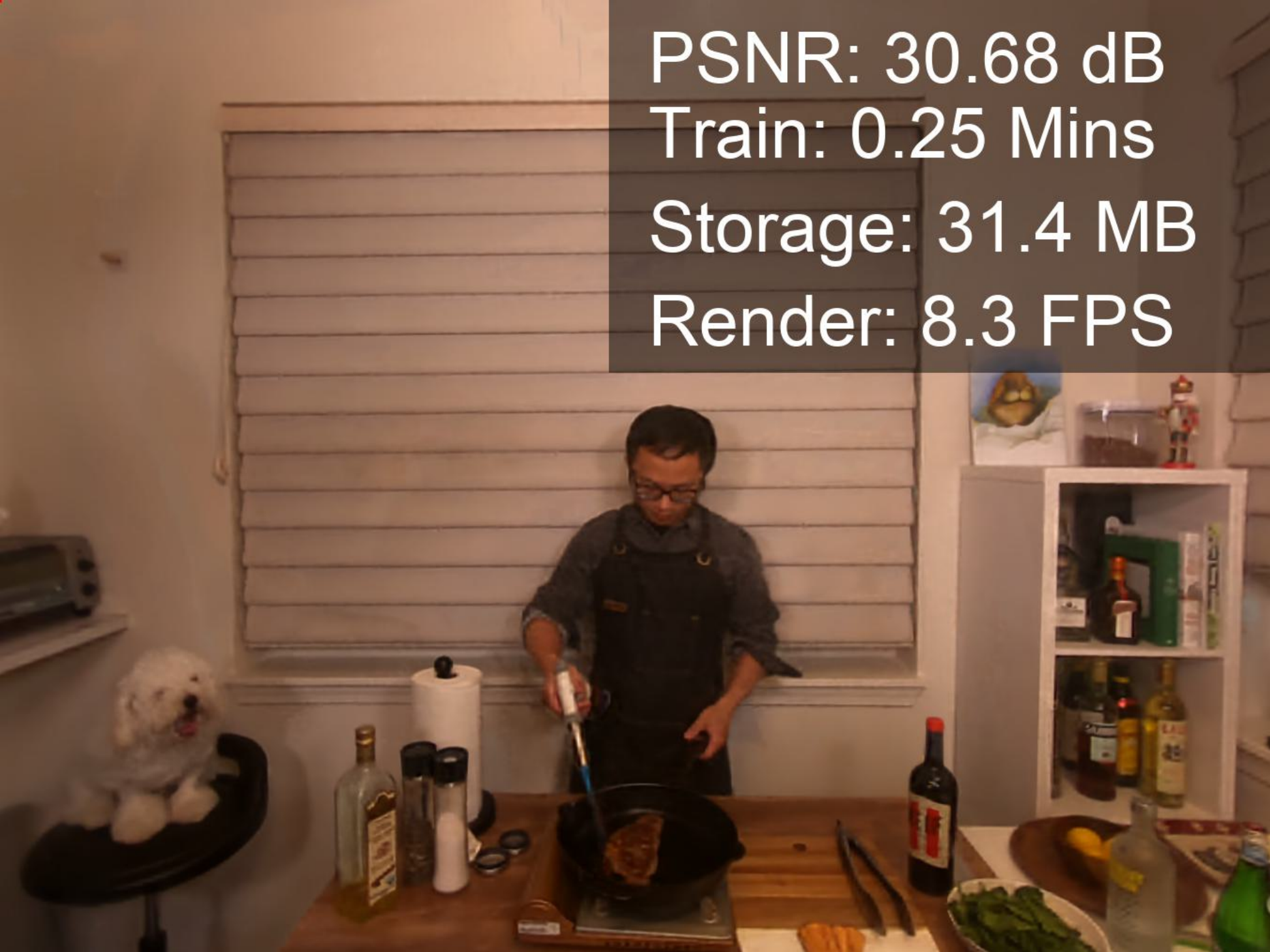}}
  
  \vspace{-0.15cm}
  \subfloat[\textnormal{3DGStream: Online}]{\includegraphics[width=0.23\textwidth]{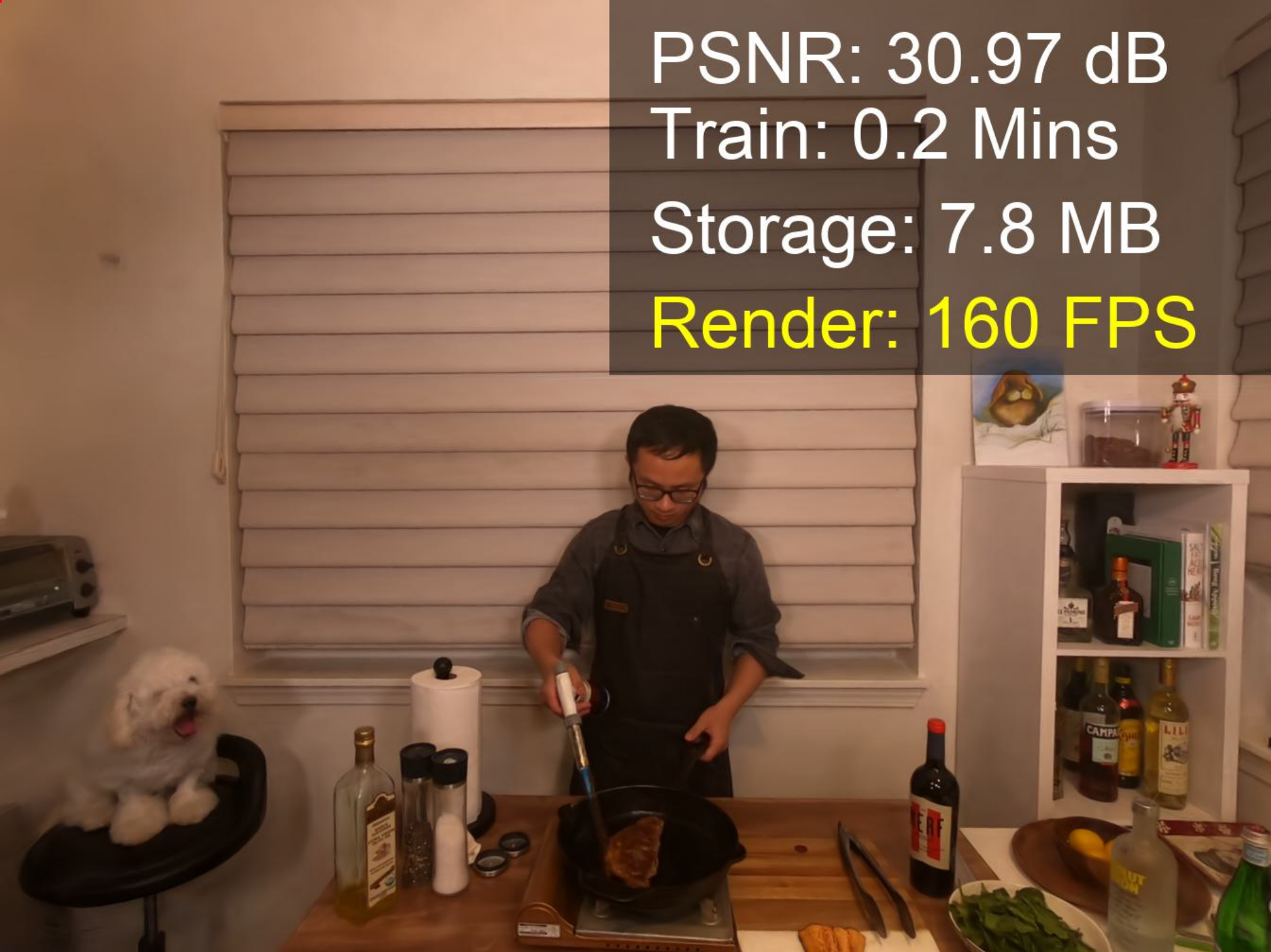}}
  \hfill
  \subfloat[\textnormal{Ours: Online}]{\includegraphics[width=0.23\textwidth]{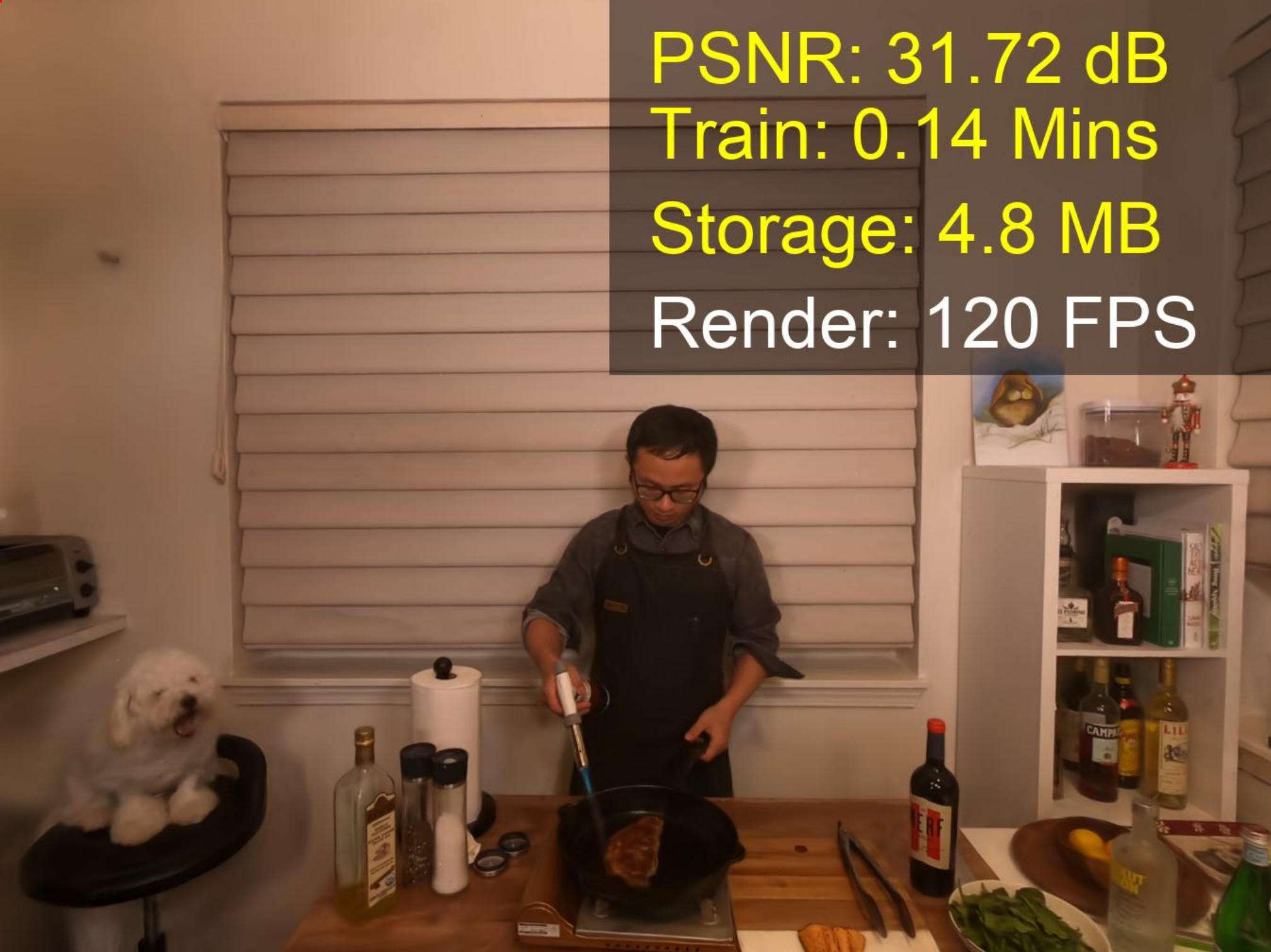}}
  \caption{Comparison on the flame steak from the N3DV dataset.}
  \label{fig:teaser_1}
\end{figure}
\par Recently, the emergence of Neural Radiance Fields (NeRF)\cite{NeRF} has substantially enhanced the ability to model static scenes under multi-view inputs, inspiring a vast body of subsequent work in various directions~\cite{https://doi.org/10.1111/cgf.70012, https://doi.org/10.1111/cgf.14449,https://doi.org/10.1111/cgf.14646,https://doi.org/10.1111/cgf.14929}. Among these advancements, many researchers have further explored the field of Free-Viewpoint Video (FVV) construction based on NeRF, producing numerous outstanding works~\cite{D-NeRF,NeuralSceneFlow,SpacetimeNerf,Hypernerf,Nerfies,StreamRF,DyNeRF,TiNeuVox, Dynibar, TemporalInterpolation,K-Planes,Hexplane}. However, these methods often face several challenges: (1) They typically require complete video sequences for training; (2) They struggle to achieve real-time rendering.

\par The recent research achievement, 3D Gaussian Splatting (3DG-S)\cite{3DGS}, has successfully achieved high-quality modeling of static scenes and significantly outperformed previous NeRF-related works in terms of rendering speed, which in turn has generated a wide variety of 3DG-S-based works~\cite{https://doi.org/10.1111/cgf.70309, https://doi.org/10.1111/cgf.70265, https://doi.org/10.1111/cgf.70271, https://doi.org/10.1111/cgf.70256,https://doi.org/10.1111/cgf.70227, https://doi.org/10.1111/cgf.70270}. Its significant advantages in reconstruction and rendering have attracted researchers to apply it in FVV construction, leading to numerous 3DG-S-related methods~\cite{SpacetimeGaussians,dynamic3dg,SC-GS,hifi4g,NeuralParaGaussian,superpointgs,3dgstream,gaussianflow}.
Despite achieving real-time rendering and high-quality FVV construction, most of these methods still rely on complete video sequences for training.
The few online training methods, such as 3DGStream~\cite{3dgstream} and Dynamic3DG~\cite{dynamic3dg} are constrained by requirements like storage, training time, or scene masks,
which significantly impede their adoption in practical applications.
To reduce the storage and training time required for online FVV construction, we propose Struct-GStream,
towards efficient FVV streaming at low-bitrates.
Our method represents the scene as structured 3DGs and free 3DGs.
Specifically, we first construct structured 3DGs for initialization based on the posed 2D images of frame 0.
For any subsequent frame $i$, we initialize by inheriting the structured 3DGs
from the previous frame along with free 3DGs if available, and then we conduct a two-stage optimization process:
(1) We jointly optimize the positions, scaling, and rotations of structured 3DGs as well as
all attributes of free 3DGs, without pruning or generating them.
(2) We carry out a global free 3DGs patching strategy in which we prune and generate
free 3DGs to compensate for reconstruction errors or represent emerging objects.
As shown in Fig. \textcolor{red}{\ref{fig:teaser_1}},
when compared to online FVV construction methods, our method outperforms them in
storage, training time, rendering quality, and achieves competitive rendering speed.
\begin{figure}[!t]
    \centering
    \includegraphics[width=\linewidth]{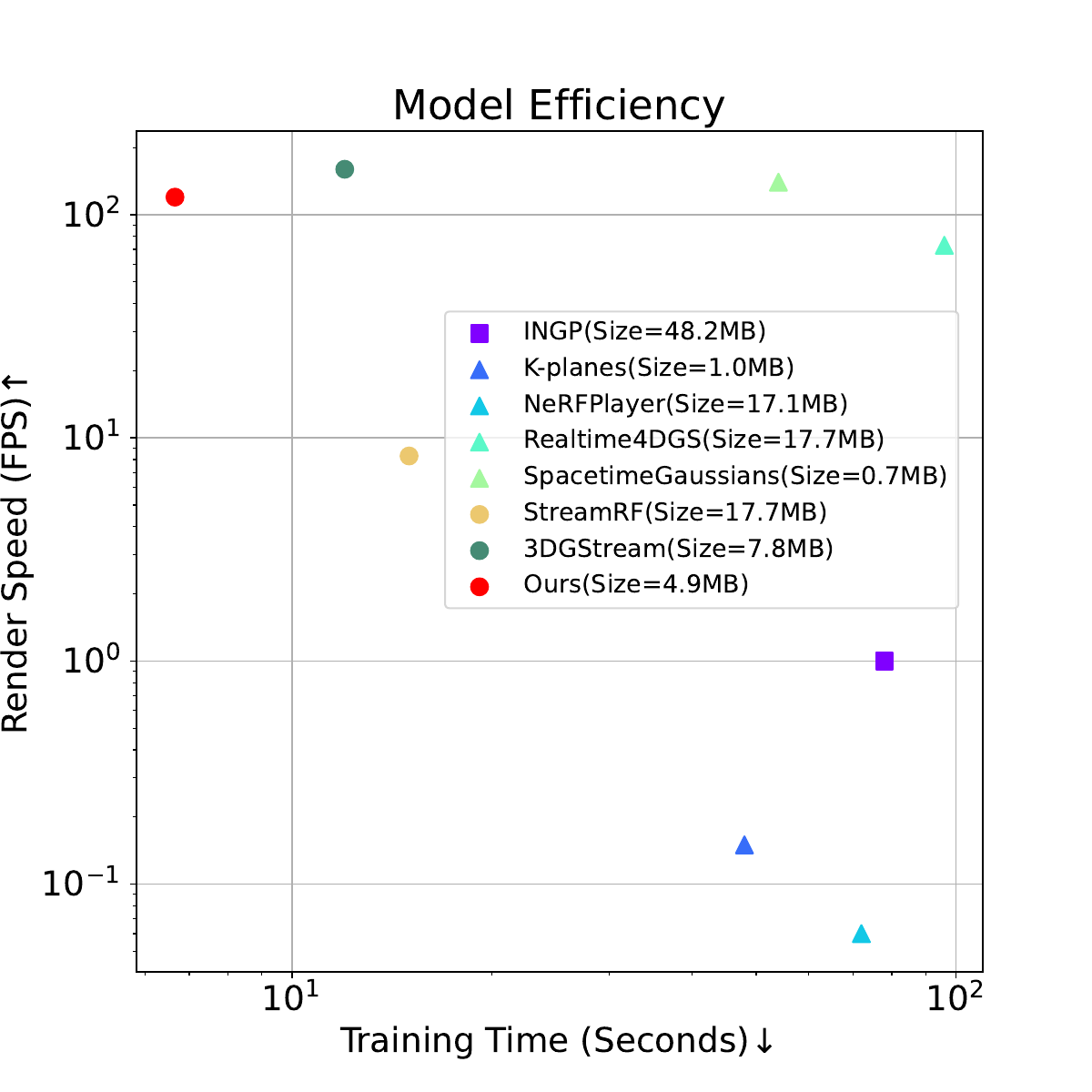}
    \caption{\textbf{Comparison of our method with other methods on the N3DV dataset.}
        $\Box$, $\bigtriangleup$ and $\bigcirc$ respectively represent training from scratch per frame,
        offline training on complete video sequences, and online training on video streams.
        Our method demonstrates state-of-the-art performance in both storage and training time
        when compared with existing online training methods,
        while achieving competitive rendering speed.}
    \label{fig:teaser_2}
\end{figure}

\par Our contributions can be summarized as follows:

\begin{itemize}
    \item We propose a novel online FVV construction method called Struct-GStream, leveraging structured 3DGs
          to achieve fast training of FVVs under low storage conditions while maintaining high rendering quality.
    \item We introduce dynamic anchor points into dynamic scene representation to generate structured 3DGs to
          model approximate scene movements based on the assumption of local rigidity in object motion.
          
    \item We employ a global free 3DGs patching strategy for generating and pruning free 3DGs, allowing for the patching and modeling of deficient areas and emerging objects in dynamic scenes
          and maintaining good temporal consistency for newly emerging objects.
    \item Extensive experiments demonstrate that Struct-GStream
          significantly outperforms existing online training methods in FVV construction in terms of training time, rendering quality and storage
          while maintaining competitive rendering speed.
\end{itemize}

\section{Related work}

\begin{figure*}[!t]
    \centering
    \includegraphics[width=\linewidth]{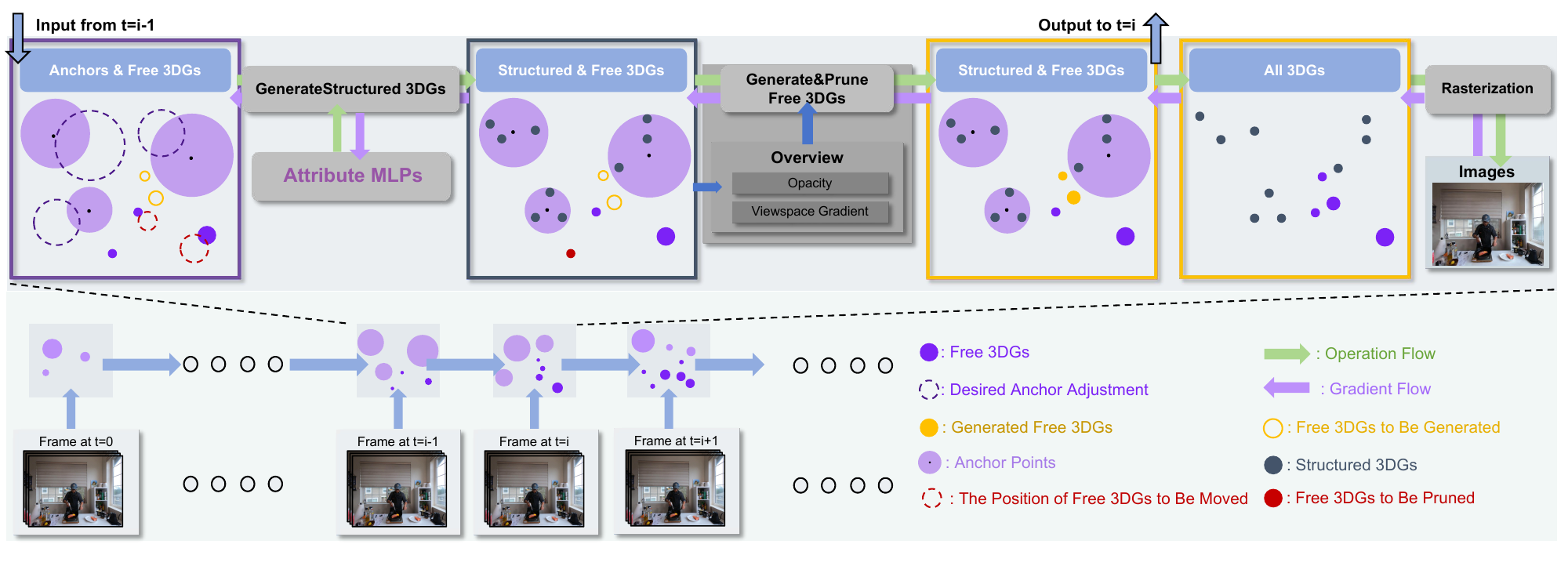}
    \caption{Overview of Struct-GStream. At frame 0, 
    we initialize the dynamic anchor points used for generating structured 3DGs. 
    For any other specific frame $i$, 
    we inherit the dynamic anchor points and free 3DGs (if any) 
    from the previous frame, and then, we prune the inherited free 3DGs 
    based on opacity. Finally, we generate new free 3DGs around all 3DGs whose viewspace 
    positional gradients exceed a certain threshold.
}
\label{fig:overview}
\end{figure*}
\subsection{Novel View Synthesis for Static Scenes}
The field of novel view synthesis has garnered increasing attention in recent years.
Traditional methods~\cite{plenoptic,lumigraph,lightfield} generate new viewpoint rays through interpolation,
and by caching these rays, they achieve real-time rendering.
In recent years, NeRF~\cite{NeRF},
a method utilizing a multi-layer perceptron (MLP) to learn neural radiance fields
with differentiable volume rendering, has ignited a research surge in novel view synthesis
due to their superior capability in synthesizing images from arbitrary viewpoints.
A series of enhancement works have since been developed based on NeRF. Some methods seek to achieve real-time rendering
\cite{fastnerf,baking,nex,plenoctrees,merf,mobilenerf} and some methods seek to achieve
fast training~\cite{plenoxels,Instant-NGP,tensorf,dicfields,trimiprf} for NeRF.
However, these efforts have not completely moved away from neural radiance fields which require numerous spatial point queries.
\par 3DG-S~\cite{3DGS} uses anisotropic 3D Gaussians as scene representations, which can effectively reconstruct scene structures and express scene details.
It not only leverages the advantages of volume rendering
to realize high-fidelity image synthesis,
but also offers a differentiable point-based rendering scheme for
real-time rendering.

\subsection{Free-Viewpoint Videos of Dynamic Scenes}
Early works on FVV construction of dynamic scenes~\cite{interpolation_video,HSFVV,Motion2fusion,Immersivelightfield}
typically employ methods related to dynamic primitives or interpolation.
With the success of NeRF in novel view synthesis for static scenes,
NeRF-related methods for dynamic scene reconstruction have begun to emerge in large numbers.
Warp-based methods~\cite{D-NeRF,NeuralSceneFlow,SpacetimeNerf,nonrigid,Nerfies,Hypernerf,nerfplayer,forwardflow}
use a deformation field to decouple the time and radiance fields, effectively handling rigid motion
by mapping points via the deformation field onto a canonical field that varies with time.
However, they are limited in capturing scenes with significant topological changes due to their foundational assumption
that the motion of the scene arises from the deformation of static structure.
There are also methods~\cite{TiNeuVox,fourier,TemporalInterpolation,K-Planes,Hexplane,mixedvoxel}
that interpolate time-space inputs,
using spatio-temporal coordinates as input to enhance the radiance field.
These methods have compact storage and satisfactory novel viewpoint synthesis ability
in FVV construction,
but the entangled input parameters limit their adaptability to downstream applications.
StreamRF~\cite{StreamRF} is an outstanding NeRF-related online training method
that can stream FVVs during the process of online training.
However, its rendering speed is unsatisfactory.
\par
With the advent of 3DG-S\cite{3DGS}, dynamic 3DG-S-related method has also attracted significant attention
due to its outstanding rendering speed and excellent explicit representation.
Some methods ~\cite{deformable3DGS,4dgs_kplanes} introduce deformation field to 3DG-S's dynamic scene representation by directly
using time and 3DG position as input or using planes to indirectly express 3DGs' spatio-temporal feature.
4D Gaussian-based methods~\cite{4dgs_towards,realtime4dgs} use more direct 4D XYZT Gaussians for dynamic scene representation
and model dynamics at each timestamp by temporally slicing the 4D Gaussian.
Curve-based methods~\cite{SpacetimeGaussians,gaussianflow} achieve fast,
high-quality 4D scene reconstruction by fitting Gaussian attribute variation
curves using polynomials or fourier coefficients.
Control point-based methods~\cite{SC-GS,hifi4g,superpointgs,NeuralParaGaussian}
achieve smooth motion trajectories and outstanding novel viewpoint synthesis
by using control points
to control local 3D Gaussians' trajectory or even more attributes. Online training methods~\cite{dynamic3dg,3dgstream,Tang_Yang_Peng_Zhai_Shen_Wang_2025} achieve online high-quality FVV construction by adjusting, spawning, or pruning 3DGs frame by frame. Among them, 3DGStream~\cite{3dgstream} delivers excellent rendering quality and speed but faces challenges in model size and training time. 
iFVC~\cite{Tang_Yang_Peng_Zhai_Shen_Wang_2025} employs a distinct optimization strategy primarily focused on extreme compression for volumetric video streaming.
In contrast, our method aims for an optimal balance between low-bitrate storage and fast online training, significantly reducing training overhead compared to previous works while maintaining competitive rendering quality and speed.
\subsection{Free-Viewpoint Video Coding and Streaming}
Traditional 2D video coding standards, such as H.264~\cite{1218189} and HEVC~\cite{6316136}, have achieved remarkable success in efficient video storage and transmission. However, constructing streamable Free-Viewpoint Videos (FVVs) from multi-view inputs introduces new challenges, requiring compact representations that handle complex 3D geometry and view-dependent effects, which cannot be directly addressed by standard 2D compression techniques.

With the advent of NeRF~\cite{NeRF} and 3DG-S~\cite{3DGS}, an increasing number of methods have focused on the compression of free-viewpoint videos; however, most of these approaches~\cite{K-Planes,Hexplane,hu2024vrvvcvariableratenerfbasedvolumetric} do not support online streaming of free-viewpoint videos. Among the few methods that enable streaming and coding, StreamRF~\cite{StreamRF} pioneered online streaming for radiance fields by transforming the task into a dynamic difference modeling problem. More recently, 3DG-S-based methods have emerged: 3DGStream~\cite{3dgstream} achieves high-fidelity online FVV construction using a compact Neural Transformation Cache based on I-NGP~\cite{Instant-NGP} to encode frame-wise attribute variations of 3D Gaussians. Similarly, iFVC~\cite{Tang_Yang_Peng_Zhai_Shen_Wang_2025} introduces a Binary Transformation Cache~\cite{shin2023binaryradiancefields} to further compress the grid structures storing attribute changes.

In a different vein, rather than focusing on optimizing grid-based neural architectures for extreme compression, our Struct-GStream explores the potential of explicit structured representations. By introducing structured 3DGs and optimizing the online training scheme, we aim to strike an optimal balance between online training efficiency, storage compactness, and reconstruction quality.

\section{Preliminaries}
3D Gaussian Splatting~\cite{3DGS} uses a collection of colored 3DGs
as the explicit representation of scenes. It splats these 3DGs
onto view space using a tile-based differentiable rasterizer,
enabling real-time rendering of high-quality novel viewpoint images.
Each 3DG $G$ has a mean vector $\mu$ representing its 3D spatial position,
as well as a 3D anisotropic covariance matrix $\Sigma$ representing its rotation
and scaling:
\begin{equation}
    G(x;\mu,\Sigma) = e^{-\frac{1}{2}(x-\mu)^T\Sigma^{-1}(x-\mu)},
\end{equation}
where $\Sigma$ is positive semi-definite and
can be decomposed into a rotation matrix $R$ and a scaling matrix $S$:
\begin{equation}
    \Sigma=RSS^{T}R^{T}.
\end{equation}
The rotation matrix $R$ can be parameterized by a quaternion $q$,
and the scaling matrix $S$ is a diagonal matrix parameterized by a 3D vector $s$.
Furthermore, each 3DG has a set of spherical harmonics coefficients $sh$ to
represent the view-dependent color, along with an opacity value $\alpha$.


\begin{figure*}[!t]
    \centering
    \subfloat{\includegraphics[width=0.19\textwidth]{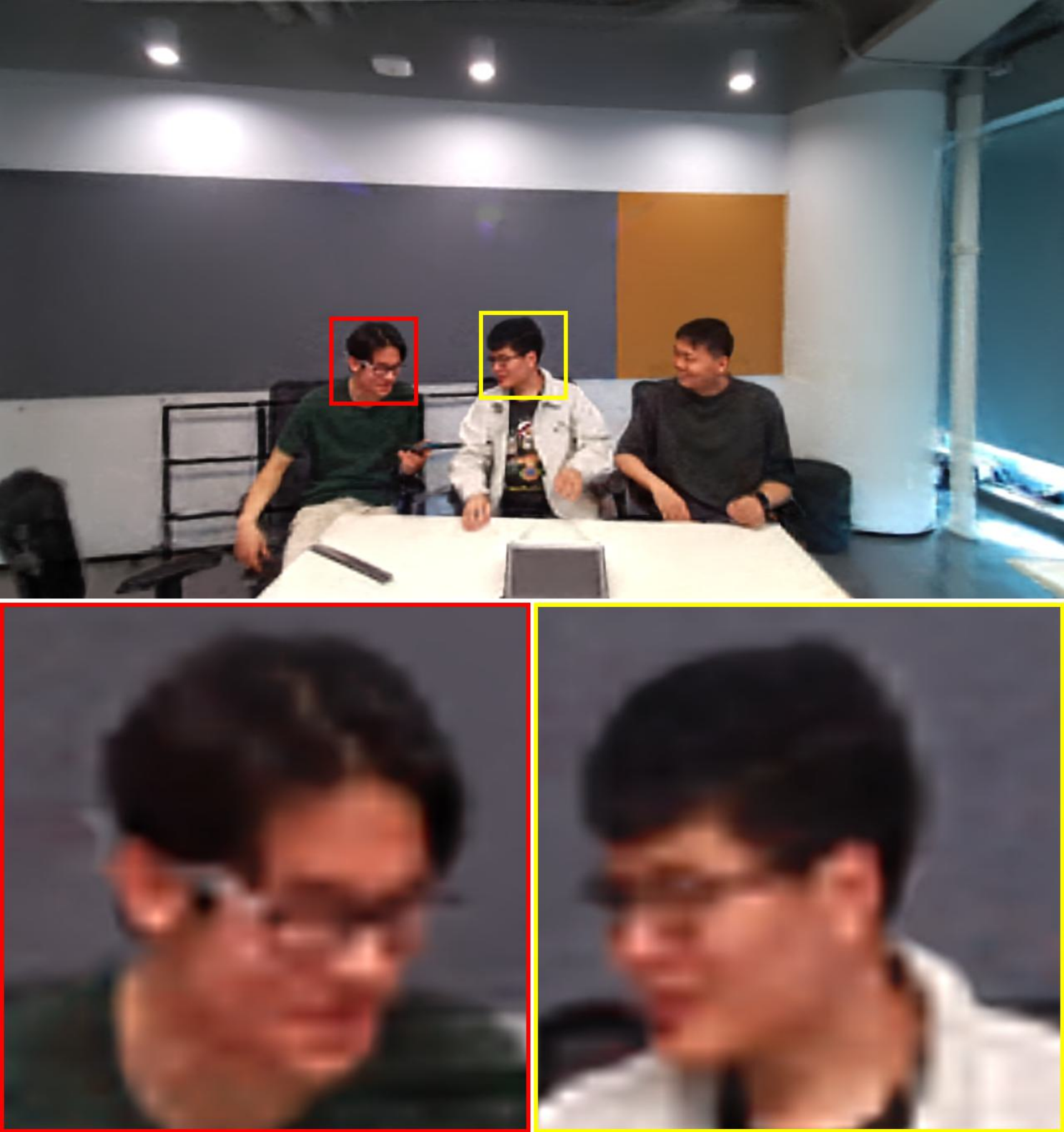}}
    \hfill
    \subfloat{\includegraphics[width=0.19\textwidth]{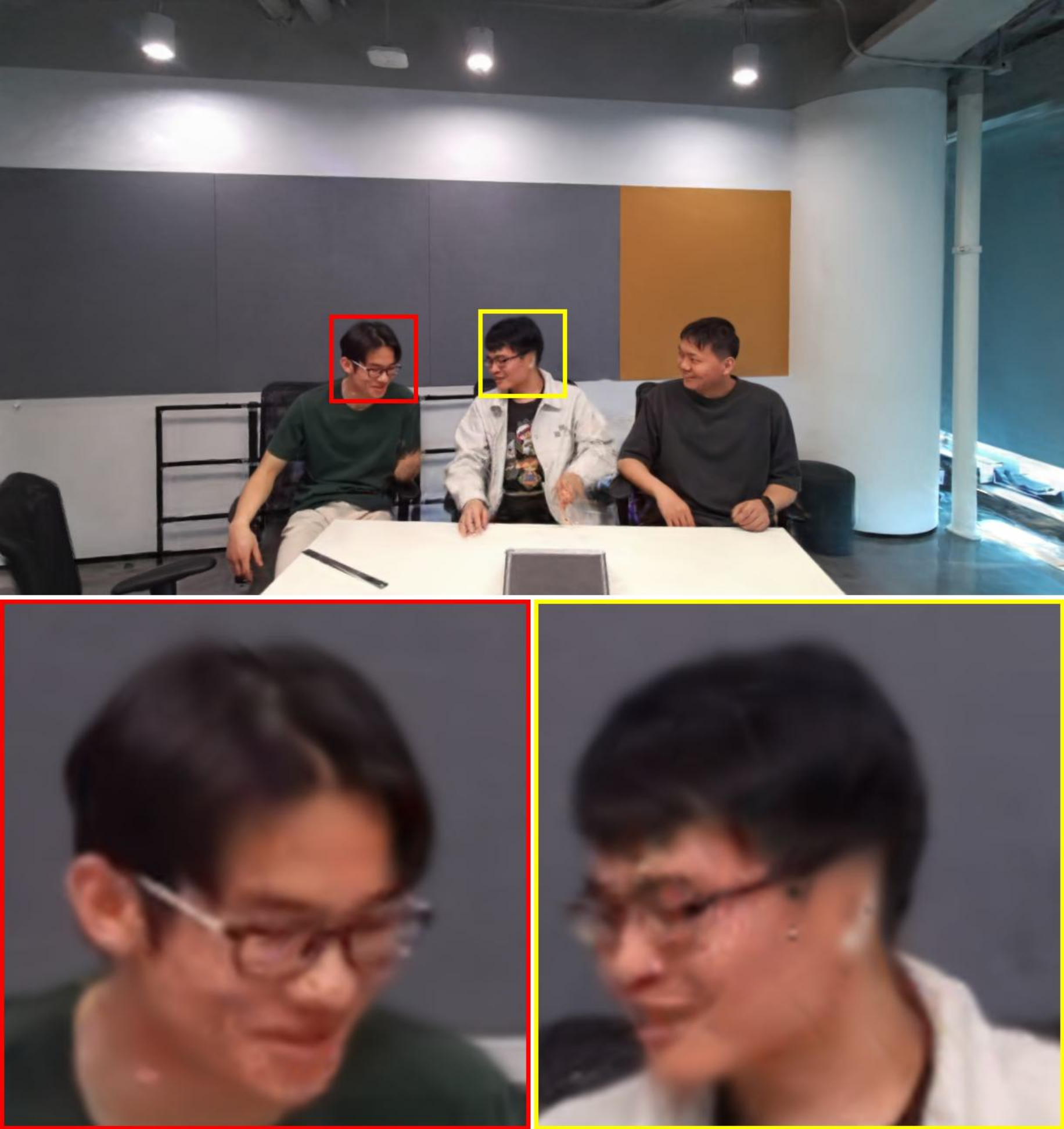}}
    \hfill
    \subfloat{\includegraphics[width=0.19\textwidth]{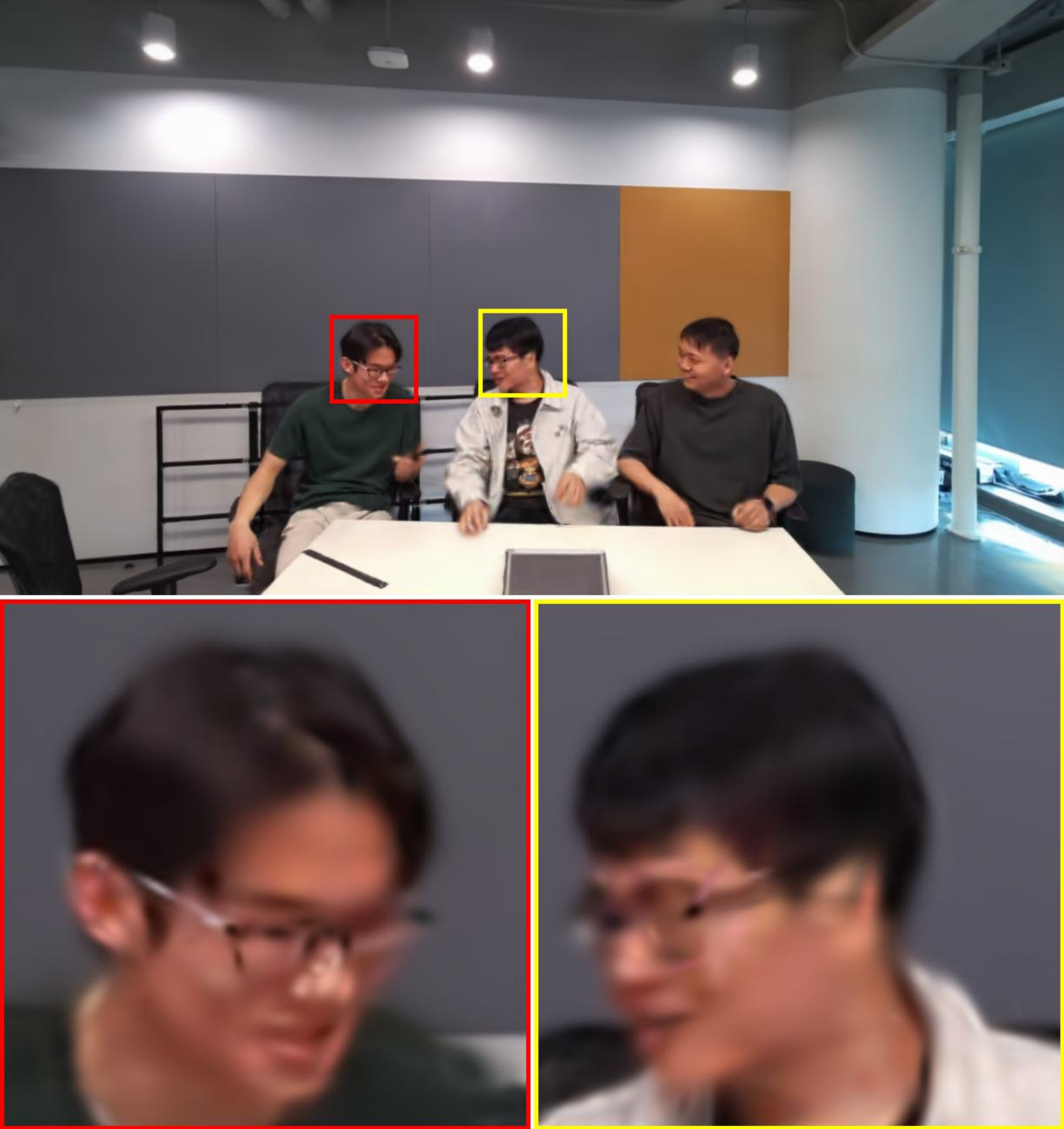}}
    \hfill
    \subfloat{\includegraphics[width=0.19\textwidth]{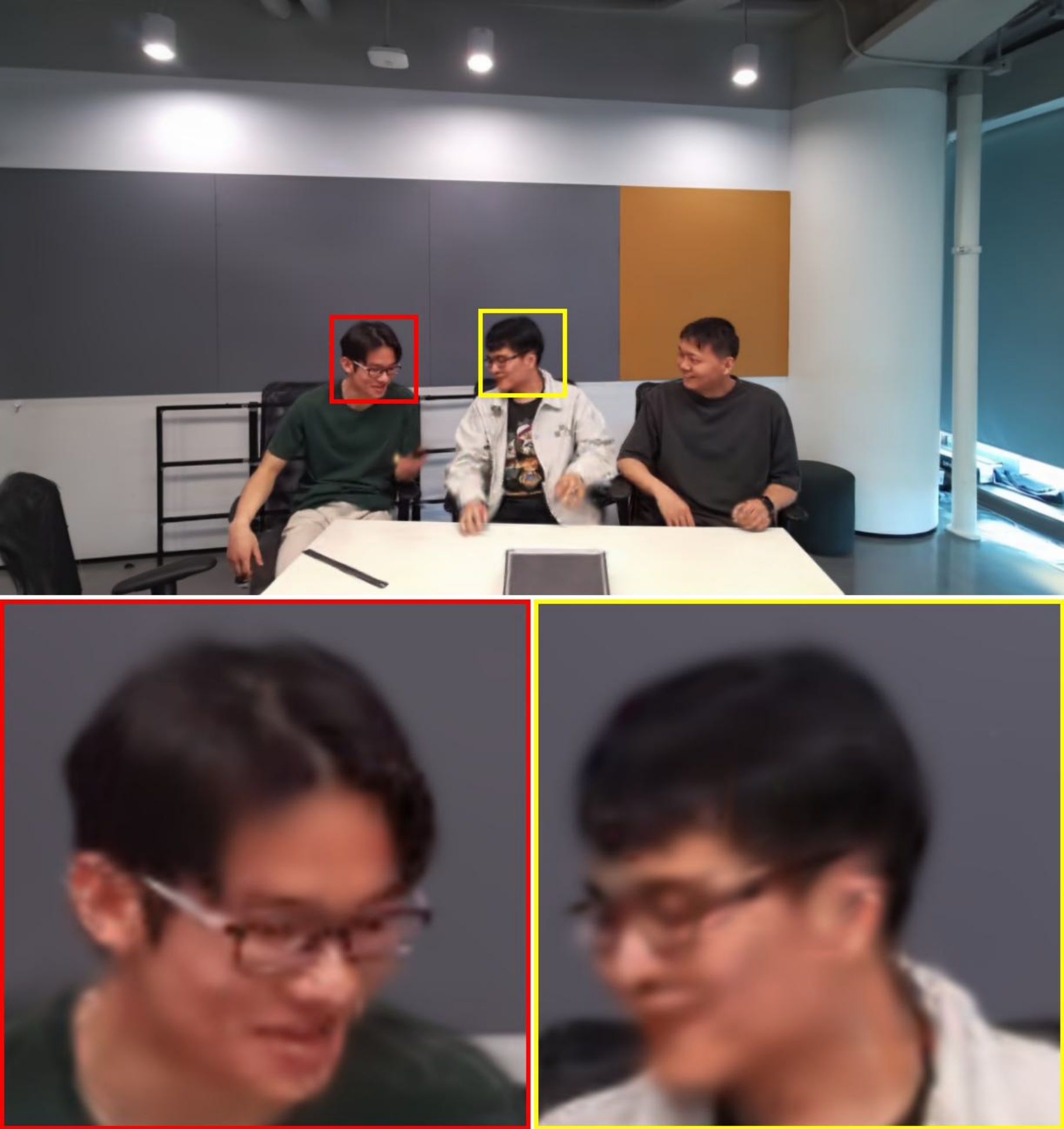}}
    \hfill
    \subfloat{\includegraphics[width=0.19\textwidth]{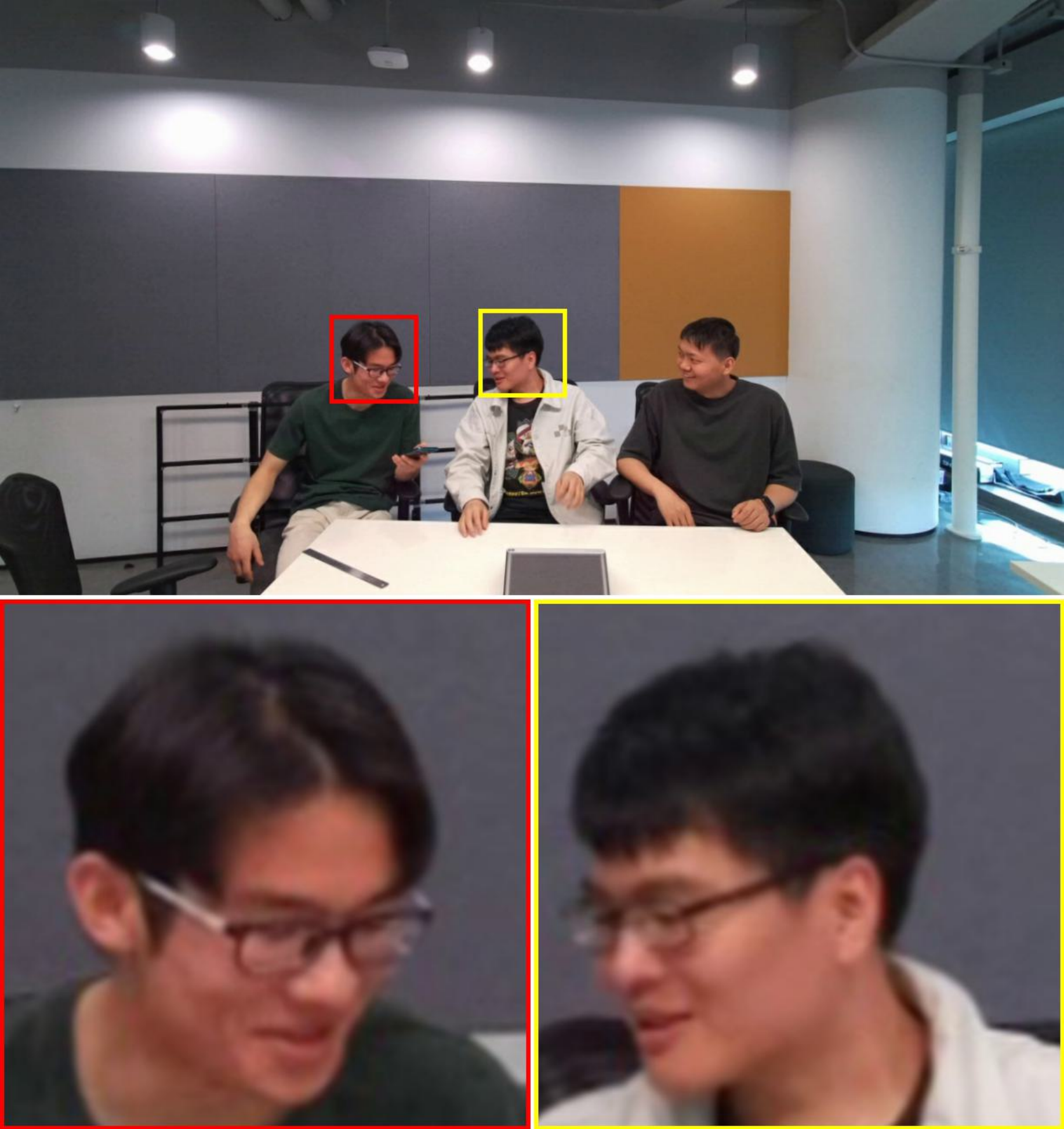}}
    
    \vspace{-0.5em} 

    \subfloat{\includegraphics[width=0.19\textwidth]{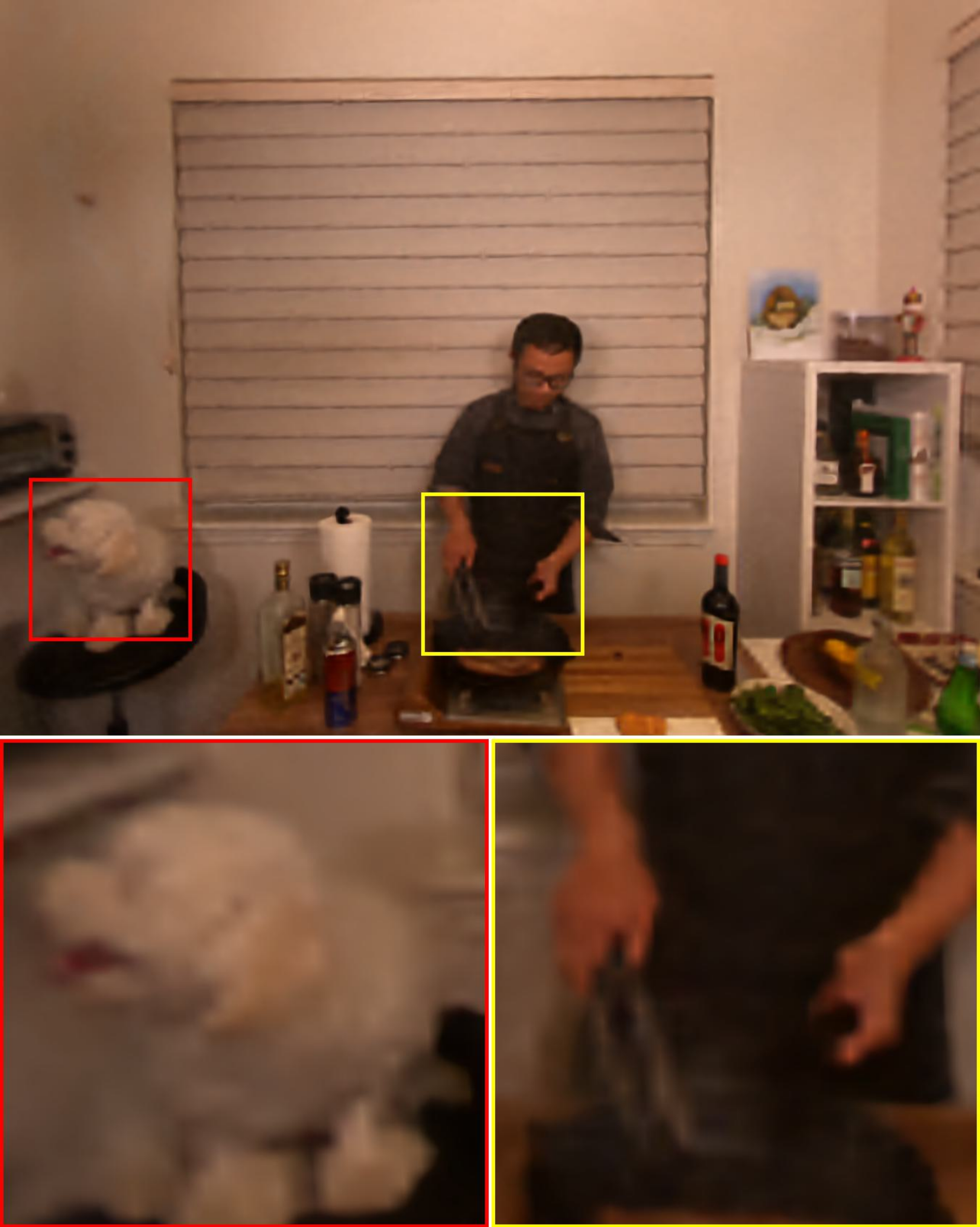}}
    \hfill
    \subfloat{\includegraphics[width=0.19\textwidth]{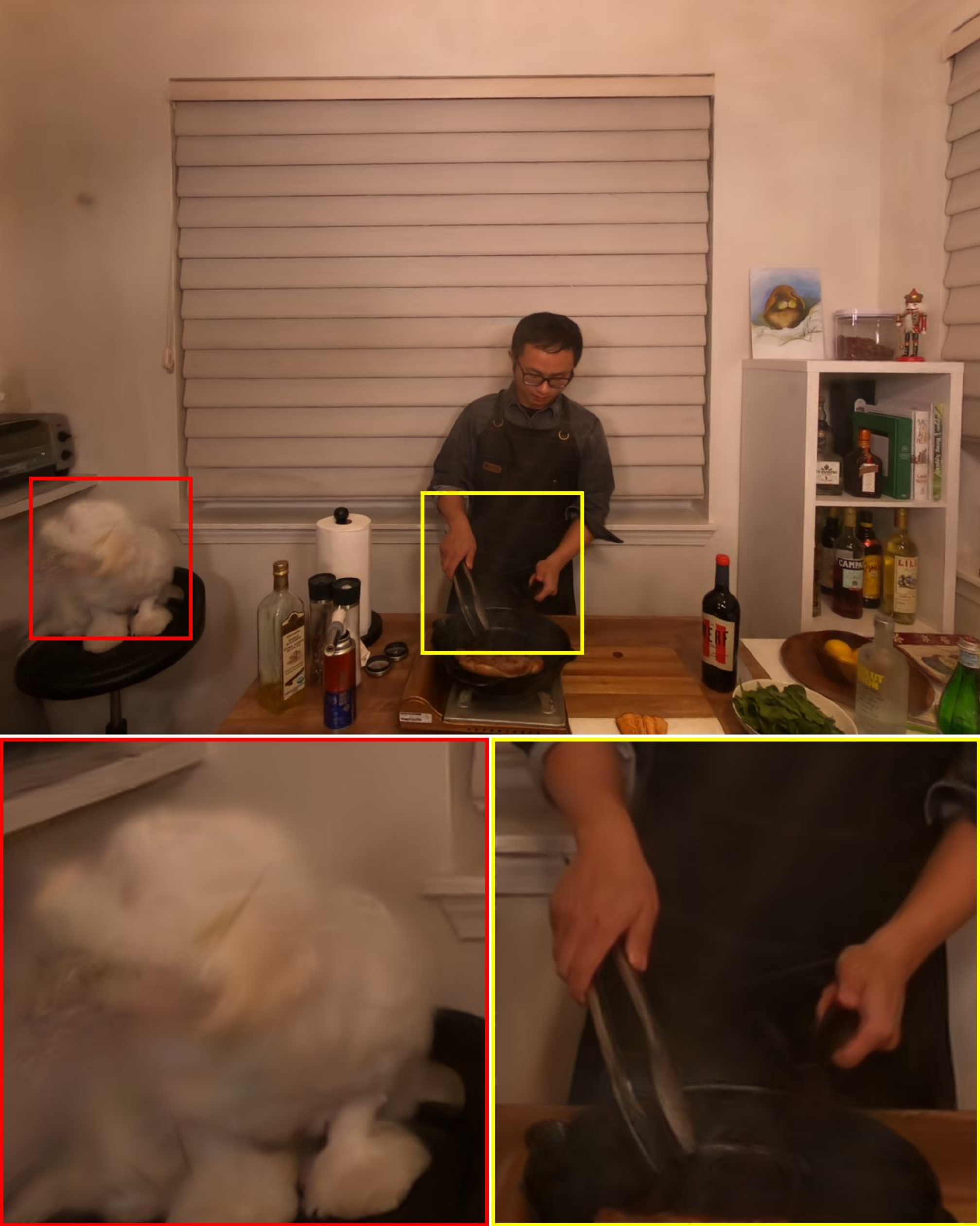}}
    \hfill
    \subfloat{\includegraphics[width=0.19\textwidth]{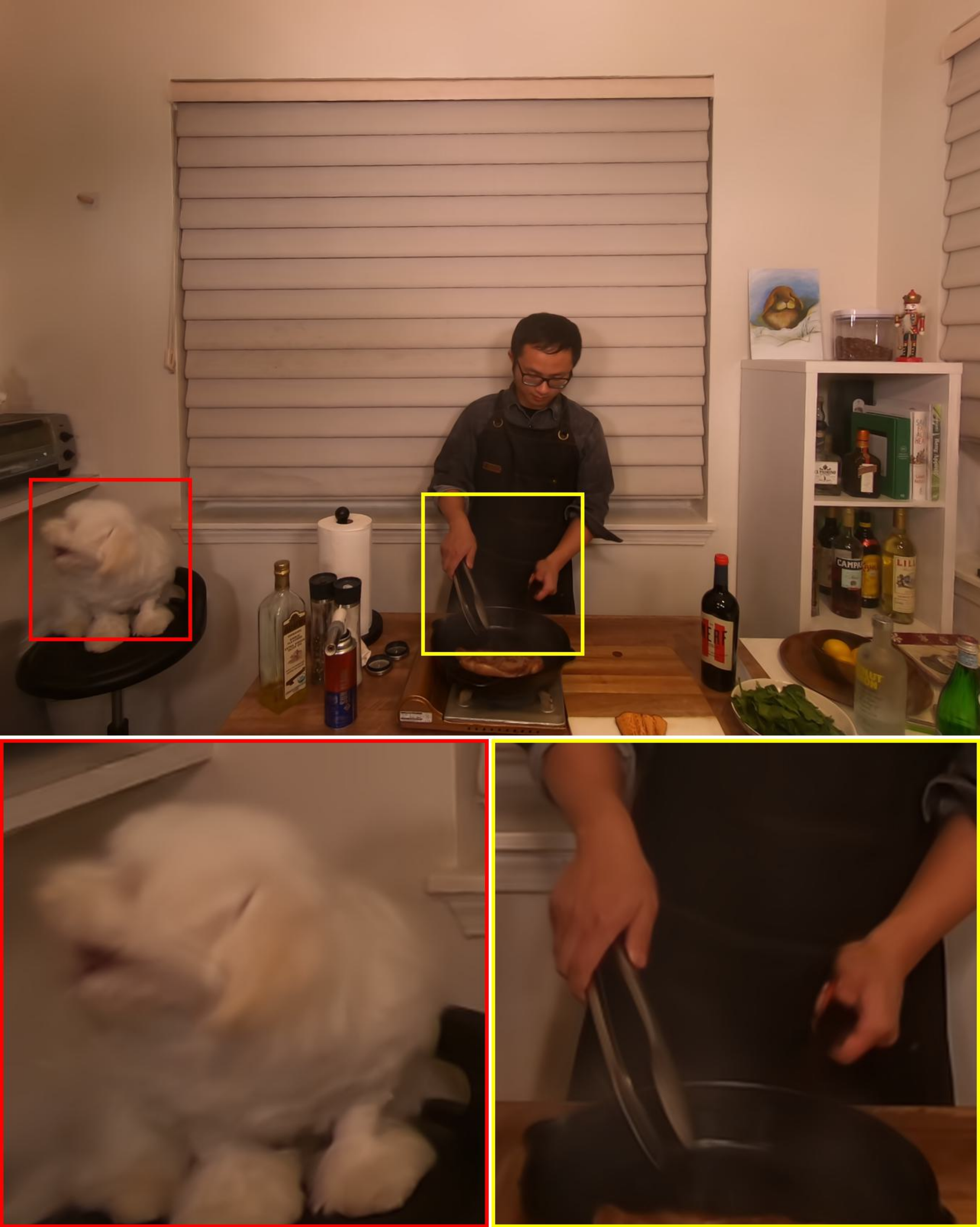}}
    \hfill
    \subfloat{\includegraphics[width=0.19\textwidth]{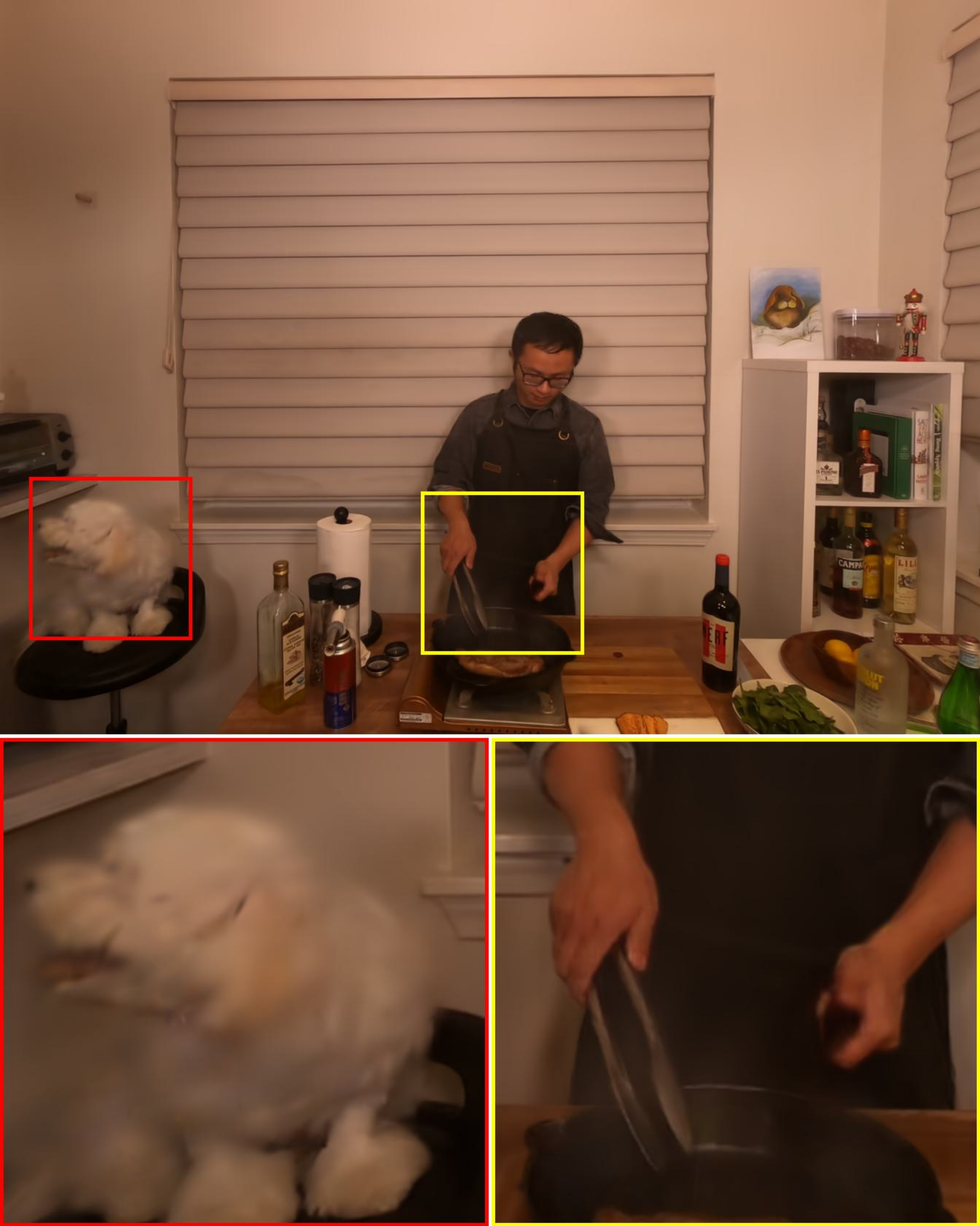}}
    \hfill
    \subfloat{\includegraphics[width=0.19\textwidth]{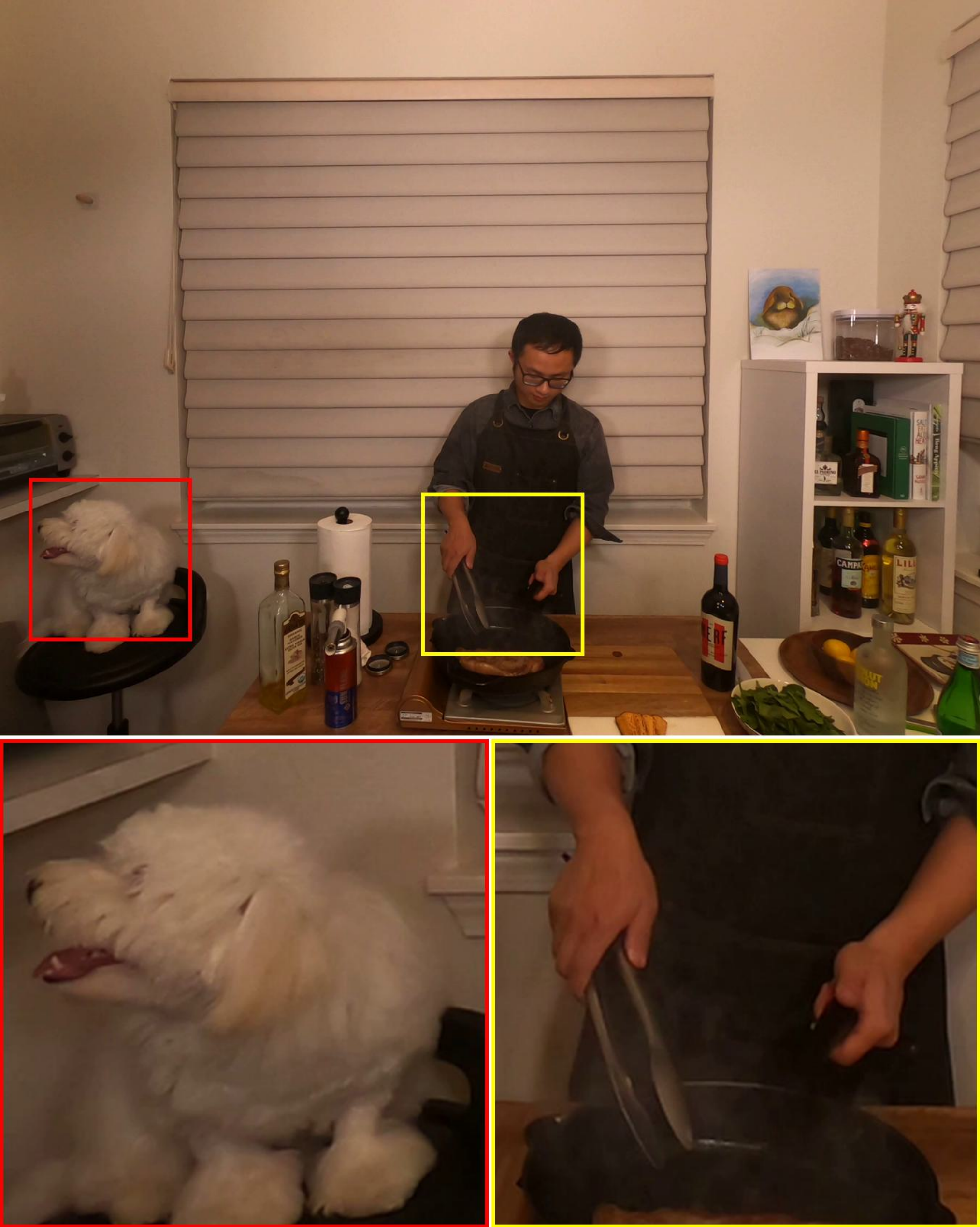}}

    \footnotesize
    \makebox[0.19\textwidth][c]{(a) StreamRF}
    \hfill
    \makebox[0.19\textwidth][c]{(b) 3DGStream}
    \hfill
    \makebox[0.19\textwidth][c]{(c) iFVC}
    \hfill
    \makebox[0.19\textwidth][c]{(d) Ours}
    \hfill
    \makebox[0.19\textwidth][c]{(e) Ground Truth}

    \caption{\textbf{Qualitative comparisons} on the \textit{discussion} scene of the Meet Room dataset and the \textit{sear steak} scene of the N3DV dataset in comparison with online training methods for free-viewpoint video construction.}
    \label{fig:visual_Comparisons}
    \vspace{-2mm}
\end{figure*}
\par
When rendering 2D images, we first need to project these 3DGs onto a 2D plane
and aggregate them using $\alpha$-blending. Specifically, the projected 3DG
are approximated as a 2D Gaussian with a mean point $\mu^{2D}$
and covariance $\Sigma^{2D}$. The computation process is as follows:
\begin{equation}
    \mu^{2D}=(P((W\mu_i)/(W\mu_i)_z))_{1:2},
\end{equation}
\begin{equation}
    \Sigma^{2D}=({JW{\Sigma}W^{T}J^{T}})_{1:2,1:2}.
\end{equation}
Here, $P$ is the projection transformation,
$W$ is the view transformation, and $J$ is the Jacobian of the affine approximation of the projection transformation $P$.
After this, we can represent a 2D Gaussian in image space as $G_{2D}(X;\mu_{2D},\sum_{2D})$.
We perform depth sorting on 2D Gaussians to obtain
$N$ ordered points covering corresponding pixels,
allowing us to render the colors of the pixels:
\begin{equation}
    C=\sum_{i=1}^{N}{c_i\alpha_i'\prod_{j=1}^{i-1}(1-\alpha_j')},
\end{equation}
where $c_i$ represents the view-dependent color obtained
by evaluating the viewing direction
and the coefficients $sh$ of the $i$-th 3DG,
while $\alpha_i'$ is obtained by multiplying the opacity $\alpha_i$ of
the $i$-th 3DG $G$ with its 2DG evaluation.
\par
By optimizing 3DGs' parameters and adaptively adjusting the density of the 3DGs,
equipped with a differentiable rasterizer,
it can render high-quality images in real-time.

\section{Methods}
Struct-GStream supports fast, online construction of photorealistic
free-viewpoint videos at low bitrates from a set of posed 2D images.
Our scene representation consists of two parts:
structured 3DGs (Sec. \textcolor{red}{\ref{sec:structured3dgs}}) and free 3DGs (Sec. \textcolor{red}{\ref{sec:free3dgs}}).
We introduce anchor points into dynamic scene
representation to generate structured 3DGs which model the foundational structure of a scene and fit rough movements.
Meanwhile, we adopt a global free 3DGs patching strategy
that generates free 3DGs
in regions where the scene reconstruction quality is deficient,
which, unlike the structured 3DGs, are not anchored
and can independently address deficient reconstructions. Fig. \textcolor{red}{\ref{fig:overview}} is an overview of our method.
\subsection{Structured 3DGs}
\label{sec:structured3dgs}
Movements in the real world typically exhibit spatial continuity and local rigidity,
occurring less frequently than stationary areas. Consequently, the current online FVV construction
methods~\cite{dynamic3dg,3dgstream} based on 3DG-S~\cite{3DGS}, face redundancy in storage and training
time due to their uniform and indiscriminate use of 3DGs for all scene content.
To address this problem, we incorporate dynamic anchor points into the representation of dynamic scenes
to generate movable structured 3DGs.
These structured 3DGs, which are used to express the coarse-grained basic structures
of dynamic scenes, can substantially reduce storage and expedite
the convergence speed during training.
\par
Structured 3DGs are an important representation of our dynamic scene.
We need to be able to use it to represent the fundamental structure of the scene
as well as roughly fit the changes occurring in the scene.
We choose to generate structured 3DGs using dynamic anchor points.
While achieving high-quality scene reconstruction, dynamic anchor points,
the coarse-grained scene representation meets our requirements
for low storage and fast convergence in streaming free-viewpoint synthesis.
At frame 0, through static reconstruction method Scaffold-GS~\cite{scaffoldgs}, we reconstruct structured 3DGs of the scene and
obtain the following structures through the reconstruction process:
a set of dynamic anchor points $V$, each of which has $k$ structured 3DGs attached to it,
and three trained MLPs $F_{\alpha}$, $F_{c}$, $F_{\Sigma}$ used to generate the opacity, color,
and covariance matrix of structured 3DGs attached to dynamic anchor points.
Unlike Scaffold-GS, we do not fix anchor points.
Instead, our anchor points have variable positions.
At the same time, we fix the number of dynamic anchor points,
which in turn fixes the number of structured 3DGs.
$F_{\Sigma}$ is inherited and optimized during the training process in each frame
whereas the other two MLPs are utilized without undergoing further optimization.
\par
For each dynamic anchor point, denoted as $v$, we have the following attributes:
a learnable position $x_v$, a learnable scaling factor $l_v\in{R^{3}}$,
structured 3DGs' offsets $O_{v}\in{R^{k*3}}$ attached to the dynamic anchor point
and a local contextual feature $f_v\in{R^{32}}$,
where $O_v$ and $f_v$ are fixed to their values at frame 0.
Structured 3DGs possess the same attributes as regular 3DGs and all attributes are obtained through
dynamic anchor points and MLPs.
\par
At each frame, we optimize the attributes of the dynamic anchor points to obtain
new positions $x_v'$ and scaling factors $l_v'$.
With these attributes, we can derive the new position $\mu'$ of the structured 3DG
attached to anchor point $v$ using structured position:
\begin{equation}
    \{\mu_0',\cdots,\mu_{k-1}'\}=x_v'+\{O_0,\cdots,O_{k-1}\}\cdot l_v',
\end{equation}
where the set $\{O_1,...,O_{k-1}\}\in{R^{k*3}}$ represents the
offsets of $k$ structured 3DGs
relative to the dynamic anchor point.
As for other attributes of structured 3DG,
the new covariance matrix is obtained
through the current frame's optimized MLP $F_\Sigma'$:
\begin{equation}
    \{\Sigma_0',\cdots,{\Sigma_{k-1}}'\}=F_\Sigma'(\vec{f_v},\delta_{vc},\vec{d_{vc}}).
\end{equation}
Here $\vec{f_v}$ is the feature vector of the dynamic anchor point,
$\delta_{vc}$ is the distance from the dynamic anchor point to the observing camera,
and $\vec{d_{vc}}$ is the direction from the dynamic anchor point to
the observing camera.
Structured 3DGs' color $c_i$ and opacity $\alpha_i$ are obtained from the initial MLP $F_{\alpha}$, $F_{c}$ with the same inputs as those used for $F_\Sigma'$.
\par
During the online training process, for structured 3DGs, we optimize the position $x_v$
and scaling factor $l_v$ of dynamic anchor points and MLP $F_{\Sigma}$ to adjust their position $\mu$, scaling $s$,
and the quaternion $q$ representing rotation.
Dynamic anchor points and MLP $F_{\Sigma}$ are optimized by the loss combined $L_1$, a SSIM term $L_{SSIM}$ and a volume regularization term $L_{vol}$:
\begin{equation}
    \label{equ:loss}
    L=L_1+\lambda_{SSIM}L_{SSIM}+\lambda_{vol}L_{vol},
\end{equation}
where $L_{vol}$ is defined as follows:
\begin{equation}
    L_{vol} = \sum_{i=1}^N{Prod}(s_i).
\end{equation}
Here $N$ represents the number of all 3DGs rendered in the scene,
and Prod(·) denotes the product of a vector of values,
with $s_i$ representing the scaling of the $i$-th 3DG.
The volume regularization term $L_{\text{vol}}$ is borrowed from Scaffold-GS~\cite{scaffoldgs}, which reduces the overlap of 3DGs by limiting scaling.
\par
During the initialization process of $0$-th frame,
we propose a distance regularization term to make the structured 3DGs' distribution more compact
and better suited for dynamic scenes. The loss function $L_{init}$ in the initialization process takes the following form:
\begin{equation}
    L_{init}=L_1+\lambda_{SSIM}L_{SSIM}+\lambda_{vol}L_{vol}+\lambda_{dist}L_{dist}.
\end{equation}
Here, $\lambda_{SSIM}$, $\lambda_{vol}$, $\lambda_{dist}$ are 0.2, 0.01, 0.005 respectively and the distance regularization term $L_{dist}$ is:
\begin{equation}
    L_{dist}=\sum_{i=1}^{N}\sum_{j=1}^{k}{{\theta(O_{i,j}-2\epsilon)}||O_{i,j}-2\epsilon|| },
\end{equation}
where $N$ is the number of anchor points and $k$ is the number of each anchor point's structured 3DGs.
$\epsilon$ represents the voxel size during the initialization of anchor points. $O_{i,j}$ represents the offset of
the $i$-th anchor point's $j$-th structured 3DGs and $\theta$ is a discriminator that outputs 0 when the input is less than or equal to 0, and outputs 1 otherwise. The distance regularization term $L_{dist}$ is applied exclusively during the initialization at frame 0. During the online training process, we simply removed the distance regularization term from the loss function. It aims to provide a compact initialization by constraining the offsets $O_v$ of structured 3DGs relative to their anchors, preventing large deviations that could lead to artifacts during subsequent streaming. In the online reconstruction phase, we only optimize the anchor attributes (position $x_v$, scaling factor $l_v$) and the covariance MLP $F_{\Sigma}$, while the offsets $O_v$ remain fixed. Since $O_v$ is not updated during streaming, the distance regularization term is no longer necessary and is therefore removed.

\begin{table}
  \centering
  \caption{\textbf{Quantitative comparison} on the N3DV dataset.
    The training time, storage, rendering speed, and PSNR are averaged over the whole 300 frames for each scene.
    $^\dagger$DyNeRF only report metrics on the \textit{flame salmon} scene.
    $^\star$Considering the initial model. Colors \coloredcircle{first},
    \coloredcircle{second}, \coloredcircle{third} mark the top three performing methods under each metric. To ensure fairness,
    we did not consider static reconstruction methods.
    We highlighted the top-performing online method with underlined formatting in the storage metric. The FPS of 3DGStream has been retested in our experimental environment.
  }
  \label{tab:N3DV_Comparisons_Avg}
  \resizebox{\columnwidth}{!}{%
    \begin{tabular}{@{}c|l|cccc|c@{}}
      \toprule
      \multirow{2}{*}{Category} & \multirow{2}{*}{Method} & PSNR$\uparrow$          & Storage$\downarrow$         & Train$\downarrow$      & Render$\uparrow$      & \multirow{2}{*}{Streamable} \\
                                &                         & (dB)                    & (MB)                        & (mins)                 & (FPS)                 &                             \\
      \midrule
      \multirow{3}{*}{Static}   & Plenoxels               & 30.77                   & 4106                        & 23                     & 8.3                   & $\checkmark$                \\
                                & I-NGP                   & 28.62                   & 48.2                        & 1.3                    & 2.9                   & $\checkmark$                \\
                                & 3DG-S                   & 32.08                   & 47.1                        & 8.3                    & 390                   & $\checkmark$                \\
                                & Scaffold-GS             & 32.25                   & 24.5                        & 1.9                    & 90                    & $\checkmark$                \\
      \midrule
      \multirow{6}{*}{Offline}  & DyNeRF                  & 29.58$^\dagger$         & 0.1                         & 260                    & 0.02                  & $\times$                    \\
                                & NeRFPlayer              & 30.69                   & 17.1                        & 1.2                    & 0.05                  & $\checkmark$                \\
                                & HexPlane                & 31.70 & 0.8                         & 2.4                    & 0.21                  & $\times$                    \\
                                & K-Planes                & 31.63                   & 1.0                         & 0.8                    & 0.9                  & $\times$                    \\
                                & HyperReel               & 31.10                   & 1.2                         & 1.8                    & 2.00                  & $\times$                    \\
                                & MixVoxels               & 30.80                   & 1.7                         & 0.27                   & 16.7                  & $\times$                    \\
                                & RealTime4DGS            & 29.95                   & 17.7                        & 1.6                    & 73                    & $\times$                    \\
                                & 4DGaussians            & 30.88                   & 0.21                        & \cellcolor{third}0.22                    & 78                    & $\times$                    \\
                                & SpacetimeGaussians      & \cellcolor{first}32.05  & 0.7                         & 0.90 & \cellcolor{second}140 & $\times$                    \\
      \midrule
      \multirow{2}{*}{Online}   & StreamRF                & 30.68                   & 17.7/31.4$^\star$           & 0.25                   & 8.3                   & $\checkmark$                \\
                                & iFVC               & \cellcolor{second}31.88                   & \underline{0.12/0.14$^\star$}             & 0.24  & 70  & $\checkmark$                \\
                                & 3DGStream               & 30.97                   & 7.6/7.8$^\star$             & \cellcolor{second}0.20  & \cellcolor{first}160  & $\checkmark$                \\
                                & Ours                    & 31.72\cellcolor{third}  & 4.7/4.8$^\star$ & \cellcolor{first}0.14  & 120\cellcolor{third}  & $\checkmark$                \\
      \bottomrule
    \end{tabular}
  }
\end{table}

\begin{figure}[!t]
    \centering
    \includegraphics[width=0.5\textwidth]{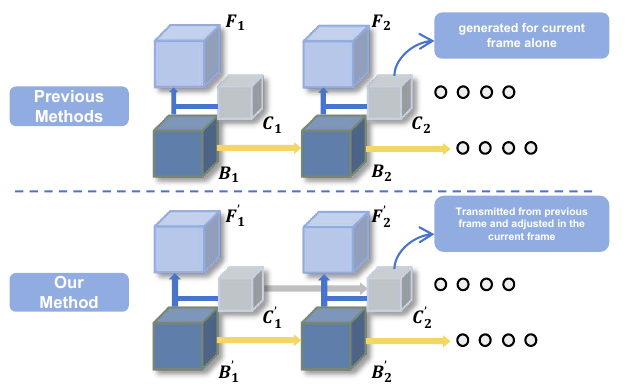}
    \caption{Overview of previous online reconstruction methods' workflow compared to ours.
        $B_i$, $C_i$, and $F_i$ represent the base scene representation, inter-frame compensation, and complete scene representation of the i-th frame, respectively.}
    \label{fig:method_comparison}
    \vspace{-2mm}
\end{figure}

\begin{figure*}[t]
    \centering
    \subfloat{\includegraphics[width=0.24\textwidth]{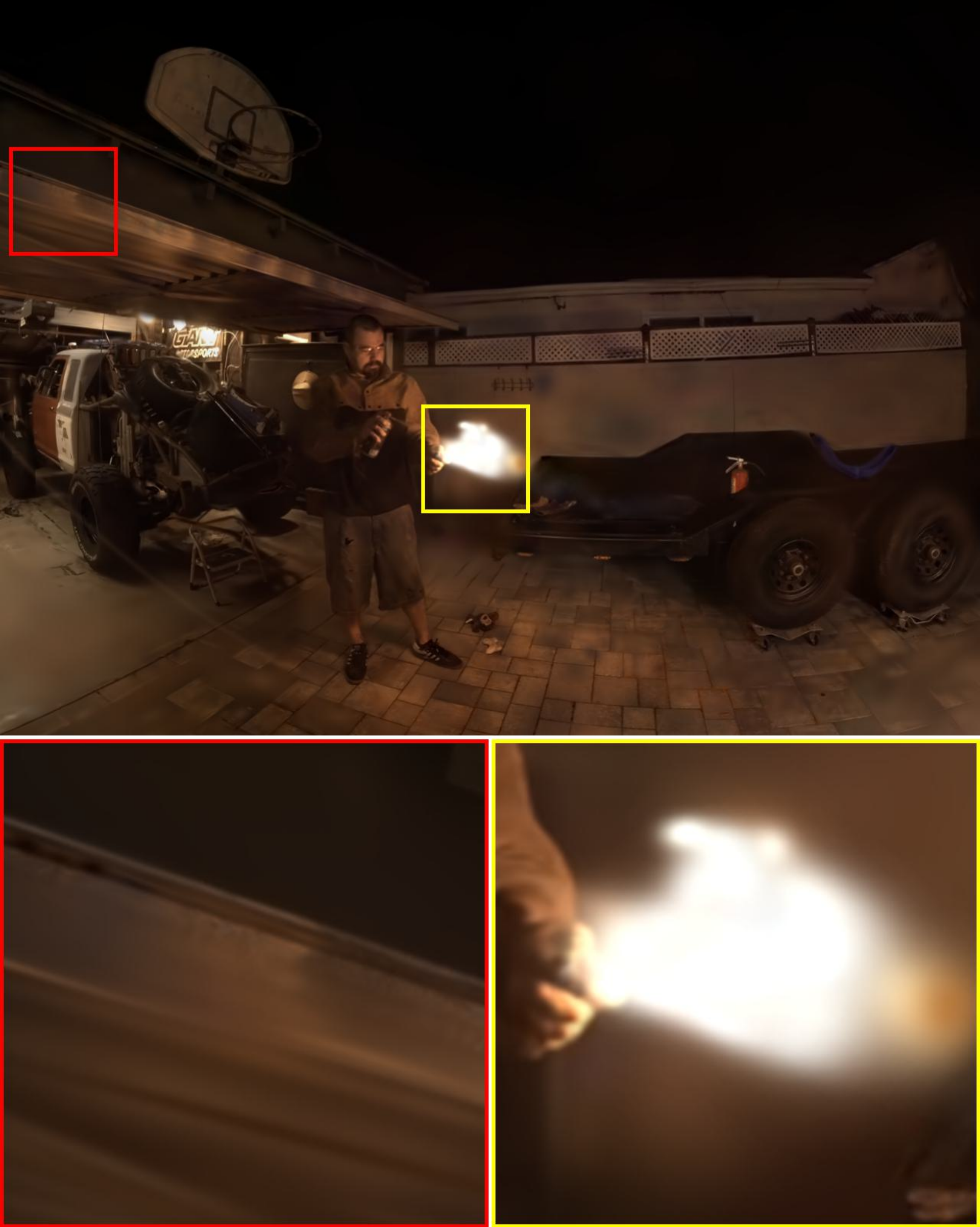}}
    \hfill
    \subfloat{\includegraphics[width=0.24\textwidth]{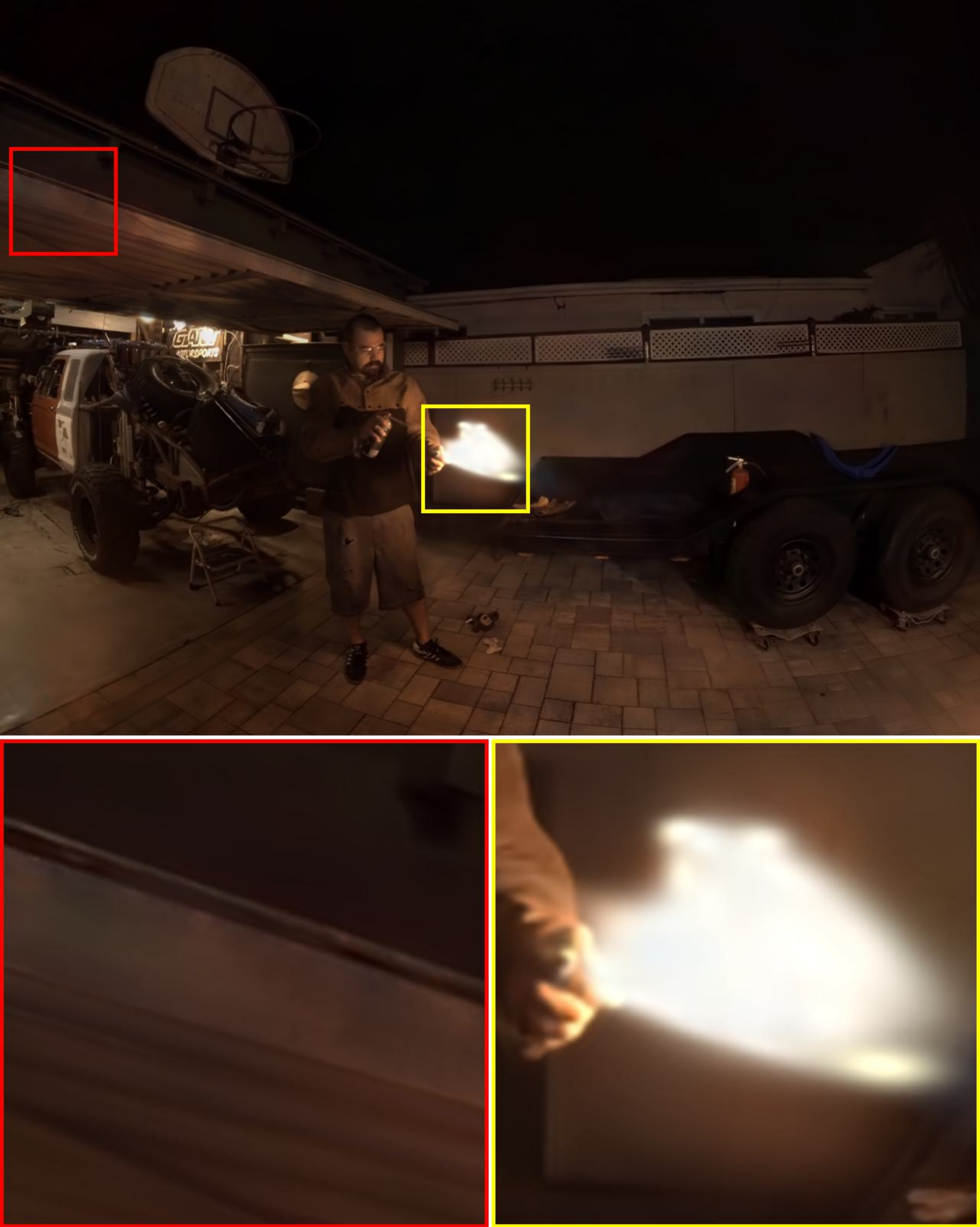}}
    \hfill
    \subfloat{\includegraphics[width=0.24\textwidth]{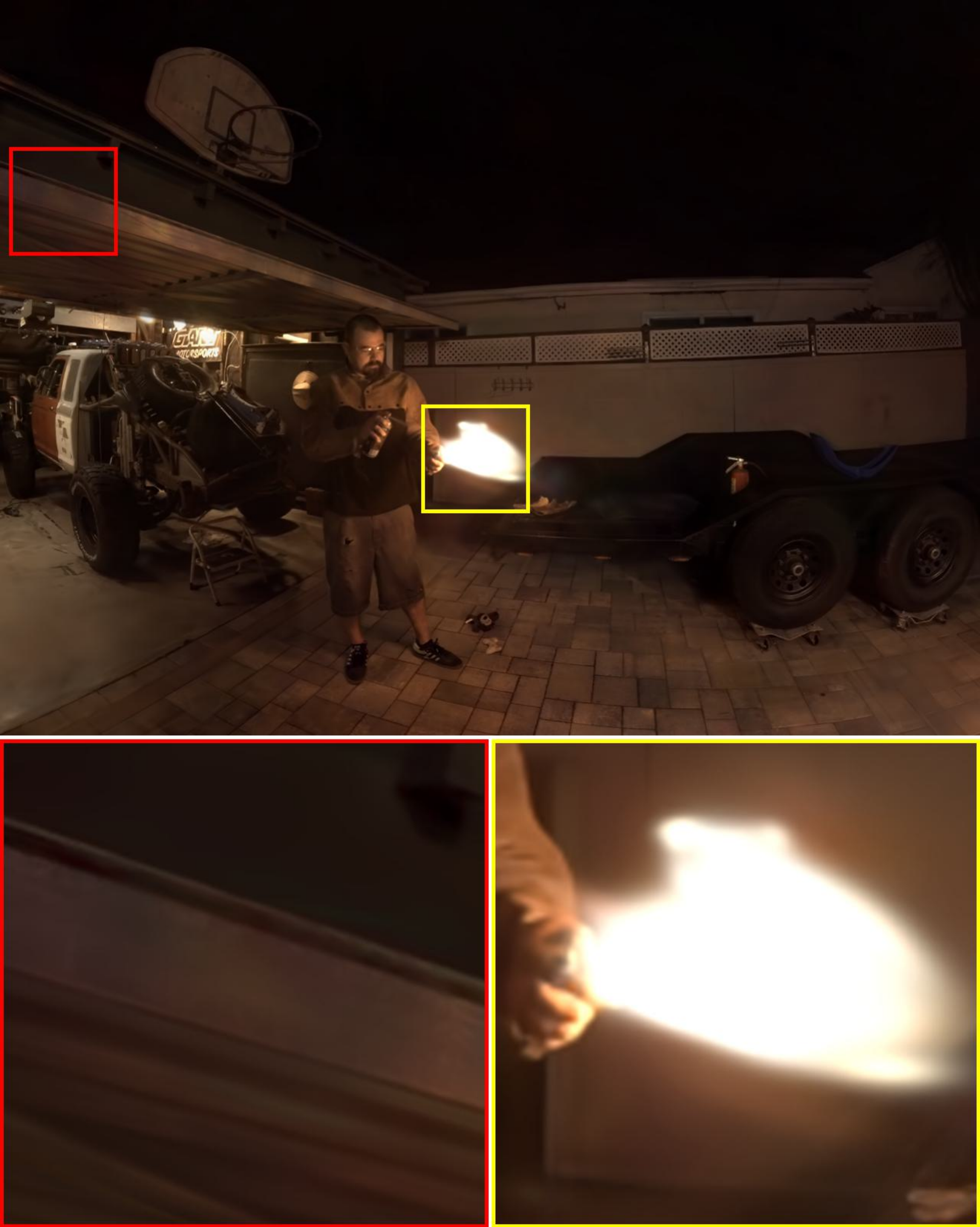}}
    \hfill
    \subfloat{\includegraphics[width=0.24\textwidth]{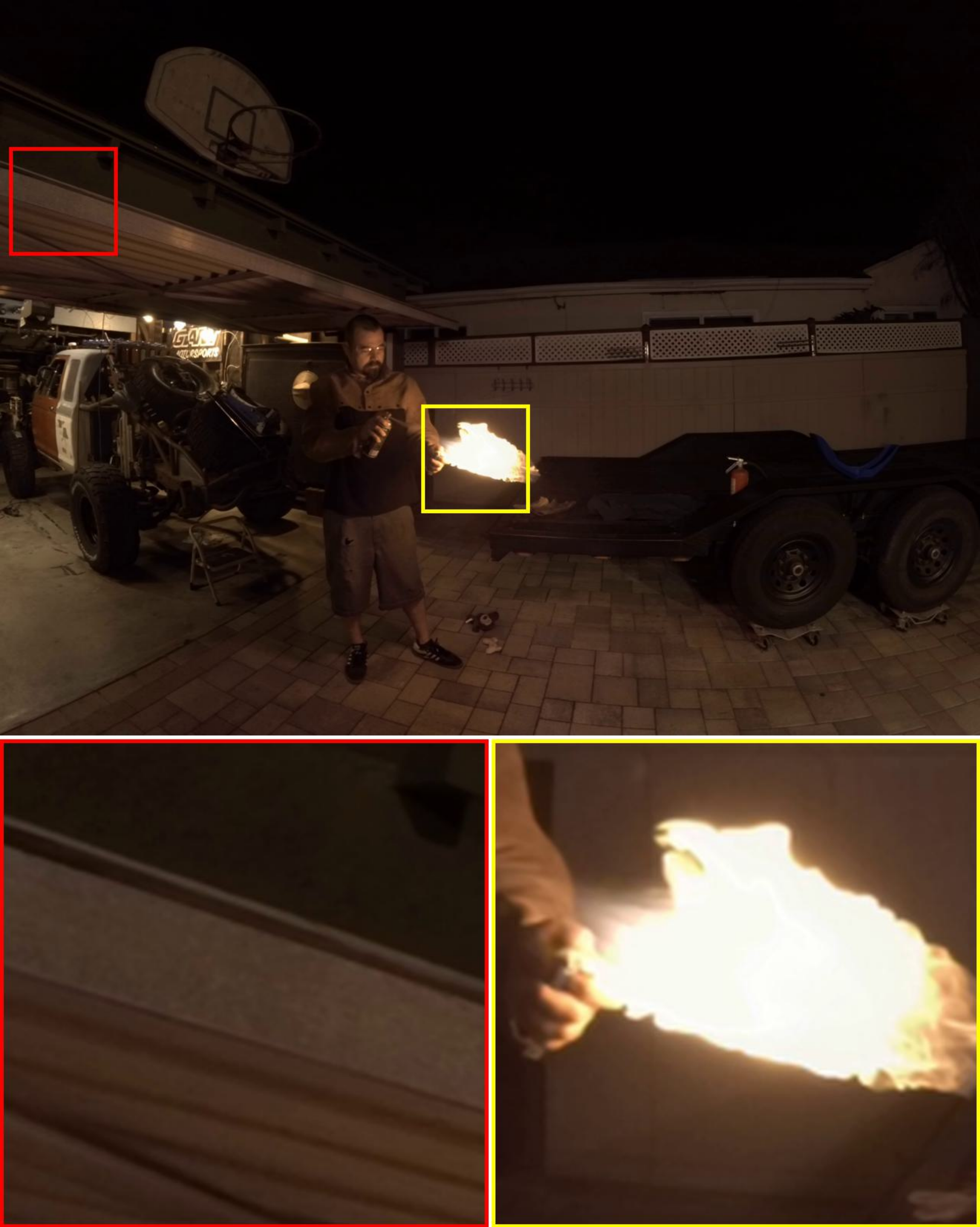}}
    
    \footnotesize
    \makebox[0.24\textwidth][c]{(a) 3DGStream}
    \hfill
    \makebox[0.24\textwidth][c]{(b) iFVC}
    \hfill
    \makebox[0.24\textwidth][c]{(c) Ours}
    \hfill
    \makebox[0.24\textwidth][c]{(d) Ground Truth}

    \caption{\textbf{Qualitative comparisons} on the \textit{02\_Flames} scene of Google Immersive Dataset.}
    \label{fig:immersive_vis}
    \vspace{-3mm}
\end{figure*}
\subsection{Global Free 3DGs Patching Strategy}
\label{sec:free3dgs}
By adjusting the structured 3DGs, we can roughly compensate for the inter-frame differences in dynamic scenes and
manage most relatively simple movements.
However, the reconstruction quality of structured 3DGs is limited
by the assumption of local rigidity,
and relying solely on it for modeling can make handling complex movements
in the scene quite challenging,
such as the movement of clothes,
or the emergence of new objects like poured coffee.
Considering that structured 3DGs are dependent on dynamic anchor points,
and these dynamic anchor points are relatively coarse scene representations compared to the regular 3DGs,
complex scene movements inevitably lead to difficulties in optimization
and result in blurriness in certain areas.
Additionally, when dealing with emerging objects in a scene,
such as flames from a flamethrower, while we maintain a fixed number of dynamic anchor points,
we can only reduce loss by optimizing to locally blur the appearance of the newly emerging object.
\par
For this reason, we choose to directly generate free 3DGs which do not rely on dynamic anchor points in the scene.
A critical question that arises is where we should generate these free 3DGs.
Through a series of experiments and observations,
we found that hard-to-capture motion and emerging objects, while causing poor reconstruction effects,
also introduce large view space positional gradients.
This is because when a region experiences overly complex motion or the introduction of a new object,
optimization aims to adjust the positions of the surrounding 3DGs to achieve better effects in complex motion areas
or move the 3DGs of surrounding objects near the emerging object to create it.
Based on this discovery, we choose to generate free 3DGs near all 3DGs
with large view space positional gradient. 
\par
For the specific free 3DGs generation process,
we set a relatively large view space positional gradient threshold of $\tau_{grad} = 0.001$,
which ensures the accurate generation of free 3DGs
in areas with complex motion or emerging objects,
while ignoring static regions where reconstruction is already satisfactory.
For any 3DG $G$ which can be either a structured or free 3DG
with its positional gradient exceeding $\tau_{grad}$,
we sample a free 3DG using $X \sim \mathcal{N}(\mu, 2\Sigma)$,
where $\mu$ is the mean of the selected 3DG $G$ and $\Sigma$ is its covariance matrix.
Attributes of the newly generated free 3DG are derived from the original 3DG.
The color and scale attributes of the sampled free 3DG are initialized using the same value with $G$,
while the rotation is set to the unit quaternion,
and the opacity is initialized to 0.1.
\begin{table}
    \caption{\textbf{Quantitative comparison} on the Meet Room dataset.
        Note that the training time, required storage and PSNR are averaged over the whole 300 frames.
        $^\star$Considering the initial model.}
    \label{tab:meetroom_Comparisons}
    \centering
    \begin{tabular}{@{}l|cccc@{}}
        \toprule
        \multirow{2}{*}{Method} & PSNR$\uparrow$          & Storage$\downarrow$                & Train$\downarrow$      & Render$\uparrow$      \\
                                & (dB)                    & (MB)                               & (mins)                 & (FPS)                 \\
        \midrule
        Plenoxels               & 27.15                   & 1015                               & 14                     & 10                    \\
        I-NGP                   & 28.10                   & 48.2                               & 1.1                    & 4.1                   \\
        3DG-S                   & \cellcolor{second}31.31 & 21.1                               & 2.6                    & \cellcolor{first}571  \\
        Scaffold-GS             & \cellcolor{first}31.52  & 17.7                               & 2.8                    & 108                   \\
        \midrule
        StreamRF                & 26.72                   & 5.7/9.0$^\star$   & 0.17  & 10                    \\
        iFVC               & \cellcolor{third}31.16                   & \underline{0.09/0.10$^\star$}  & \cellcolor{third}0.15 & 74 \\
        3DGStream               & 30.45                   & 4.0/4.1$^\star$  & \cellcolor{second}0.10 & \cellcolor{second}197 \\
        Ours                    & 30.95  & 2.97/3.02$^\star$ & \cellcolor{first}0.08  & \cellcolor{third}123  \\

        \bottomrule
    \end{tabular}
\end{table}
\begin{table}
    \caption{\textbf{Quantitative comparison} on the Google Immersive dataset.
        $^\star$Considering the initial model.}
    \label{tab:immersive_Comparisons}
    \centering
    \begin{tabular}{@{}l|cccc@{}}
        \toprule
        \multirow{2}{*}{Method} & PSNR$\uparrow$          & Storage$\downarrow$                & Train$\downarrow$      & Render$\uparrow$      \\
                                & (dB)                    & (MB)                               & (mins)                 & (FPS)                 \\
        \midrule
        iFVC & \cellcolor{first}28.65   & \underline{0.14/0.16$^\star$}  & \cellcolor{third}0.70 & \cellcolor{third}56 \\
        3DGStream               & \cellcolor{third}27.69                   & 8.5/8.7$^\star$  & \cellcolor{second}0.41 & \cellcolor{first}135 \\
        Ours                    &\cellcolor{second}28.48   & 5.2/5.4$^\star$ &\cellcolor{first}0.26   & \cellcolor{second}96  \\

        \bottomrule
    \end{tabular}
         \vspace{-3mm}
\end{table}
\par
The next issue to be addressed is how to handle the generated free 3DGs.
Initially inspired by previous online training methods~\cite{ReRF,3dgstream}, we began by independently generating free 3DGs for each frame.
However, through experimentation,
we found that because these free 3DGs are only used for the current frame and are discarded for the next frame,
next frame's free 3DGs are re-generated based on the positional gradients of structured 3DGs. Consequently,
the free 3DGs generated between the two frames are different to some extent,
leading to poor temporal consistency.
Additionally, for some challenging emerging objects,
such as a dog's protruding tongue,
this approach fails to yield good results due to
insufficient frame-by-frame compensation.
Based on these observations,
we choose not to process free 3DGs for each frame independently,
but instead, combine the free 3DGs of the current frame with the structured 3DGs and jointly optimize them,
while also passing free 3DGs between adjacent frames.
This approach effectively maintains reconstruction quality and temporal consistency for emerging objects. In Fig. \textcolor{red}{\ref{fig:method_comparison}}, we roughly depict the general process differences between our method and previous methods.
\par
Additionally, we prune free 3DGs by their opacities to limit them to a reasonable quantity.
Specifically, we have set an opacity threshold of $\tau_\alpha = 0.005$ and during the middle of the training phase,
we prune away free 3DGs with opacities below this threshold.

\begin{table}
    \caption{\textbf{Ablation study of structured 3DGs} for the \textit{sear steak} scene.
        Note that our full model doesn't contain the optimization of opacity and color MLP.}
    \label{tab:rendering}
    \centering
    \begin{tabular}{@{}l|cc@{}}
        \toprule
        Variant                                 & PSNR  \\
        \midrule
        \phantom{} \textit{w/o} Scaling Factor  & 33.53 \\
        \phantom{} \textit{w/o} Convariance MLP & 33.68 \\
        \phantom{} \textit{w} Opacity MLP       & 33.78 \\
        \phantom{} \textit{w} Color MLP         & 33.41 \\
        \midrule
        Full Model                              & 33.81 \\
        \bottomrule
    \end{tabular}
    
  \vspace{-2mm}
\end{table}
\begin{figure}[!t]
  \centering
  \subfloat[w/o Scaling Factor]{\includegraphics[width=0.23\textwidth]{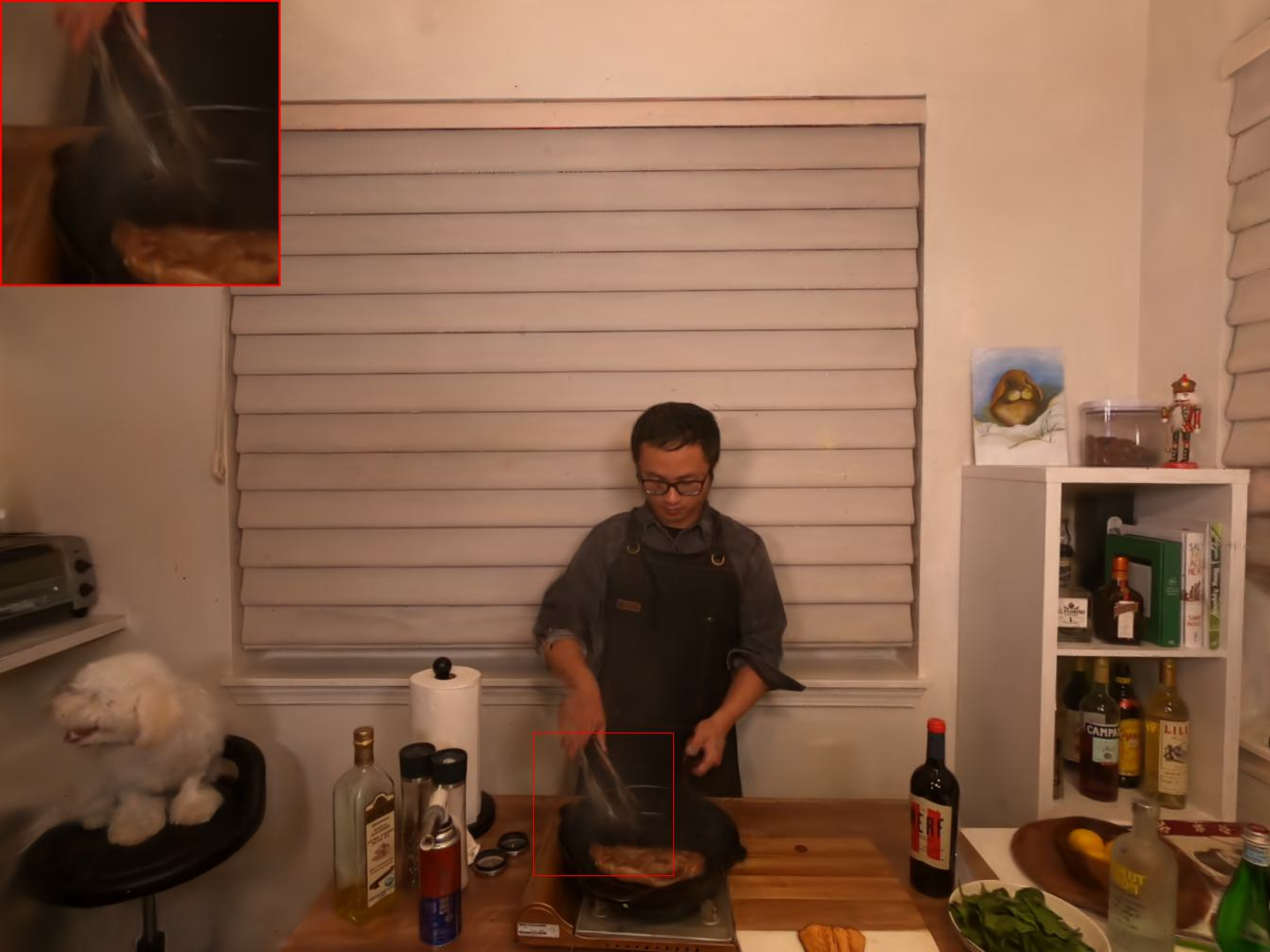}}
  \hfill
  \subfloat[w/o MLP $F_{v}$]{\includegraphics[width=0.23\textwidth]{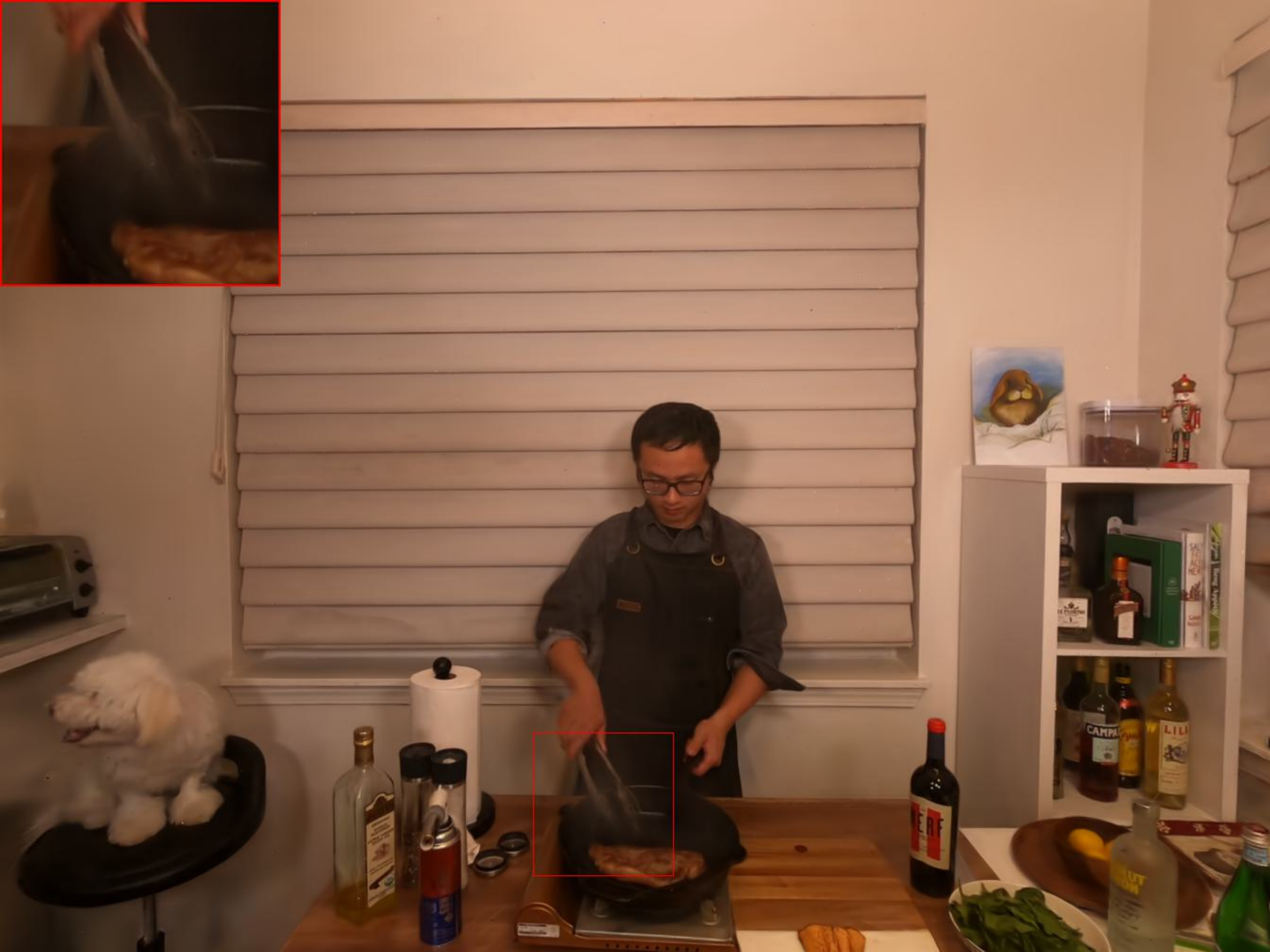}}

  \vspace{0.05cm}

  \subfloat[Full Model]{\includegraphics[width=0.23\textwidth]{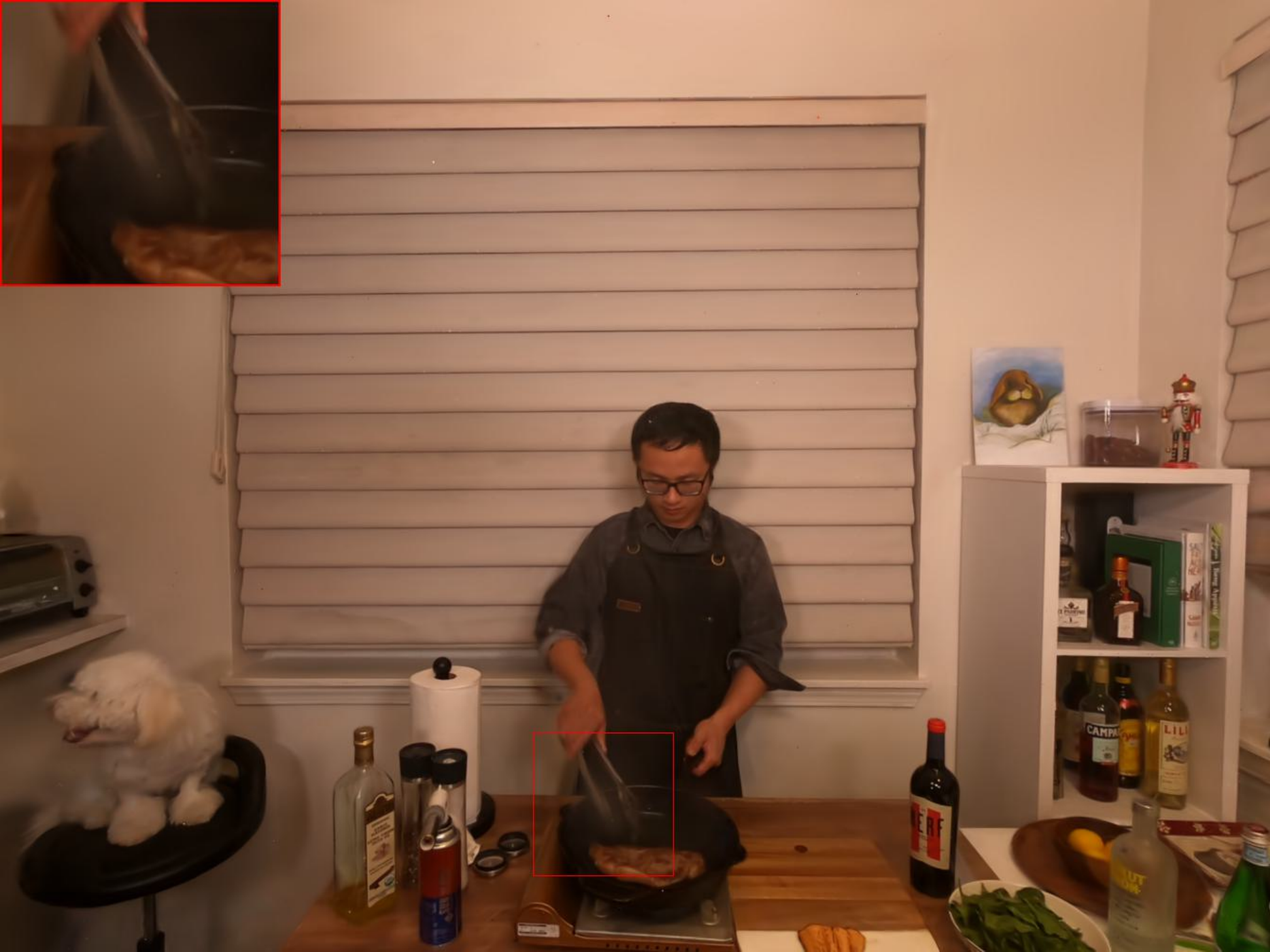}}
  \hfill
  \subfloat[Ground Truth]{\includegraphics[width=0.23\textwidth]{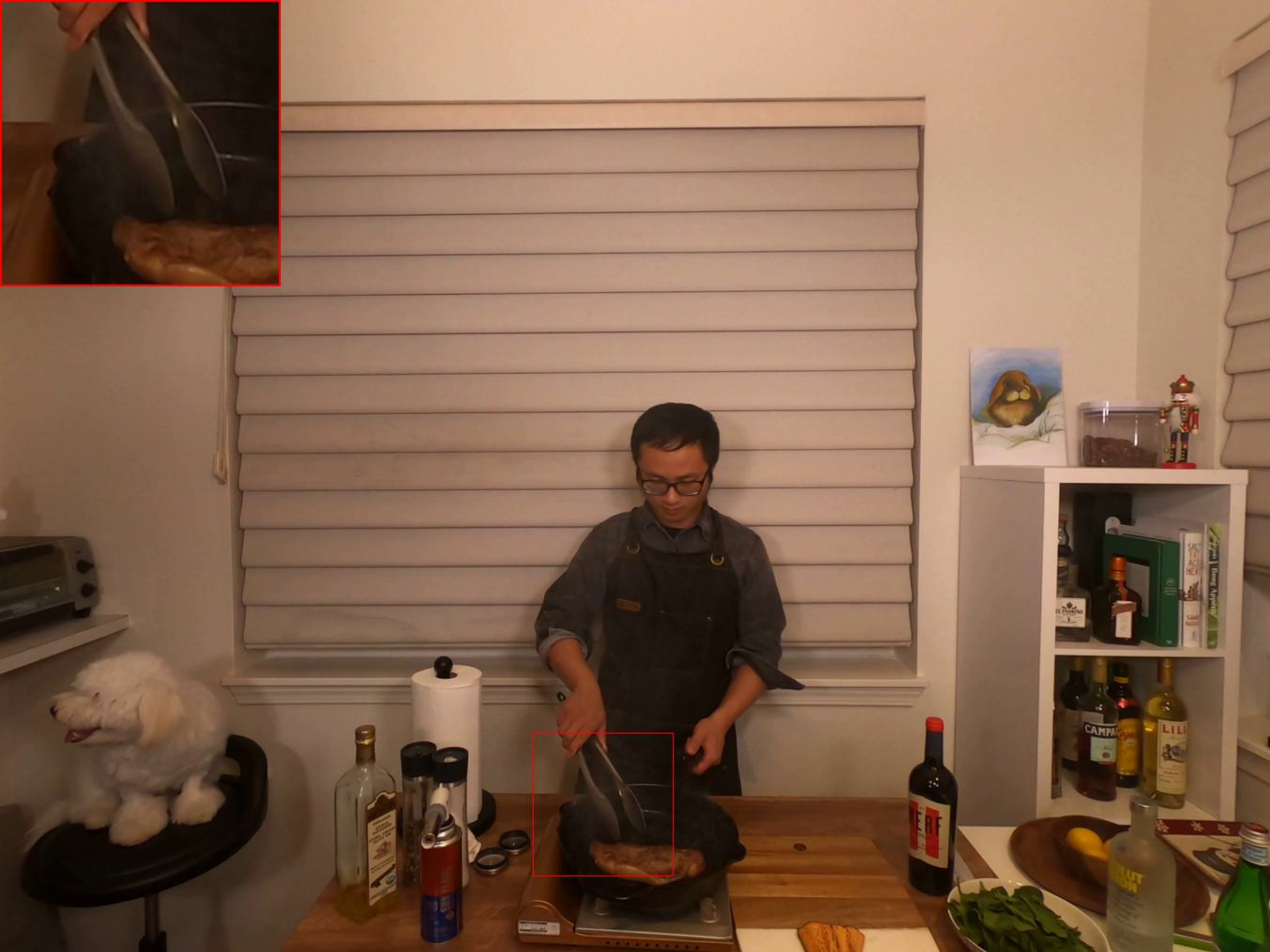}}

  \caption{\textbf{Qualitative results of the ablation study of structured 3DGs} conducted on the \textit{sear steak} scene.}
  \label{fig:ablation_study}
  
  \vspace{-2mm}
\end{figure}
\begin{figure}[!t]
    \centering
    \includegraphics[width=\linewidth]{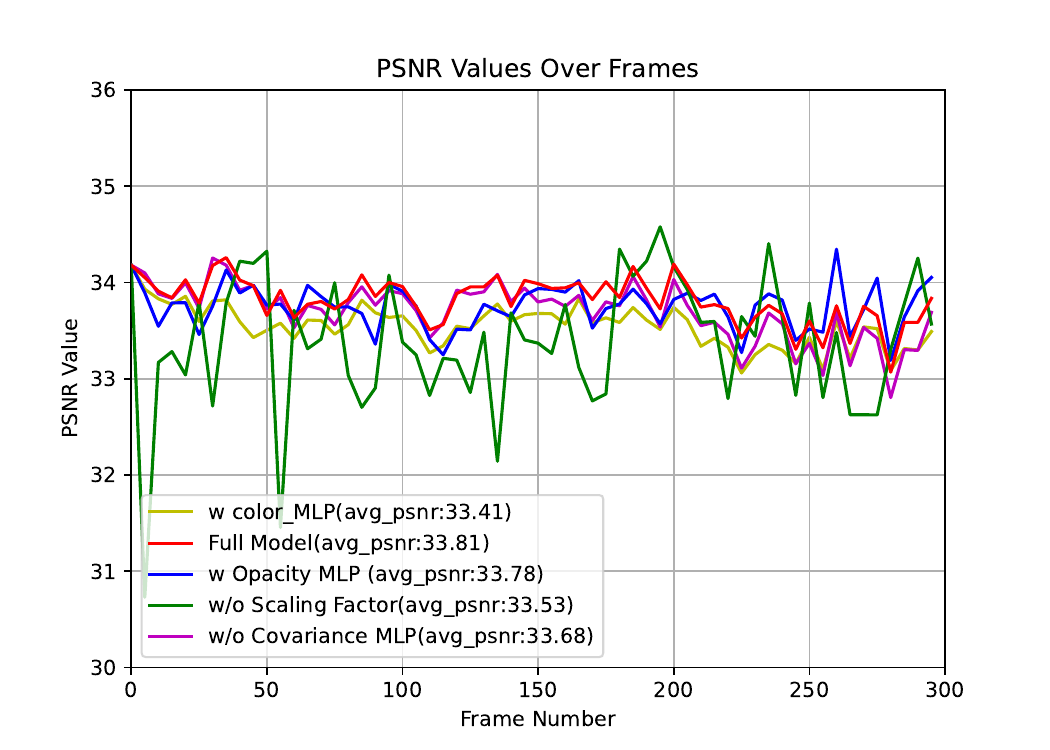}
    \caption{\textbf{Comparison of different variants for structured 3DGs on the} \textit{sear steak} \textbf{scene.}}
\label{fig:s1_frame}
\vspace{-5mm}
\end{figure}

\section{Experiments}
\subsection{Implementations}
\par
We implement Struct-GStream based on the Scaffold-GS~\cite{scaffoldgs} codebase and conduct experiments on an NVIDIA RTX 3090 GPU. Unless specified otherwise, the learning rates and experimental settings for the initial static scene reconstruction remain consistent with Scaffold-GS. During the initial frame training, we set the voxel size $\epsilon$ to 0.005. The number of structured 3DGs attached to each anchor point, $k$, is set to 5. We initiate the densification of Scaffold-GS at the 1500th iteration and select the anchor points and MLPs at the 4000th iteration as the initialization for subsequent processes. During the online training process, we perform 100 iterations for training both structured and free 3DGs for each frame.

\par
\textit{Training structured 3DGs and covariance MLP.} During online training, we specify the learning rates as follows: the anchor position $\lambda_{pos} = 0.0004$, the scaling factor $\lambda_{scale} = 0.005$, and the covariance MLP $\lambda_{cov} = 0.004$.

\par
\textit{Training free 3DGs.} The learning rates for the position, scale, rotation, opacity, and color of free 3DGs are set to 0.0002, 0.005, 0.001, 0.05, and 0.0025, respectively. We do not impose a hard limit on the number of free 3DGs; instead, their population is adaptively regulated by the model's densification and pruning mechanism. Regarding the pruning frequency, the generation and pruning of free 3DGs are performed once per frame, specifically at the 50th iteration out of the 100 total training iterations for each frame.

\subsection{Datasets}
We evaluate our method on three real-world dynamic scene datasets:  the N3DV dataset~\cite{DyNeRF}, the Meet Room dataset~\cite{StreamRF}, and the Google Immersive Dataset~\cite{10.1145/3386569.3392485}.
\par
\textbf{N3DV dataset} is captured using 21 cameras
at a resolution of 2704 $\times$ 2028, with each camera recording a 10-second video at 30 FPS.
To ensure a fair comparison, the images are downsampled to a resolution of 1352 $\times$ 1014 to align with all the methods being compared.
\par
\textbf{Meet Room dataset} is captured using 13 cameras,
with each camera recording a 10-second video at a resolution of
1280 $\times$ 720 and 30 FPS. In line with StreamRF~\cite{StreamRF} and 3DGStream~\cite{3dgstream}, we employed 12 views from each scene for training and set aside 1 for testing.
\par
\par
\textbf{Google Immersive dataset} features large-scale outdoor and indoor environments captured by a rig of 46 fisheye cameras. Since standard 3DG-S is designed for pinhole camera models, we undistorted the original fisheye images to rectilinear projection before training. We selected one camera as the test view and used the remaining cameras for training. All images were downsampled to a resolution of 1280 $\times$ 960.

\subsection{Comparisions}

\begin{table}[t]
    \caption{\textbf{Ablation study of free 3DG generation and pruning} on the \textit{flame\_steak} scene.
        Metrics are averaged over the whole 300 frames.}
    \label{tab:s2_eva}
    \centering
    \resizebox{\linewidth}{!}{
    \begin{tabular}{@{}l|cc|cc@{}}
        \toprule
        \multirow{2}{*}{Variant} & PSNR $\uparrow$ & External Storage $\downarrow$ & T-LPIPS $\downarrow$ & $E_{warp}$ $\downarrow$ \\
         & (dB) & (MB) & ($\times 10^{-2}$) & ($\times 10^{-3}$) \\
        \midrule
        Baseline & 33.26 & \cellcolor{first}0 & 1.55 & 4.89 \\
        Rnd. Generate & 33.48 & 2.28 & 1.52 & 4.85 \\
        \textit{w/o} Quant.Ctrl & \cellcolor{second}33.93 & 2.87 & \cellcolor{second}1.45 & \cellcolor{second}4.70 \\
        Full Model & \cellcolor{first}33.98 & \cellcolor{second}1.40 & \cellcolor{first}{1.42} & \cellcolor{first}{4.41} \\
        \bottomrule
    \end{tabular}
    }
    
  \vspace{-2mm}
\end{table}
\begin{figure}[!t]
    \centering
    \subfloat[\textit{w/o} Free 3DGs]{\includegraphics[width=0.23\textwidth]{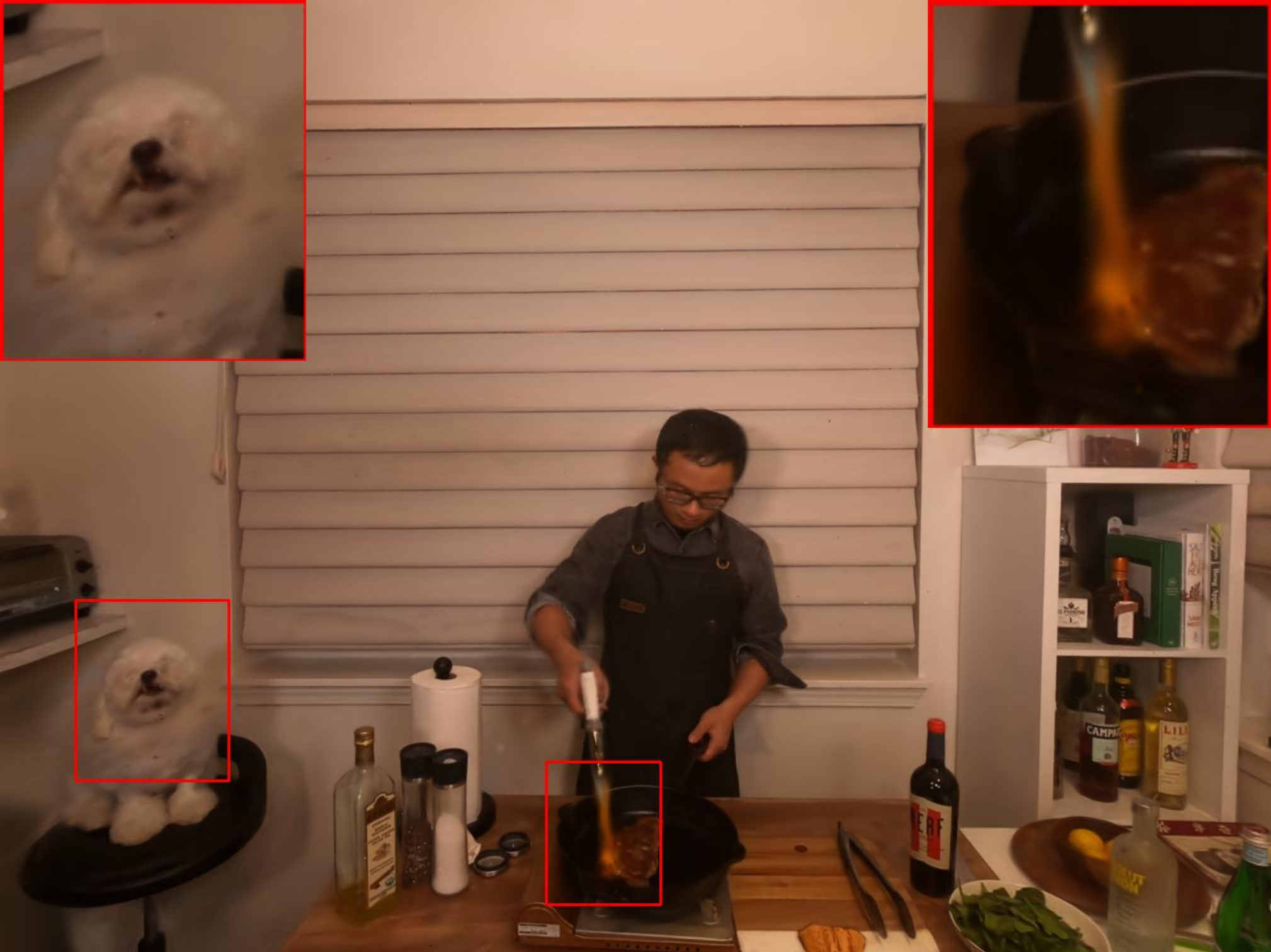}}
    \hfill
    \subfloat[\textit{w} Rnd. Generate]{\includegraphics[width=0.23\textwidth]{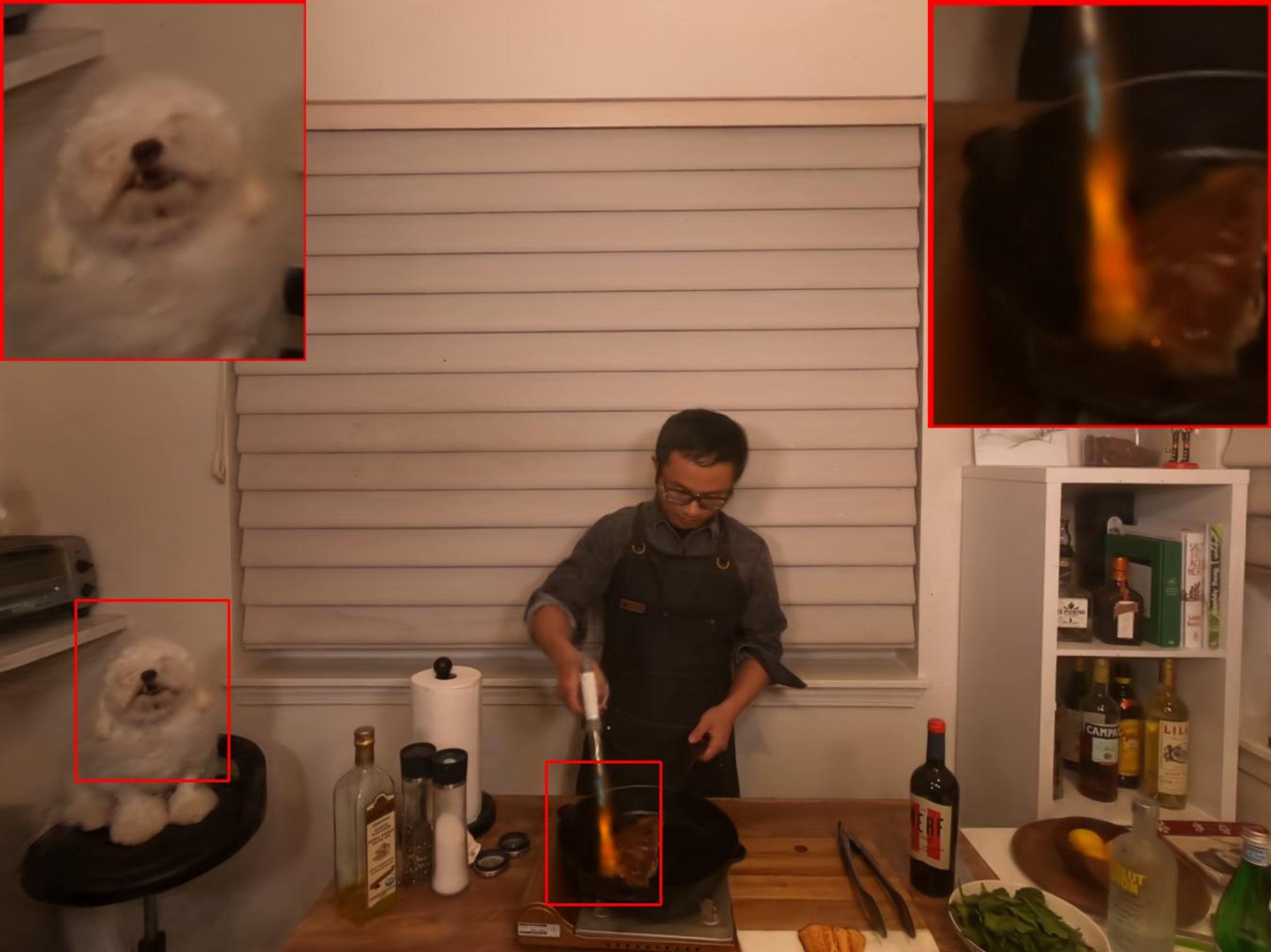}}
    \hfill
    \subfloat[\textit{w/o} Quant.Ctrl.]{\includegraphics[width=0.23\textwidth]{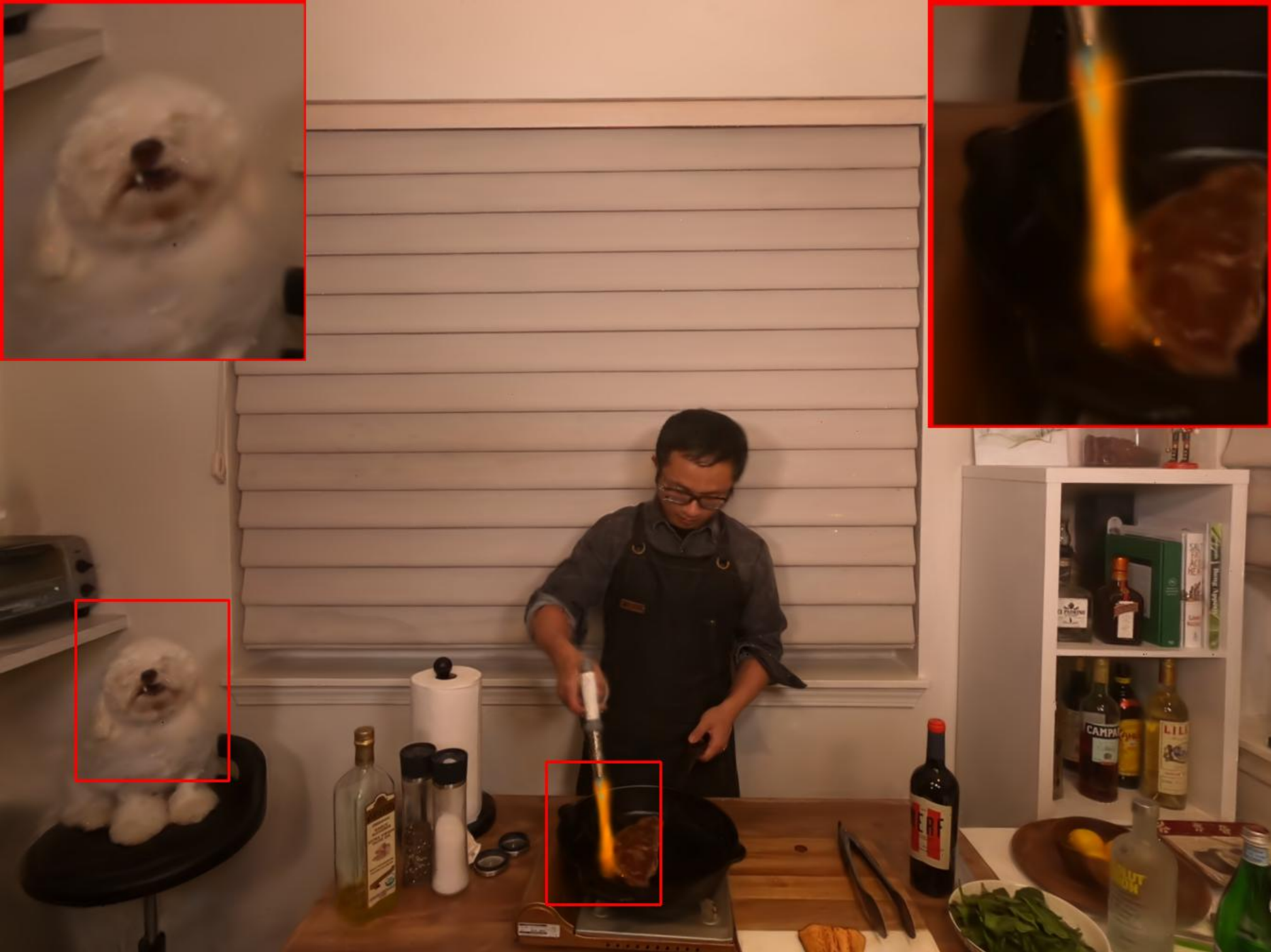}}
    \hfill
    \subfloat[Full Model]{\includegraphics[width=0.23\textwidth]{figs/abl/flame_steak80/processed_zoom/Full_Model_80.pdf}}
    \hfill
    \subfloat[Ground Truth]{\includegraphics[width=0.23\textwidth]{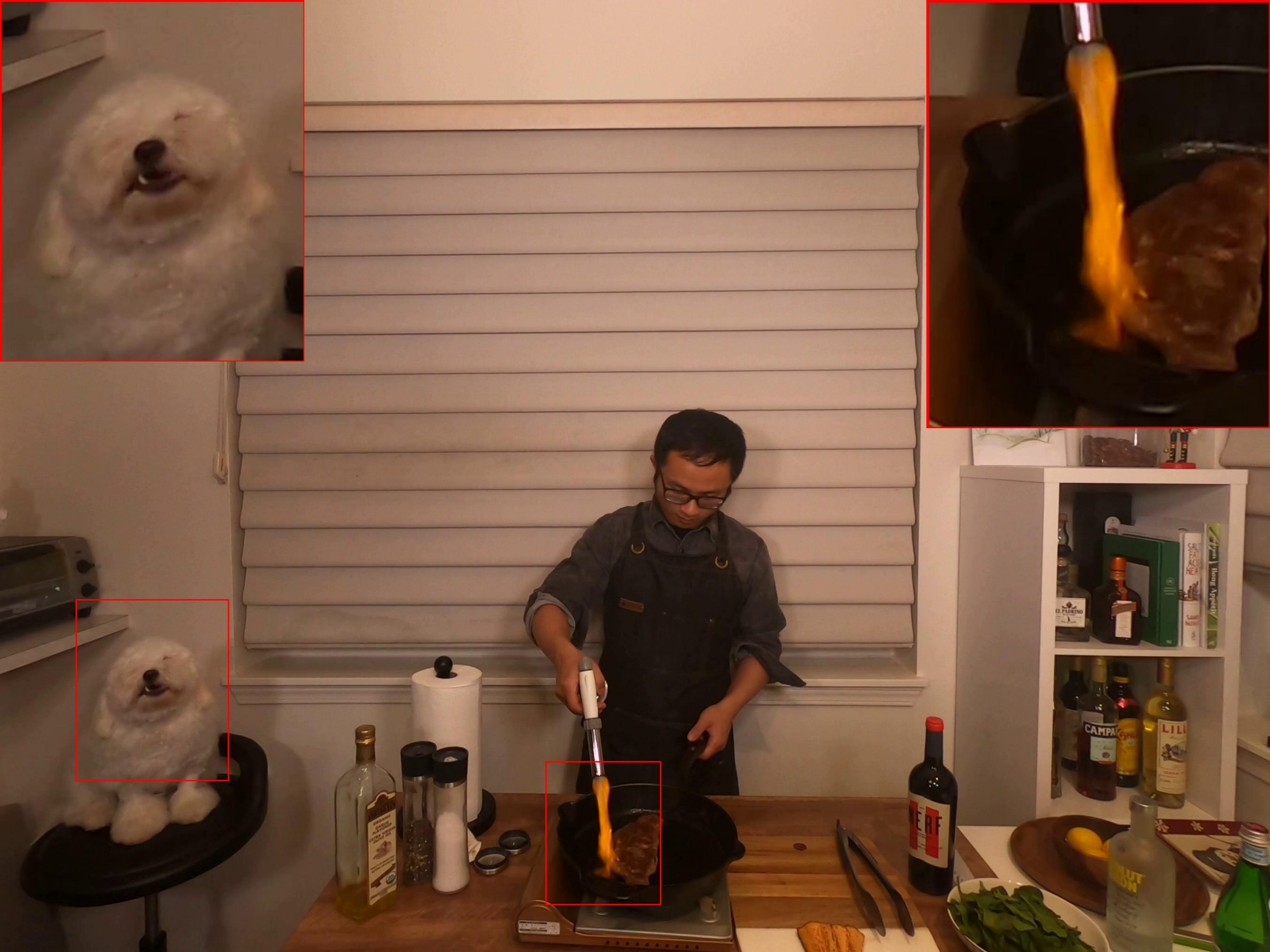}}
    \caption{\textbf{Qualitative results of the ablation study of free 3DGs} conducted on the \textit{flame steak} scene.}
    \label{fig:s2_abl}
    
  \vspace{-2mm}
\end{figure}
\begin{figure}
    \centering
    \includegraphics[width=\linewidth]{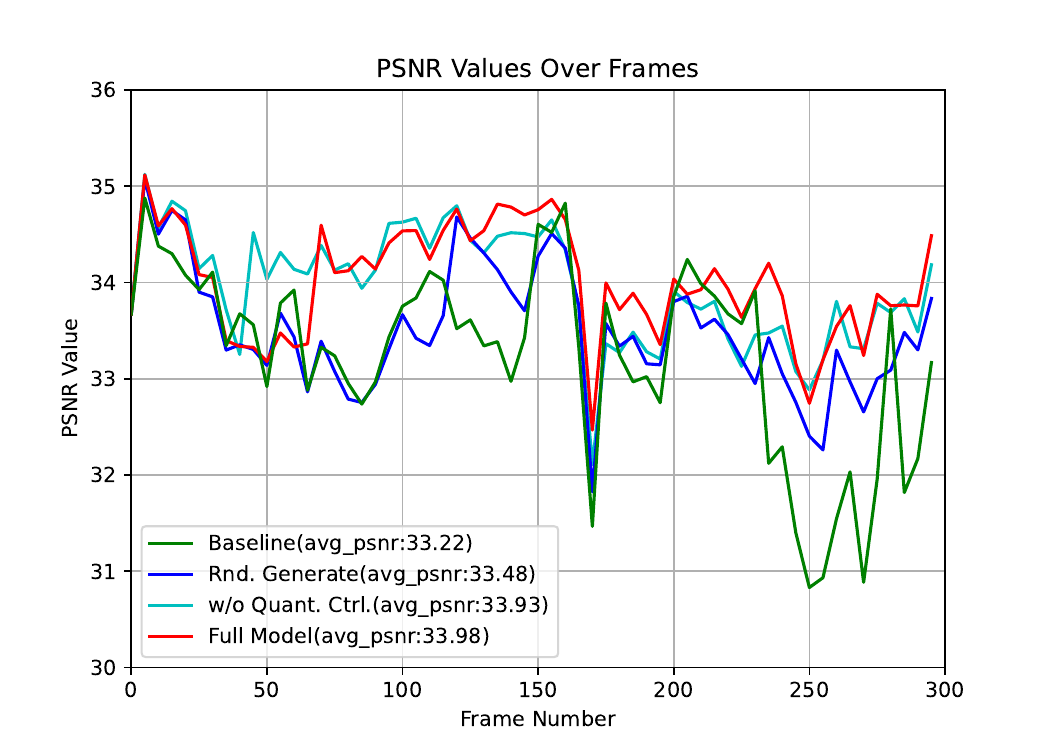}
    \caption{\textbf{Comparison of different variants for free 3DGs on the} \textit{flame steak} \textbf{scene.}}
    
  \vspace{-2mm}
\label{fig:s2_frame}
\end{figure}
\begin{figure}[!t]
    \centering
    \subfloat[Baseline]{\includegraphics[width=0.23\textwidth]{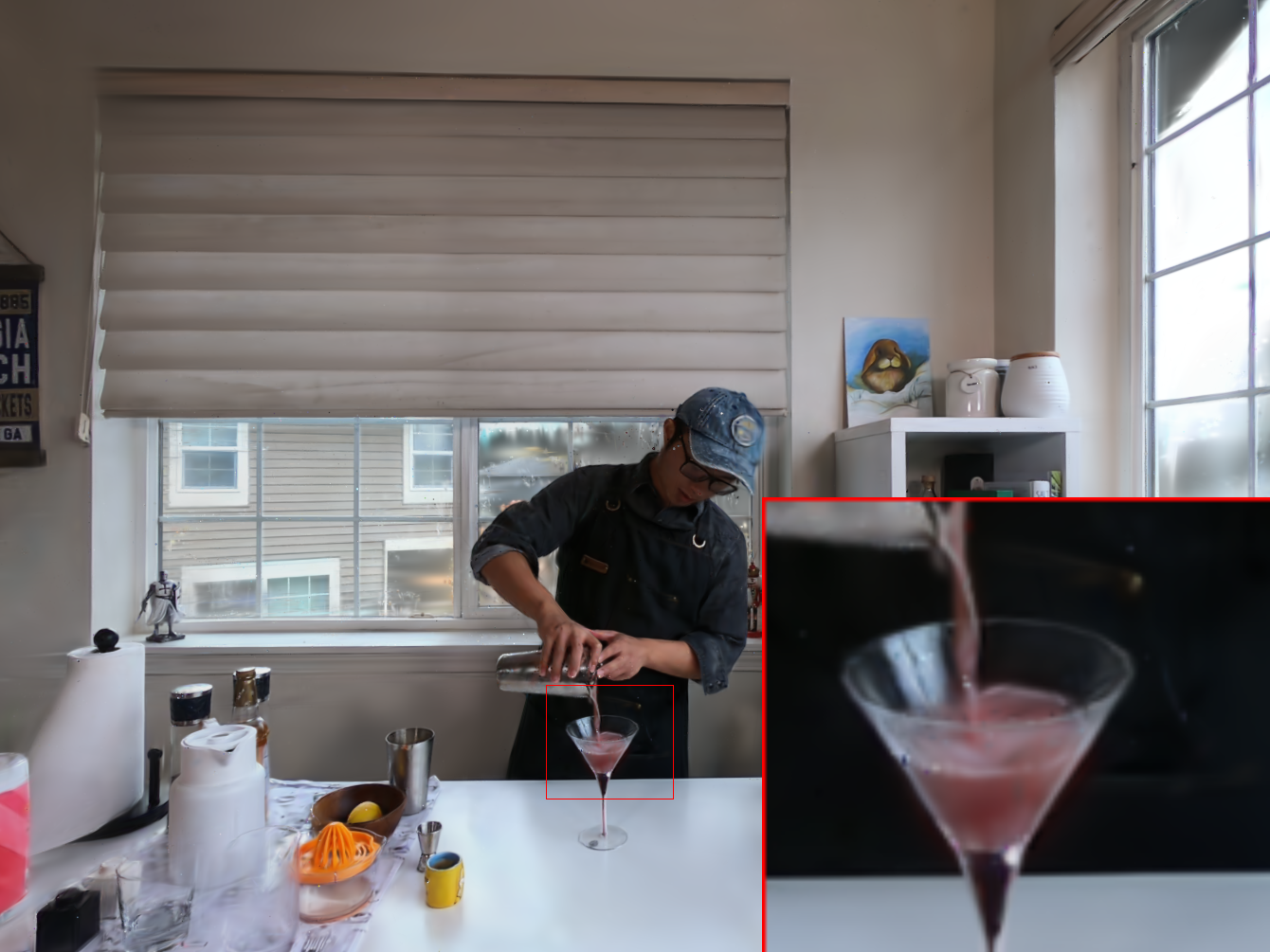}}
    \hfill
    \subfloat[I1]{\includegraphics[width=0.23\textwidth]{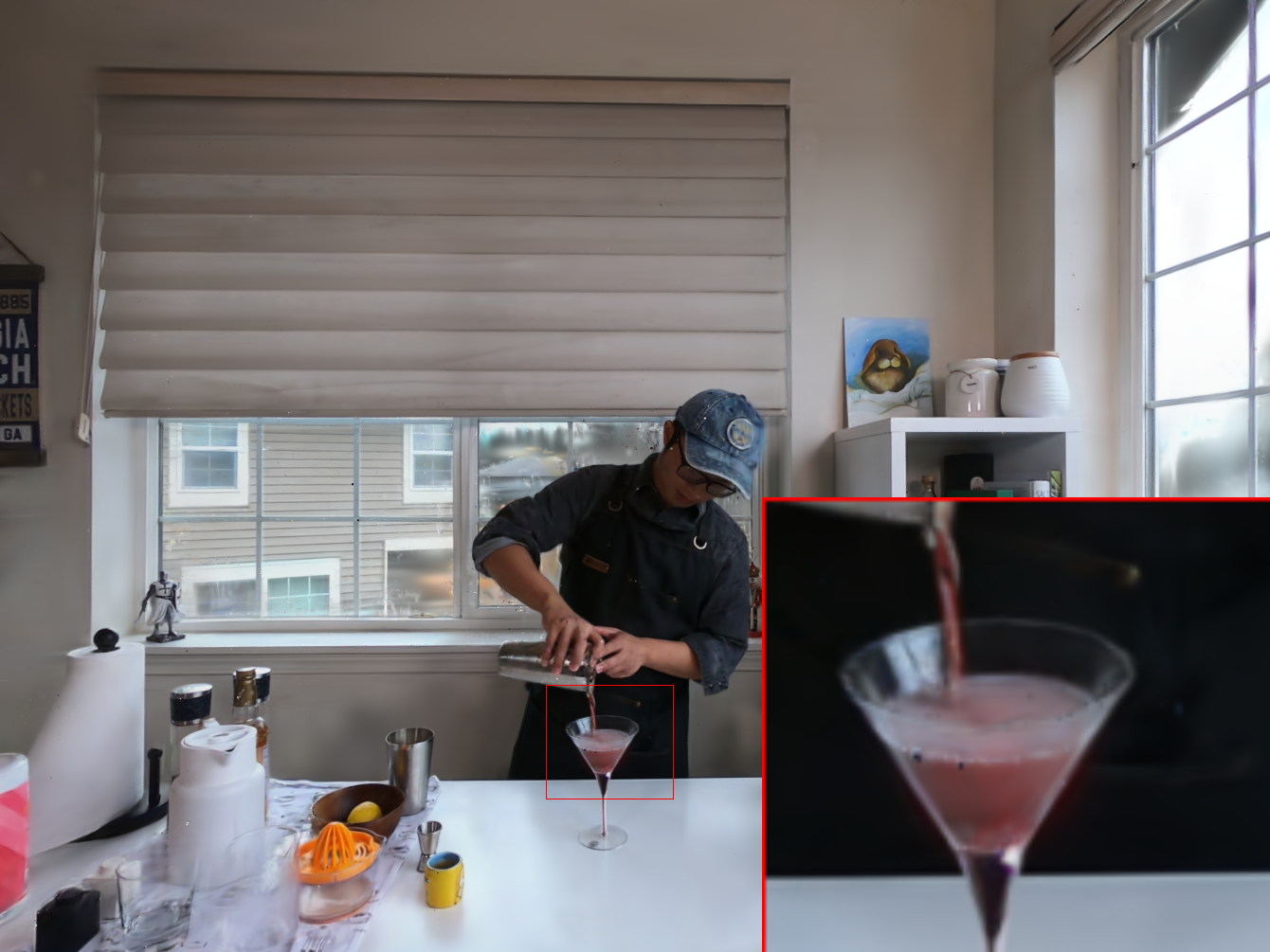}}
    \hfill
    \subfloat[I2]{\includegraphics[width=0.23\textwidth]{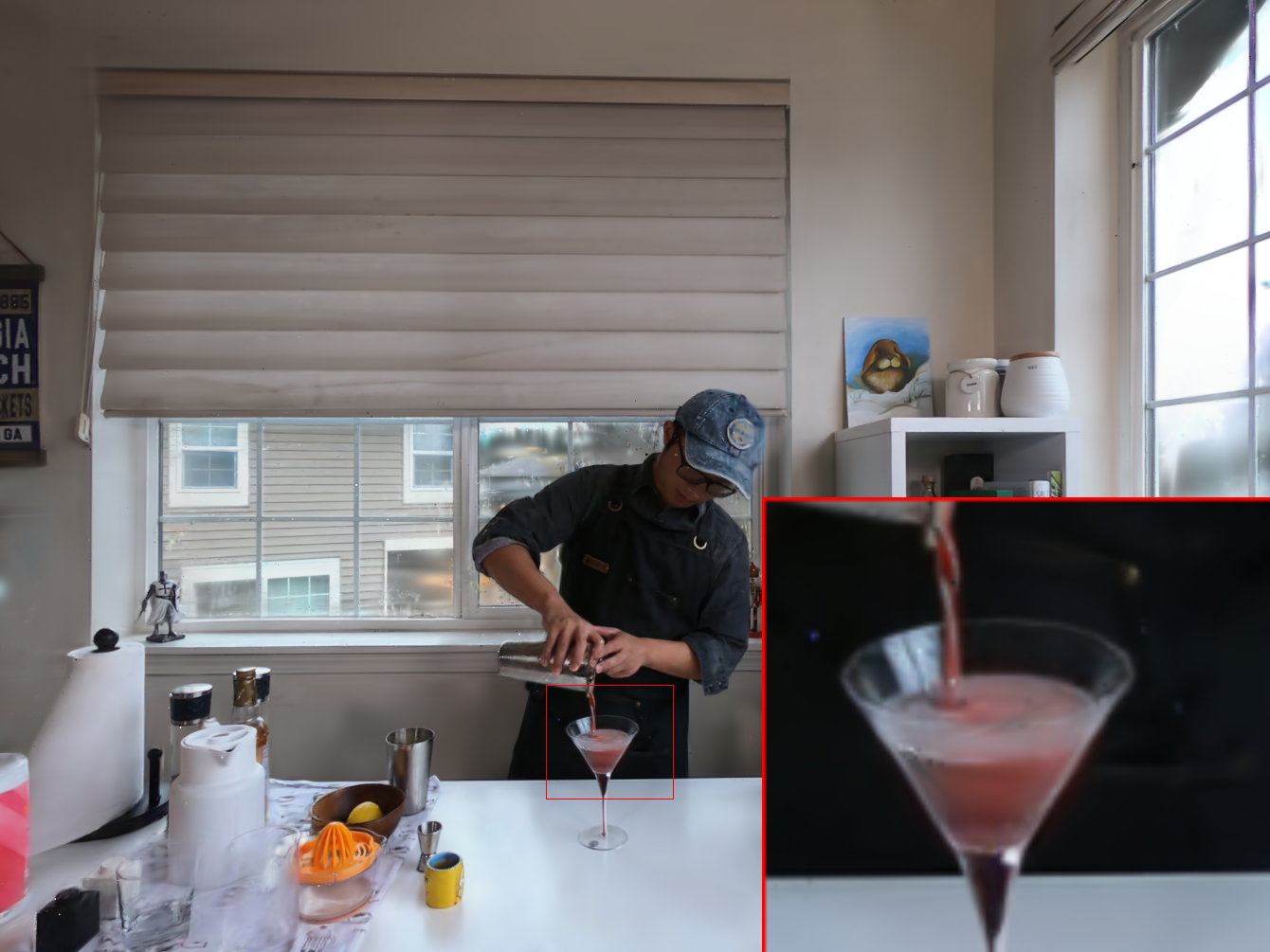}}
    \hfill
    \subfloat[I3]{\includegraphics[width=0.23\textwidth]{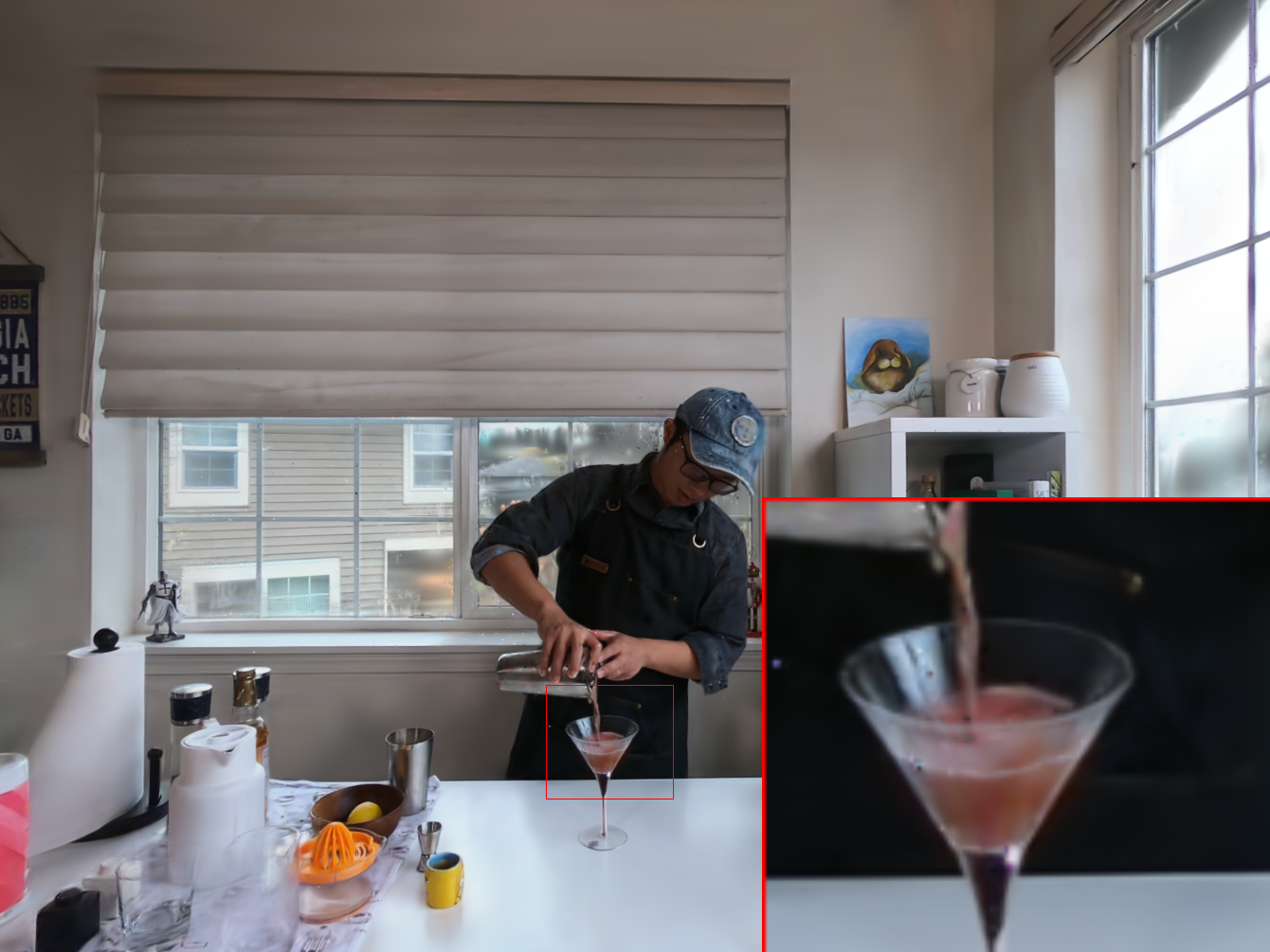}}
    \hfill
    \subfloat[Ours]{\includegraphics[width=0.23\textwidth]{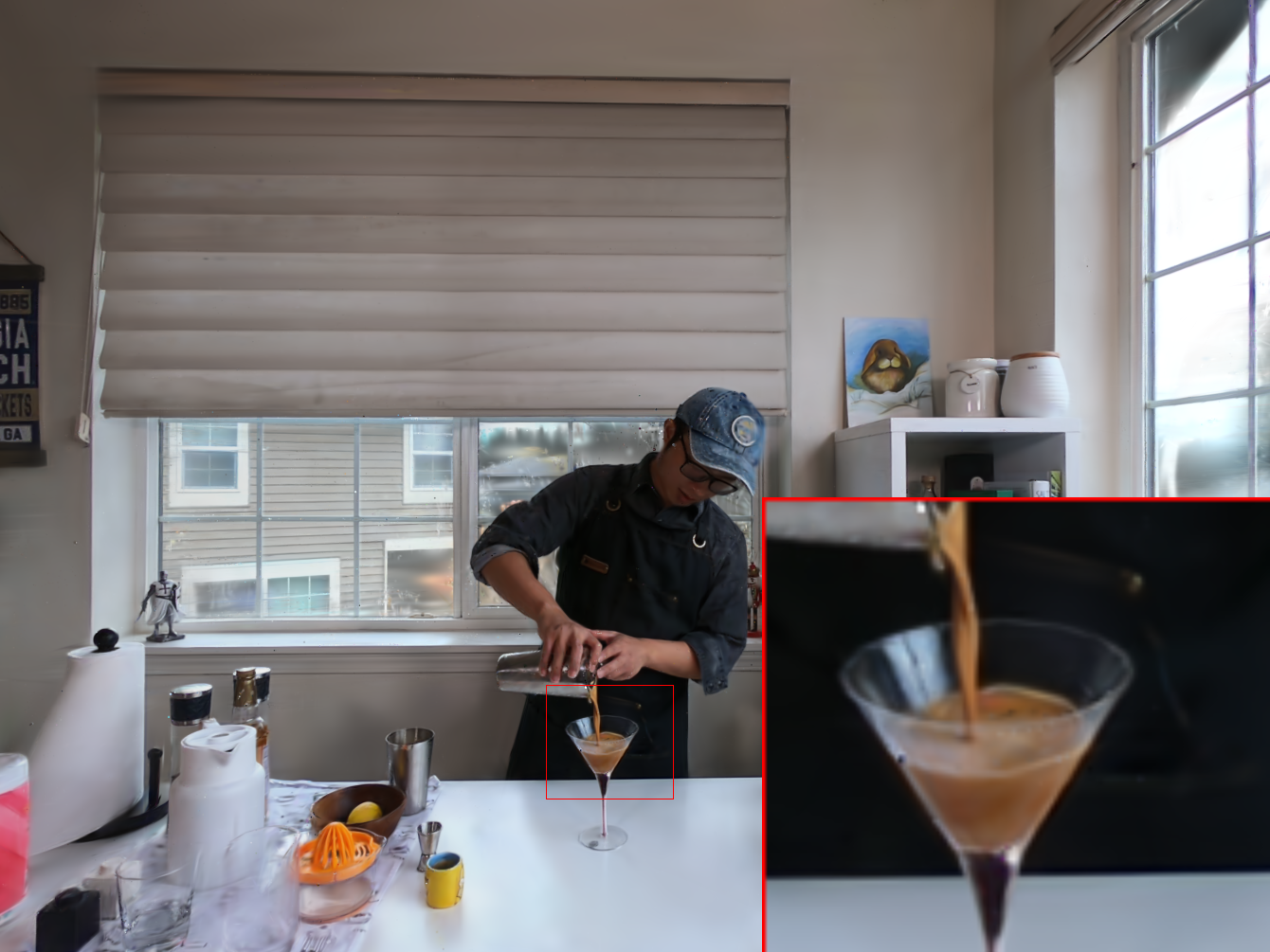}}
    \hfill
    \subfloat[GT]{\includegraphics[width=0.23\textwidth]{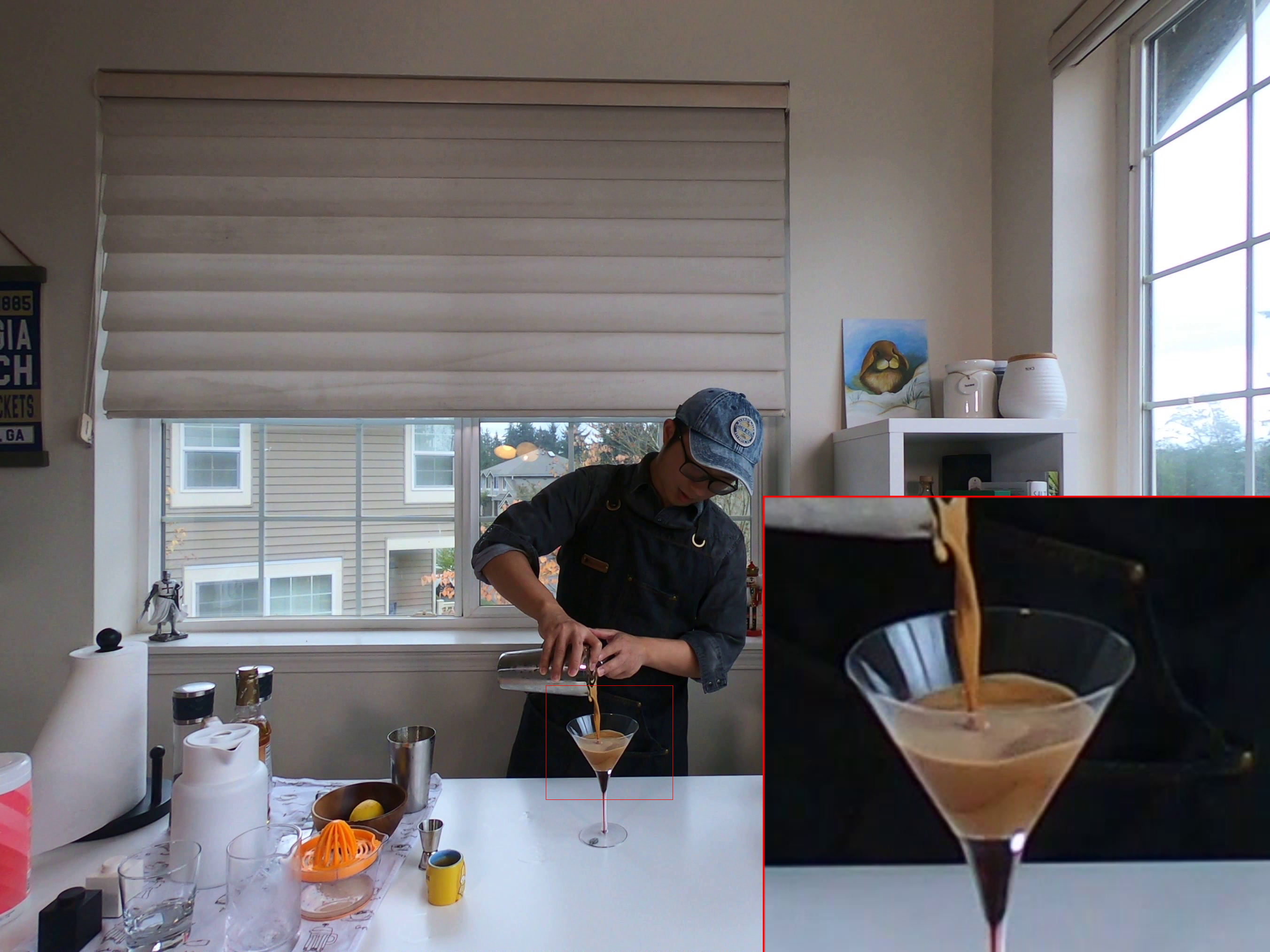}}
    \hfill
    \caption{\textbf{Results of the ablation study of global patching strategy} conducted on the \textit{coffee martini} scene.}
    \label{fig:s2_abl_global}
    
  \vspace{-3mm}
\end{figure}
\begin{figure}[!t]
  \centering
  \subfloat[3DGStream: Frame44]{\includegraphics[width=0.23\textwidth]{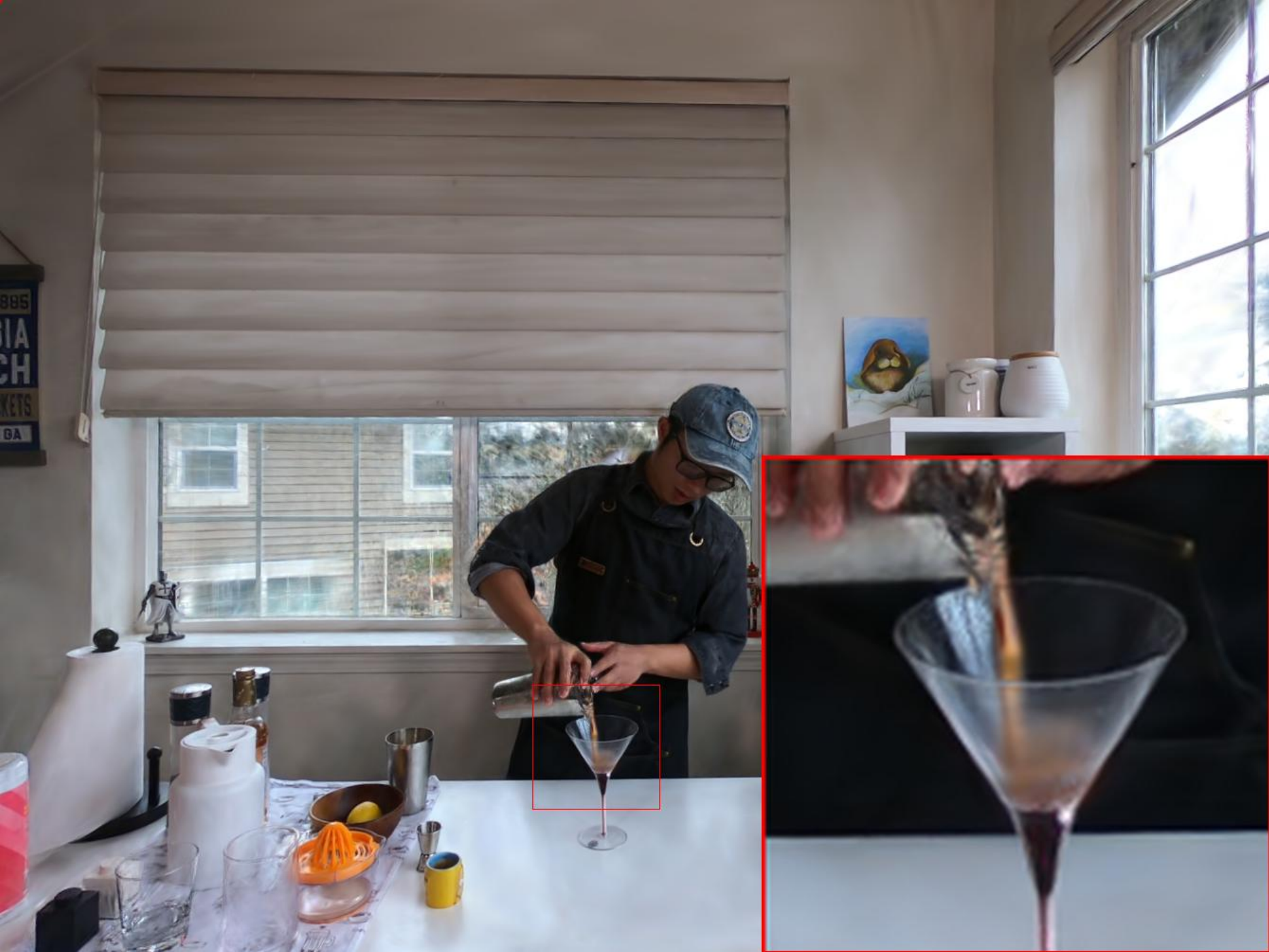}}
  \hfill
  \subfloat[3DGStream: Frame45]{\includegraphics[width=0.23\textwidth]{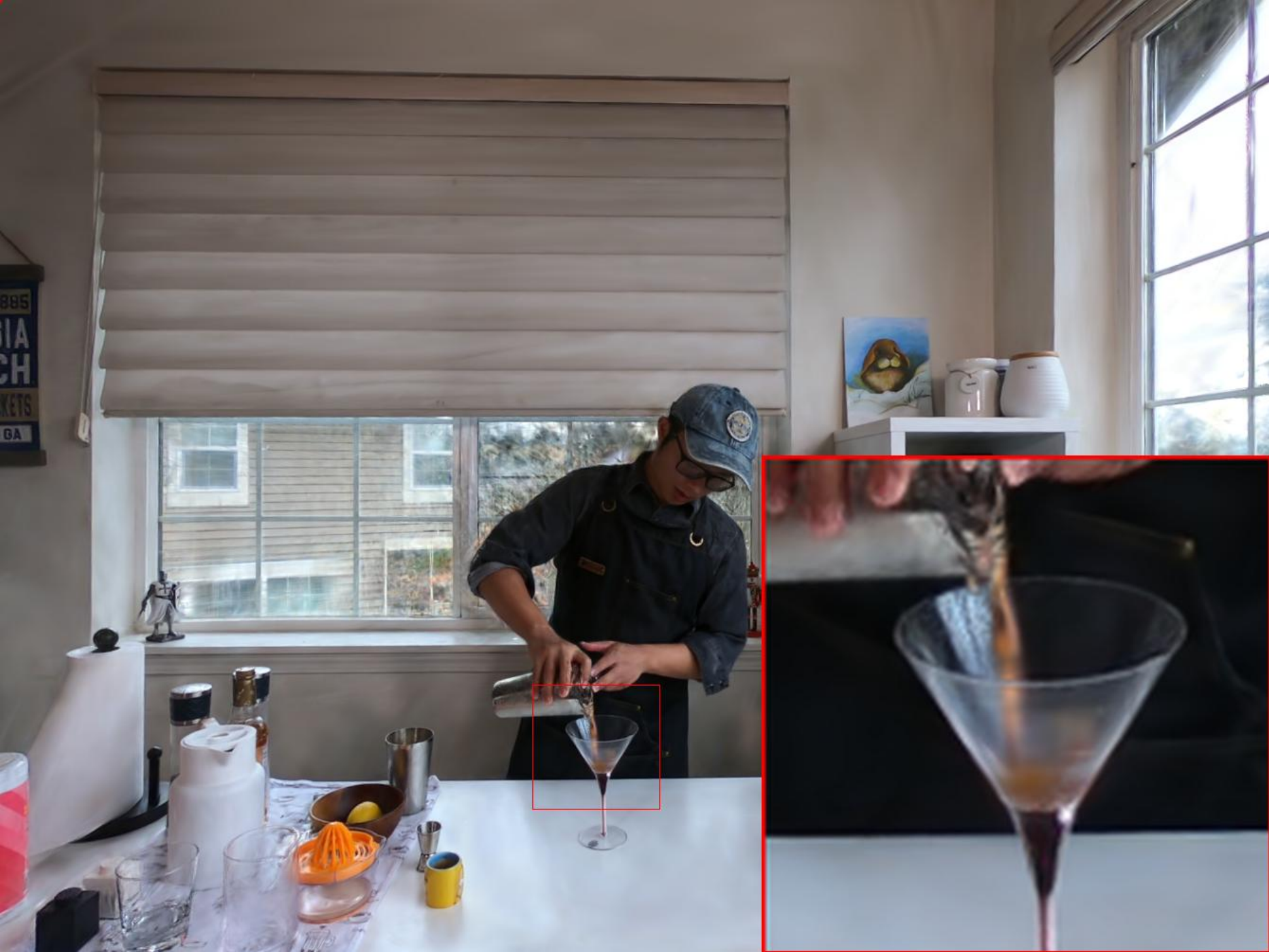}}

  \subfloat[Ours: Frame44]{\includegraphics[width=0.23\textwidth]{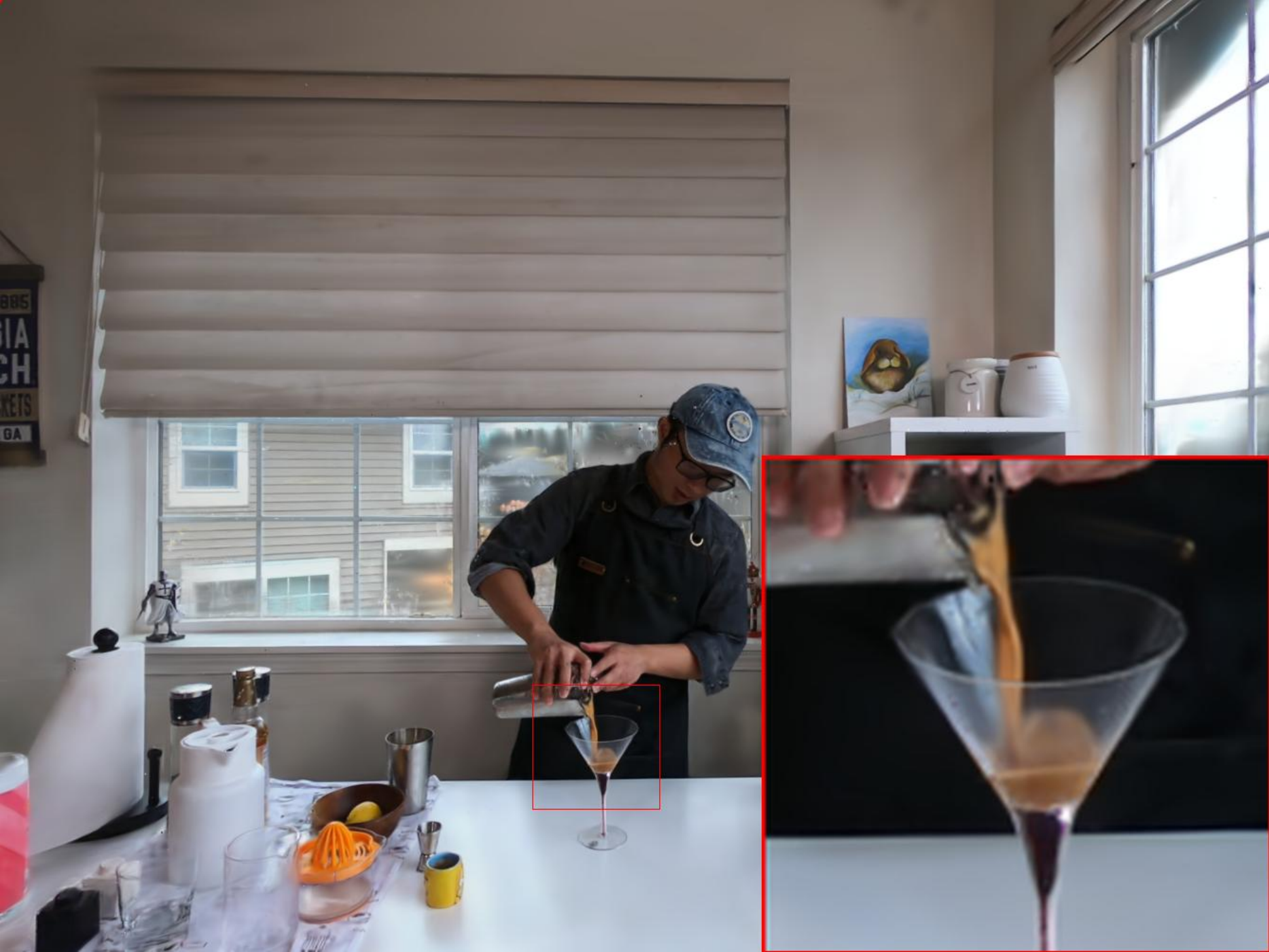}}
  \hfill
  \subfloat[Ours: Frame45]{\includegraphics[width=0.23\textwidth]{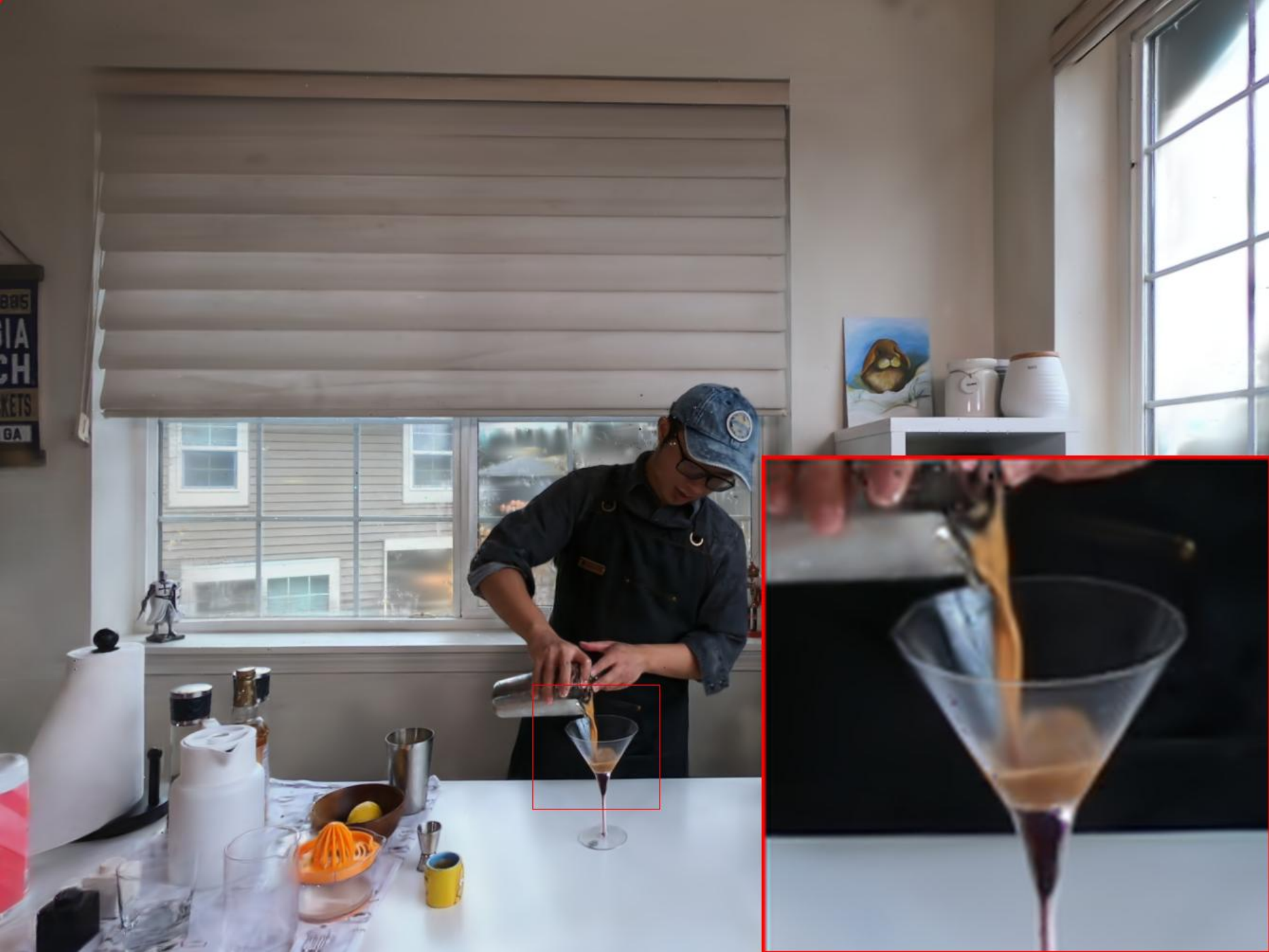}}

  \subfloat[GT: Frame44]{\includegraphics[width=0.23\textwidth]{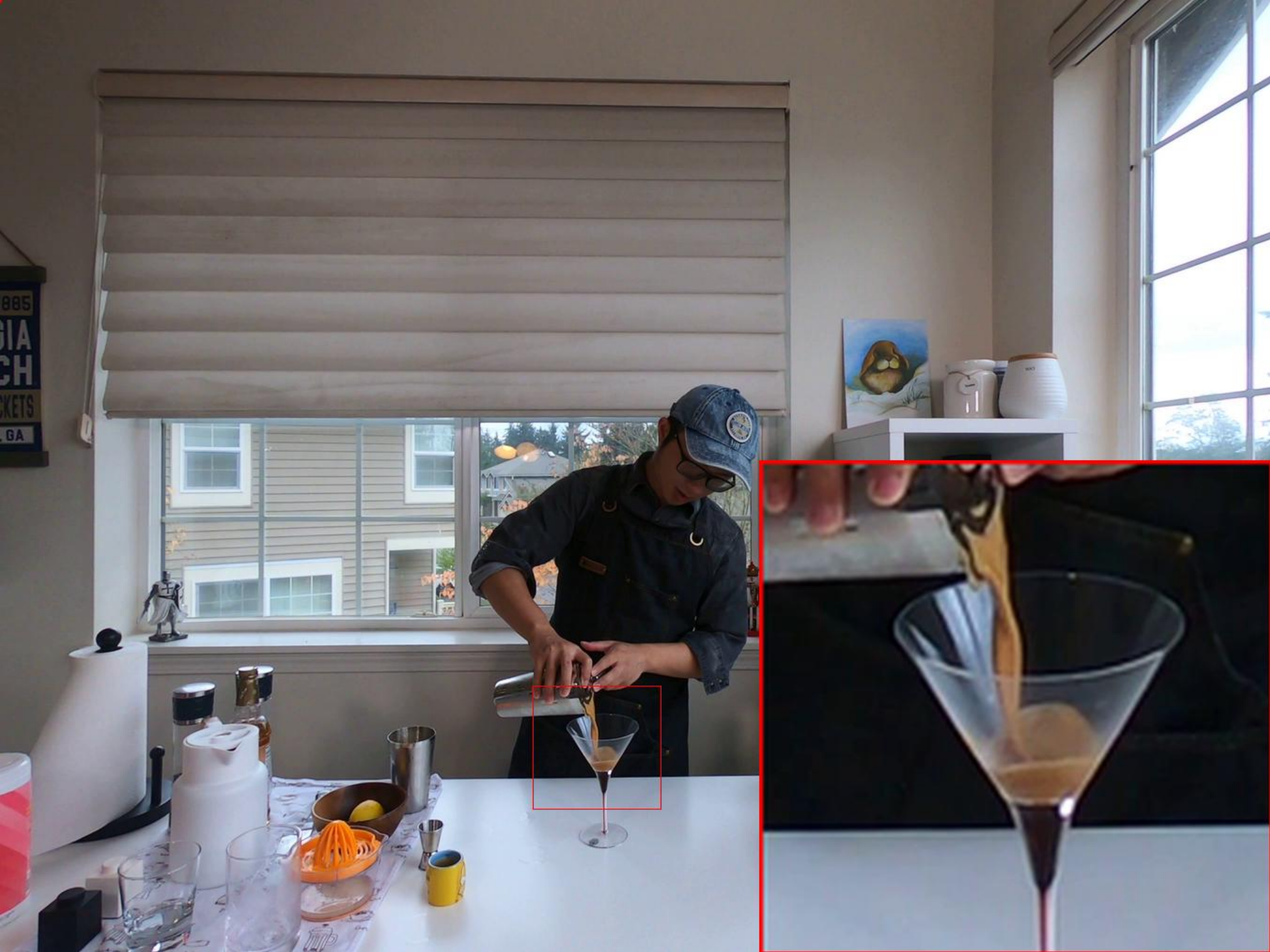}}
  \hfill
  \subfloat[GT: Frame45]{\includegraphics[width=0.23\textwidth]{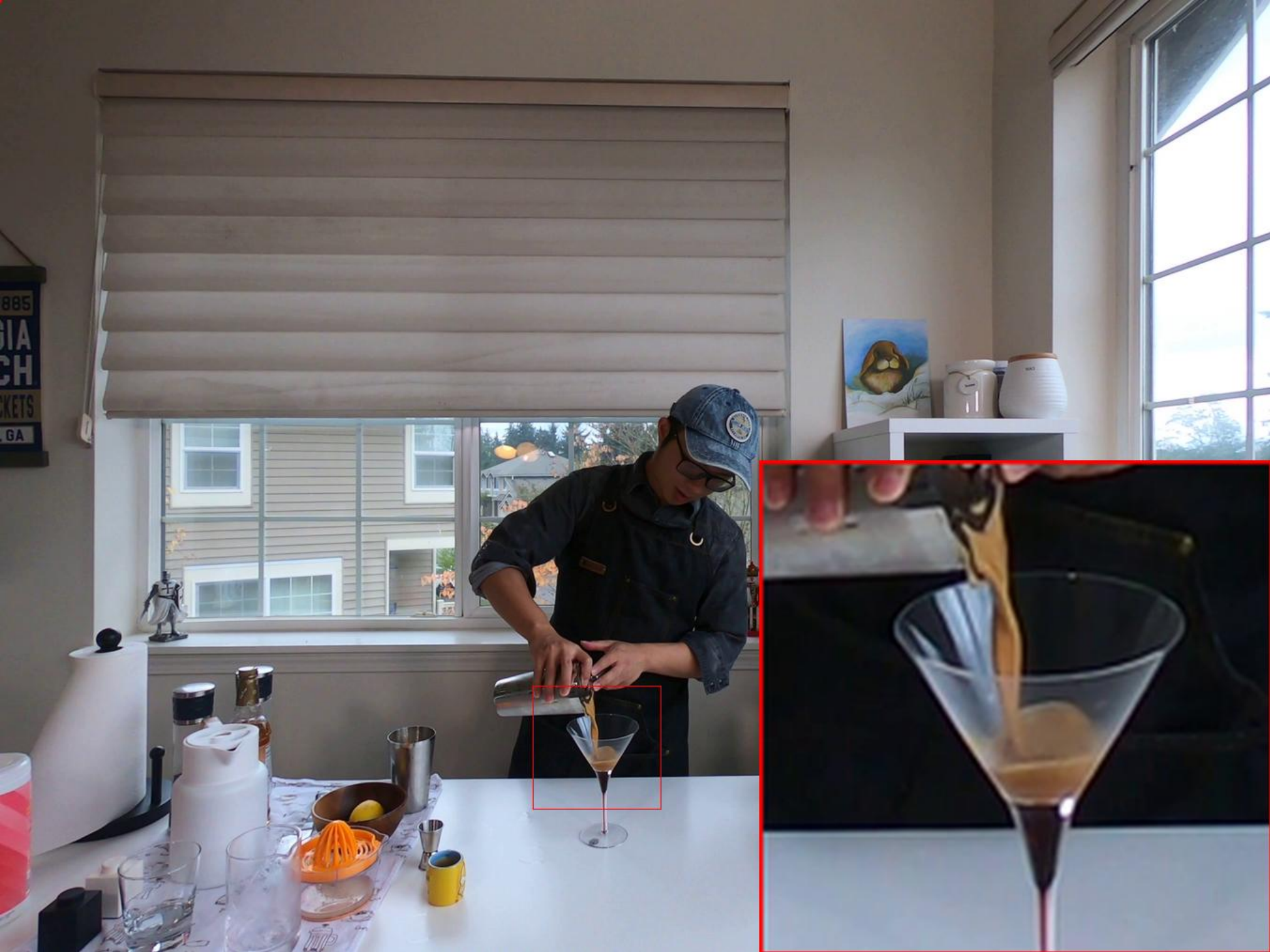}}
  
  \caption{\textbf{Time consistency comparison} of emerging object modeling in coffee martini of N3DV dataset.}
  \label{fig:timeconsistency}
  
  \vspace{-2mm}
\end{figure}

\subsubsection{Quantitative comparisons}
We compare our method with previous state-of-the-art NeRF-related methods
and contemporaneous 3DG-S-related work.
We compare our method with Plenoxels~\cite{plenoxels}, I-NGP~\cite{Instant-NGP}, 3DG-S~\cite{3DGS}, and Scaffold-GS~\cite{scaffoldgs}
as representatives of static reconstruction methods, training them frame by frame from scratch. 
For 3DG-based methods, we used the same point cloud input.
In dynamic reconstruction methods,
to showcase compact storage, fast training advantages, and competitive rendering quality and speed,
we not only compare our method with online training methods StreamRF~\cite{StreamRF}, iFVC~\cite{Tang_Yang_Peng_Zhai_Shen_Wang_2025} and 3DGStream~\cite{3dgstream},
but also conduct comparisons with offline training methods DyNeRF~\cite{DyNeRF}, NeRFPlayer~\cite{nerfplayer}, HexPlane~\cite{Hexplane},
K-Planes~\cite{K-Planes}, HyperReel~\cite{Hyperreel}, MixVoxels~\cite{mixedvoxel}, Realtime4DGS~\cite{realtime4dgs}, 4DGaussians~\cite{4dgs_kplanes},
and SpacetimeGaussians~\cite{SpacetimeGaussians}.
\par
We analyze method performance on the N3DV dataset in Tab. \textcolor{red}{\ref{tab:N3DV_Comparisons_Avg}}, showing our approach's strong results, notably achieving the fastest training speed and superior rendering quality, along with highly competitive storage efficiency when compared to other online training methods.
We also present the results of the Meet Room dataset and Google Immersive dataset in Tab. \textcolor{red}{\ref{tab:meetroom_Comparisons}} and Tab. \textcolor{red}{\ref{tab:immersive_Comparisons}}. Experiments conducted on these datasets
serve to further demonstrate the comprehensive advantages of our method, encompassing rendering quality, storage requirements, and
training time.
Moreover, they highlight the generalizability of our approach.

\par
\subsubsection{Qualitative comparisons}
We present the comparative results with StreamRF, iFVC, and
3DGStream for online FVV construction
on the N3DV dataset, the Meet Room dataset, and the Google Immersive dataset
in Fig. \textcolor{red}{\ref{fig:visual_Comparisons}} and Fig. \textcolor{red}{\ref{fig:immersive_vis}}.
Our method achieves superior rendering quality compared to the state-of-the-art,
underscoring its competitive advantage in FVV construction.

\subsection{Evaluations}
\subsubsection{Structured 3DGs} In our experiment, we optimize the position $x_{v}$,
scaling factor $l_v$ of dynamic anchor points and MLP $F_{\Sigma}$.
However, we did not include the optimization of structured 3DGs offsets $O_v$ and the remaining two MLPs $F_{\alpha}$ and $F_{c}$.
Because including the optimization of structured 3DGs offsets $O_v$
would not only reduce convergence speed but also increase the attributes that need to be stored,
leading to a significant increase in storage requirements.
The insufficient optimization of the other two MLPs $F_{\alpha}$ and $F_{c}$ can result in a degradation of scene reconstruction quality.
We validate the effectiveness of our method through four variants which
are conducted by removing scaling factor's optimization (\textit{w/o} Scaling Factor), removing
covariance MLP's optimization (\textit{w/o} Covariance MLP), adding opacity MLP's optimization (\textit{w} Opacity MLP), adding color MLP's optimization (\textit{w} Color MLP).
Tab. \textcolor{red}{\ref{tab:rendering}} demonstrates that without training scaling factor or covariance MLP,
our method experiences a certain decline in reconstruction quality which is also shown in Fig. \textcolor{red}{\ref{fig:ablation_study}}. Meanwhile, incorporating $F_{\alpha}$ and $F_{c}$ for optimization does not
lead to any improvement in our method but rather results in a decline in image quality. We also provide an overview of our training results of variants in structured 3DGs' evaluation on the sear steak scene in Fig. \textcolor{red}{\ref{fig:s1_frame}}.

\par
\begin{table}[t]
\centering
\caption{\textbf{Ablation Study of the Global Patching Strategy} on the \textit{coffee\_martini} scene.}
\label{tab:ablation_global_patch}
\resizebox{\linewidth}{!}{
\begin{tabular}{@{}l|c|cc|cc@{}}
\toprule
\multirow{2}{*}{Method} & Generation & PSNR $\uparrow$ & External Storage $\downarrow$ & T-LPIPS $\downarrow$ & $E_{warp}$ $\downarrow$ \\
 & Strategy & (dB) & (MB) & ($\times 10^{-2}$) & ($\times 10^{-3}$) \\
\midrule
Baseline & \textit{w/o} Free 3DGs & 27.60 & {0} & 1.46 & 73.1 \\
Variant I1 & Indep. ($\tau_{grad}=0.00100$) & 27.74 & {0.01} & 1.44 & 68.0 \\
Variant I2 & Indep. ($\tau_{grad}=0.00015$) & 27.67 & 0.29 & 1.44 & 67.6 \\
Variant I3 & Indep. ($\tau_{grad}=0.00010$) & 27.70 & 0.41 & 1.43 & 67.1 \\
Ours & Global Patching & {27.83} & 0.38 & {1.39} & {65.9} \\
\bottomrule
\end{tabular}
}
\end{table}
\subsubsection{Free 3DGs}
As illustrated in Tab. \textcolor{red}{\ref{tab:s2_eva}}, we evaluated the impact of free 3DGs' generation and pruning on the storage, image quality, and temporal consistency in FVV construction through several approaches:
removing all free 3DGs (Baseline), generating free 3DGs without pruning (\textit{w/o} Quant. Ctrl),
generating free 3DGs in random position (Rnd. Generate), and Full Model. To rigorously quantify temporal consistency, we introduced Temporal LPIPS (T-LPIPS) and Optical Flow Warping Error ($E_{warp}$), where lower values indicate better temporal consistency.
\par
By comparing with a baseline that does not generate free 3DGs at all and the full model, we found that our free 3DGs significantly improve the qualitative and quantitative
rendering effect of images at a relatively low storage cost and better temporal consistency. The qualitative results of the ablation study are described in Fig. \textcolor{red}{\ref{fig:s2_abl}}. The quality comparison between Rnd. Generate and Full Model demonstrates the effectiveness of our method for generating free 3DGs based on view space positional gradients. 
 The quantitative results in Tab. \textcolor{red}{\ref{tab:s2_eva}} confirm that our Full Model achieves the best trade-off across all metrics. While the \textit{w/o} Quant. Ctrl variant achieves respectable quality and temporal consistency, it suffers from high storage costs. By incorporating our pruning strategy, the Full Model not only halves the storage but also further improves temporal consistency, demonstrating that our method effectively eliminates redundant primitives to ensure a compact, high-quality, and temporally coherent video stream.
We also provide an overview of our training results of variants in free 3DGs' evaluation on the flame steak scene in Fig. \textcolor{red}{\ref{fig:s2_frame}}.

\par
\begin{figure}[!t]
  \centering
  \subfloat[\textit{w/o} dist]{\includegraphics[width=0.23\textwidth]{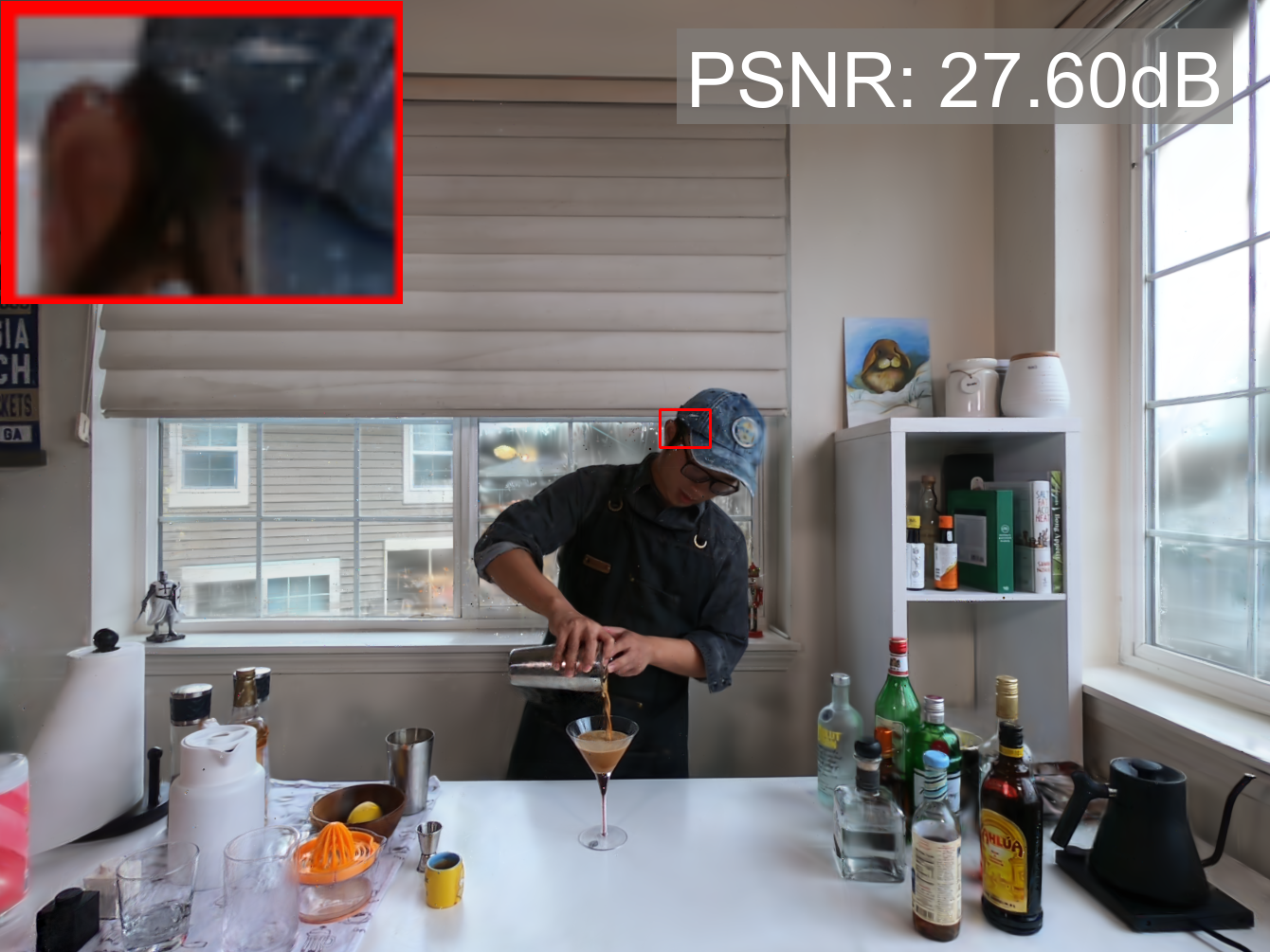}}
  \hfill
  \subfloat[Full Model]{\includegraphics[width=0.23\textwidth]{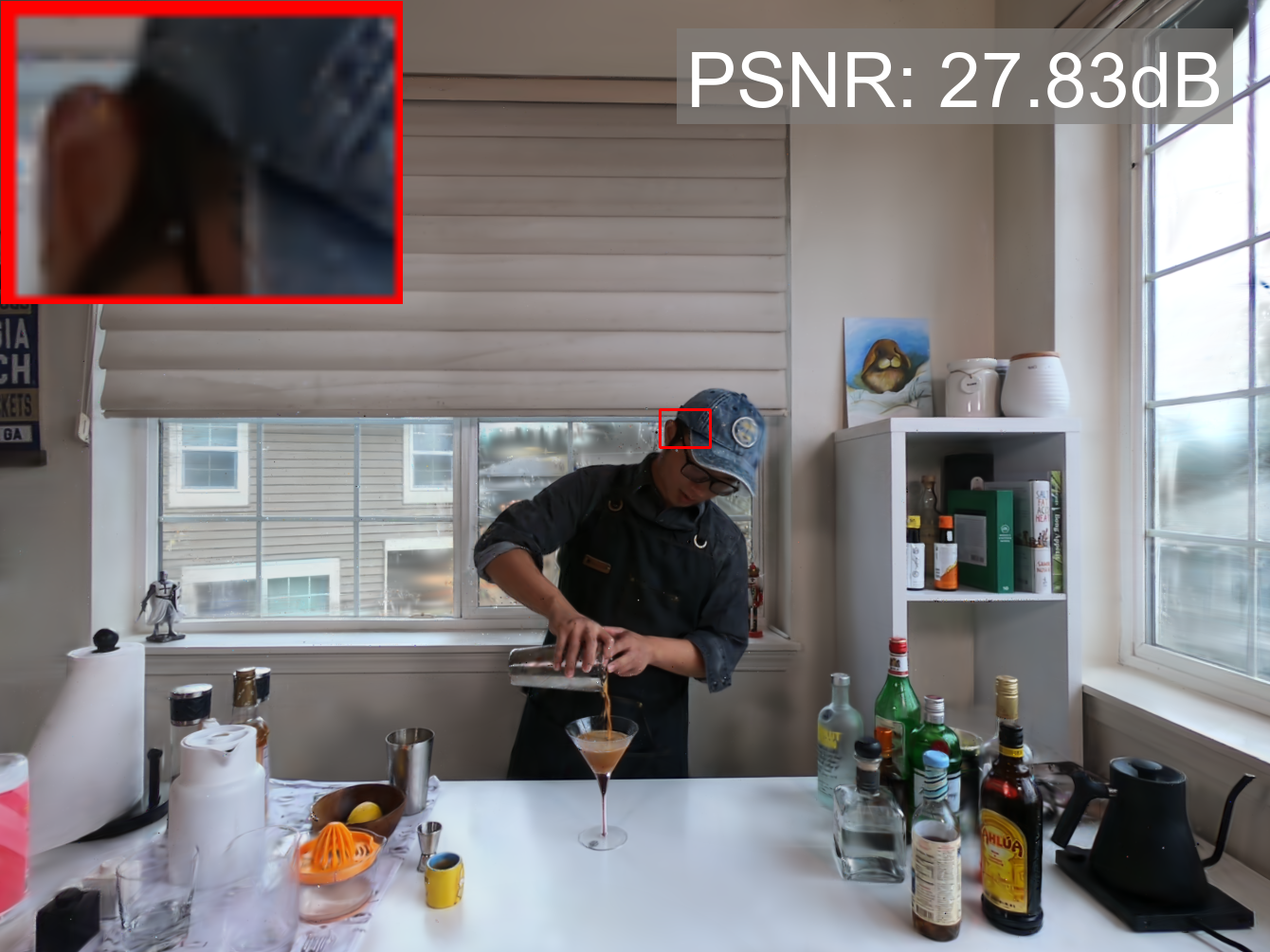}}
  
  \caption{\textbf{Quantitative and qualitative results of the ablation study of distance regularization} conducted on the \textit{coffee martini} scene. Metrics are averaged over  total 300 frames.}
  \label{fig:dist_abl}
  
  \vspace{-2mm}
\end{figure}
\subsubsection{Global Patching Strategy}
We validated the effectiveness of our global patching strategy by comparing the image quality and storage requirements against independent generation approaches (re-spawning free 3DGs for each frame). Specifically, we compared our method (globally generating with $\tau_{grad} = 0.001$) against three independent baselines: 
(I1) independently generating for each frame with the same threshold $\tau_{grad} = 0.001$; 
(I2) using a lowered threshold $\tau_{grad} = 0.00015$ (configuration of 3DGStream); 
and (I3) using a further lowered threshold $\tau_{grad} = 0.00010$ to match the storage consumption of our method. 
The qualitative results are presented in Fig. \textcolor{red}{\ref{fig:s2_abl_global}}.

Clearly, independent free 3DGs generation methods fall short compared to ours.
Variant I1, with its high threshold, yields fewer free 3DGs, making it hard to model coffee,
while Variant I2 and I3, by lowering the threshold, produce more free 3DGs but with inaccurate color.
In contrast, our approach significantly enhances the reconstruction quality of emerging objects with minimal storage increase.
In independent methods, structured 3DGs are misaligned to fit new objects,
accompanied by color MLP-generated false hues,
and new free 3DGs inherit these flawed attributes, resulting in incorrect coffee color.
Conversely, in the global patching approach,
free 3DGs keep structured 3DGs mostly aligned,
and most new free 3DGs derive from previous frames' free 3DGs, continuously refining to correct colors.

We further applied these temporal metrics ($E_{warp}$ and T-LPIPS) to verify the Global Patching Strategy. Independent generation methods (Variants I1, I2, and I3) suffer from severe flickering and misalignment, resulting in high warping errors. In contrast, our approach, by enforcing temporal continuity through global patching, drastically reduces the warping error and achieves superior T-LPIPS scores, confirming its superiority in maintaining coherent dynamics.

Our global patching strategy brings better reconstruction and temporal consistency to emerging objects.
Quantitatively, our method achieves lower Temporal LPIPS (0.00139 vs. 0.00145) and Warping Error (0.00659 vs. 0.00723) compared to 3DGStream.
In Fig. \textcolor{red}{\ref{fig:timeconsistency}}, we provide a visual comparison with 3DGStream~\cite{3dgstream} on adjacent frames of coffee martini, further demonstrating our method's superior temporal consistency and reconstruction quality.

\par 
\subsubsection{Distance Regularization}
We validated the effectiveness of distance regularization by training the variant, namely \textit{w/o} dist on the \textit{coffee\_martini} scene without distance regularization. As shown in Fig. \ref{fig:dist_abl}, the experiments demonstrate that our distance regularization indeed plays a significant role in noise reduction.

\par

\section{Conclusion and Limitation}
\textit{Conclusion.} In this paper, we introduce Struct-GStream, a method for efficient FVV streaming at low bitrates.
Struct-GStream represents scenes by dividing 3DGs into two distinct parts:
Structured 3DGs for representing the basic scene
and free 3DGs for patching inadequate scene reconstruction
and handling emerging objects. Owing to this architecture,
Struct-GStream achieves online training in less than 10 seconds per frame
and real-time rendering at over 100 FPS,
with superior rendering quality and an average storage cost of approximately 3.0--5.4 MB per frame at megapixel resolution across the evaluated datasets.
\par
\textit{Limitation.} During the experimental process, we observed that our results were constrained by the quality of the initial frame reconstruction. Particularly in the indoor corners with sparse perspectives and the unseen outdoor areas, these regions were challenging to reconstruct, leading to jitter during the online optimization process.
Many methods use a set of spherical shell point clouds to enclose scenes and assist in constructing outdoor scenes, such as  RealTime4DGS~\cite{realtime4dgs}. We consider this approach suboptimal, as it only enhances PSNR but does not contribute to the actual reconstruction quality; the rendering of areas outside the window remains blurry.
Any method that can improve the quality of the initial frame reconstruction is eagerly anticipated, as it would benefit our approach.
Furthermore, although our method has outperformed current state-of-the-art online training methods in terms of storage and training time,
there is still a considerable gap when compared to the standards of practical free-viewpoint video systems.

\bibliographystyle{eg-alpha-doi} 
\bibliography{egbibsample}       




\end{document}